\documentclass{article}

\PassOptionsToPackage{numbers,compress}{natbib}
\usepackage[main,final]{neurips_2026}

\usepackage[utf8]{inputenc}
\usepackage[T1]{fontenc}
\usepackage{amsmath}
\usepackage{amssymb}
\usepackage{amsfonts}
\usepackage{amsthm}
\usepackage{mathtools}
\usepackage{booktabs}
\usepackage{caption}
\usepackage{subcaption}
\usepackage{graphicx}
\usepackage[table]{xcolor}
\definecolor{gain}{RGB}{0,140,60}
\usepackage{tabularx}
\usepackage{multirow}
\usepackage{tcolorbox}
\tcbuselibrary{breakable}
\usepackage{enumitem}
\usepackage{wrapfig}
\usepackage{float}
\usepackage{needspace}
\usepackage{algorithm}
\usepackage{algorithmic}
\usepackage{xspace}
\usepackage{microtype}
\usepackage{nicefrac}
\usepackage{url}
\usepackage{hyperref}
\usepackage[capitalize,noabbrev]{cleveref}

\newcommand{\RETURN}{\STATE \textbf{return}}
\newcommand{\method}{VeoPlace\xspace}

\title{See it to Place it: Evolving Macro Placements with Vision-Language Models}

\author{%
  Ikechukwu Uchendu\textsuperscript{1,2,$\ast$} \And
  Swati Goel\textsuperscript{1} \And
  Karly Hou\textsuperscript{1} \And
  Ebrahim Songhori\textsuperscript{3,$\dagger$} \AND
  Kuang-Huei Lee\textsuperscript{3} \And
  Joe Wenjie Jiang\textsuperscript{3,$\dagger$} \And
  Vijay Janapa Reddi\textsuperscript{1} \And
  Vincent Zhuang\textsuperscript{3,$\dagger$}
}

\begin{document}

\maketitle
{\footnotetext[1]{Harvard University.}\footnotetext[2]{Kempner Institute, Harvard University.}\footnotetext[3]{Google DeepMind.}\renewcommand{\thefootnote}{\fnsymbol{footnote}}\footnotetext[1]{Work done partly as a Student Researcher at Google. Correspondence: \texttt{iuchendu1@gmail.com}.}\footnotetext[2]{Work done while at Google DeepMind.}}

\begin{abstract}
We propose using Vision-Language Models (VLMs) for macro placement in chip floorplanning, a complex optimization task that has recently shown promising advancements through machine learning methods. Because human designers rely heavily on spatial reasoning to arrange components on the chip canvas, we hypothesize that VLMs with strong spatial reasoning capabilities can effectively complement existing placement algorithms. We introduce \method{} (Visual Evolutionary Optimization Placement), a novel framework that uses a VLM---without any fine-tuning---to guide the actions of a base placer by constraining it to subregions of the chip canvas. The VLM proposals are iteratively optimized through an evolutionary search strategy with respect to resulting placement quality. Using global half-perimeter wirelength (gHPWL) after standard-cell placement and legalization as the evaluation metric, \method{} boosts ChiPFormer performance on 9 of 10 benchmarks with peak reductions exceeding 32\%. We further demonstrate that \method{} generalizes to analytical placers, improving DREAMPlace on all 8 evaluated Superblue benchmarks with gains up to 4.4\%. Our approach opens new possibilities for electronic design automation tools that leverage foundation models to solve complex physical design problems.

\end{abstract}

\section{Introduction}
\label{sec:introduction}
Computer chip floorplanning is a critical step in the integrated circuit design process, involving the strategic arrangement of macros on the chip canvas. Determining the optimal placement is a complex multi-objective problem, in which performance, power, and area (PPA) must be optimized while minimizing routing congestion. The vast combinatorial design space makes manual chip floorplanning a time-consuming and expertise-driven task while posing a significant challenge for automated methods.

A variety of approaches have been proposed for automated chip floorplanning, including black-box optimization \citep{wiremask}, analytical methods \citep{lin_dreamplace_2019,cheng2018replace,lu2015eplace}, and learning-based methods \citep{circuit_training_paper,lai2022maskplace,chipformer,chip_placement_with_diffusion}.
Among these, learning-based approaches have achieved state-of-the-art performance, but have a severe limitation: policies trained from scratch struggle to generalize to \textit{unseen chips} without additional interaction, an issue exacerbated by the limited training data available in chip design.
In contrast, human designers leverage high-level prior knowledge and spatial reasoning to efficiently tackle new design spaces.
Our work aims to bridge this gap by harnessing Vision-Language Models (VLMs) to provide spatial reasoning and guide the exploration of existing placement algorithms. While human designers develop placement intuition over years of experience, VLMs can condition on dozens of prior floorplans and their evaluation metrics in a single prompt, allowing them to extract visual placement patterns and refine proposals through in-context learning.

Existing learning-based approaches first pre-train models on a set of training chips, then fine-tune on sampled placements from unseen chips \citep{circuit_training_paper,lai2022maskplace,chipformer}. We consider a general formulation of this setup: for an unseen chip and a fixed budget of $B$ placement attempts, what is the best possible placement that can be generated? We posit that efficiently using this budget of online attempts requires spatial reasoning as well as learning from prior attempts, both of which have been exhibited by modern VLMs.

\begin{figure}[!htbp]
    \centering
    \begin{subfigure}[t]{0.32\textwidth}
        \centering
        \includegraphics[width=\linewidth]{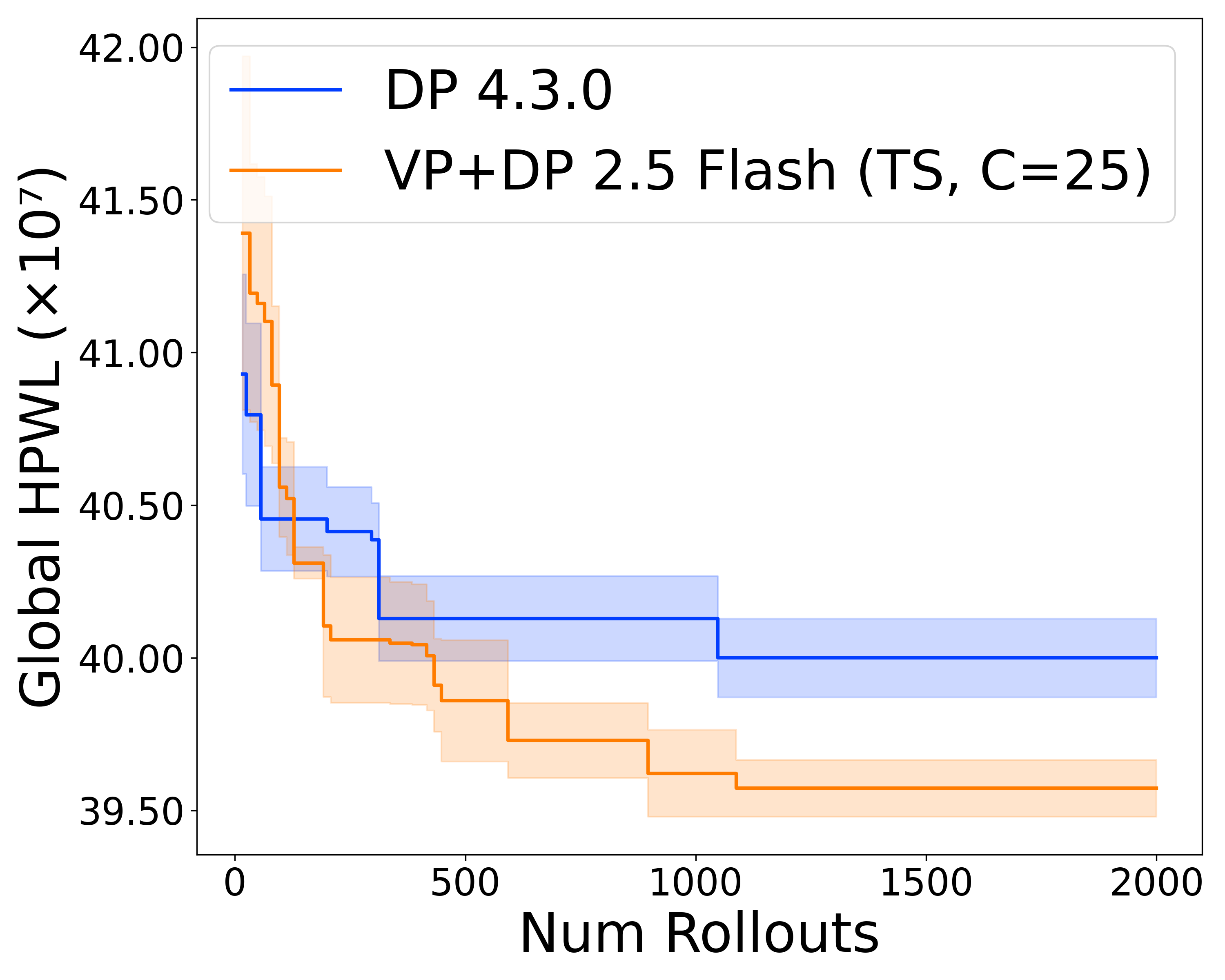}
        \caption{\texttt{superblue5}}
    \end{subfigure}
    \hfill
    \begin{subfigure}[t]{0.32\textwidth}
        \centering
        \includegraphics[width=\linewidth]{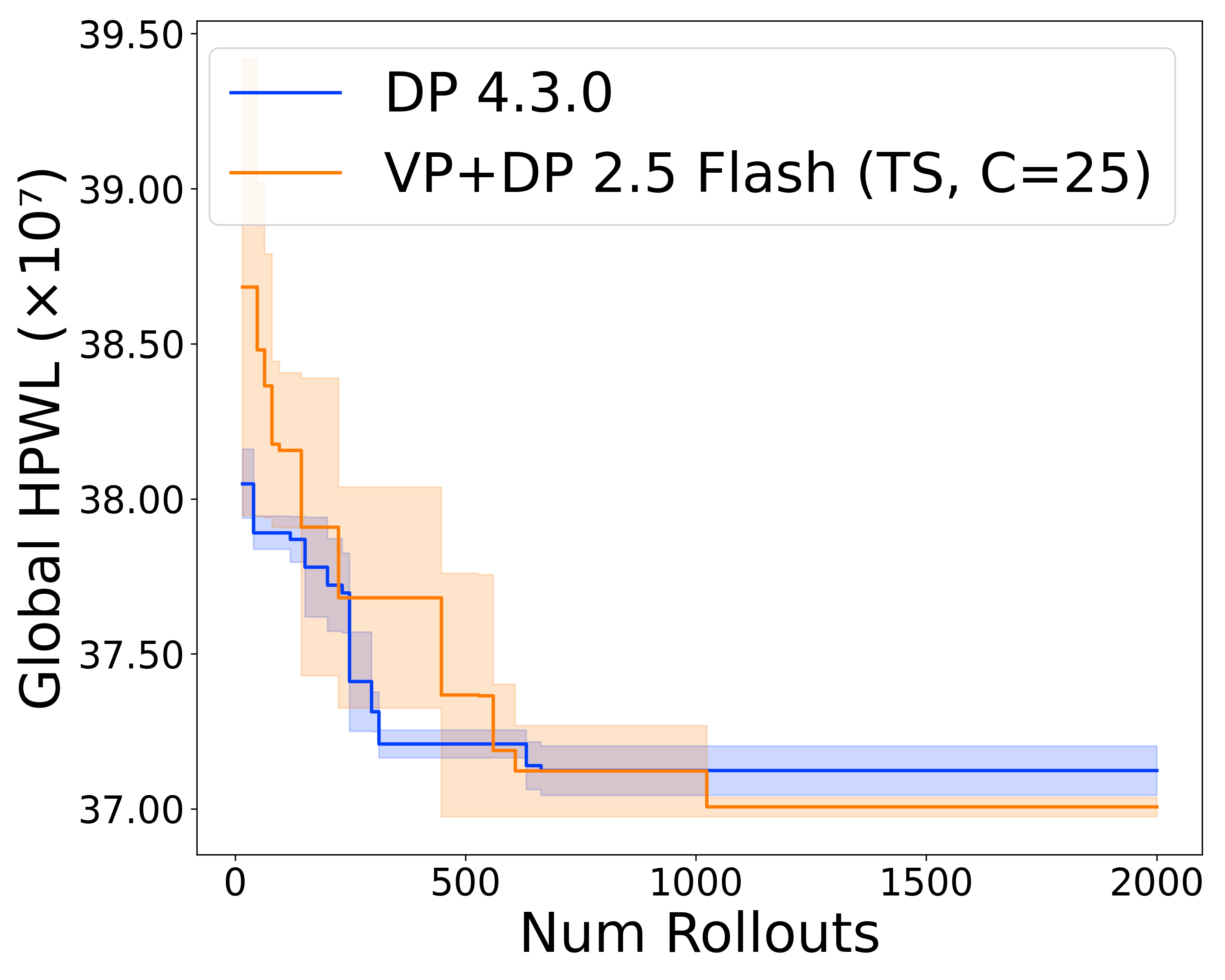}
        \caption{\texttt{superblue16}}
    \end{subfigure}
    \hfill
    \begin{subfigure}[t]{0.32\textwidth}
        \centering
        \includegraphics[width=\linewidth]{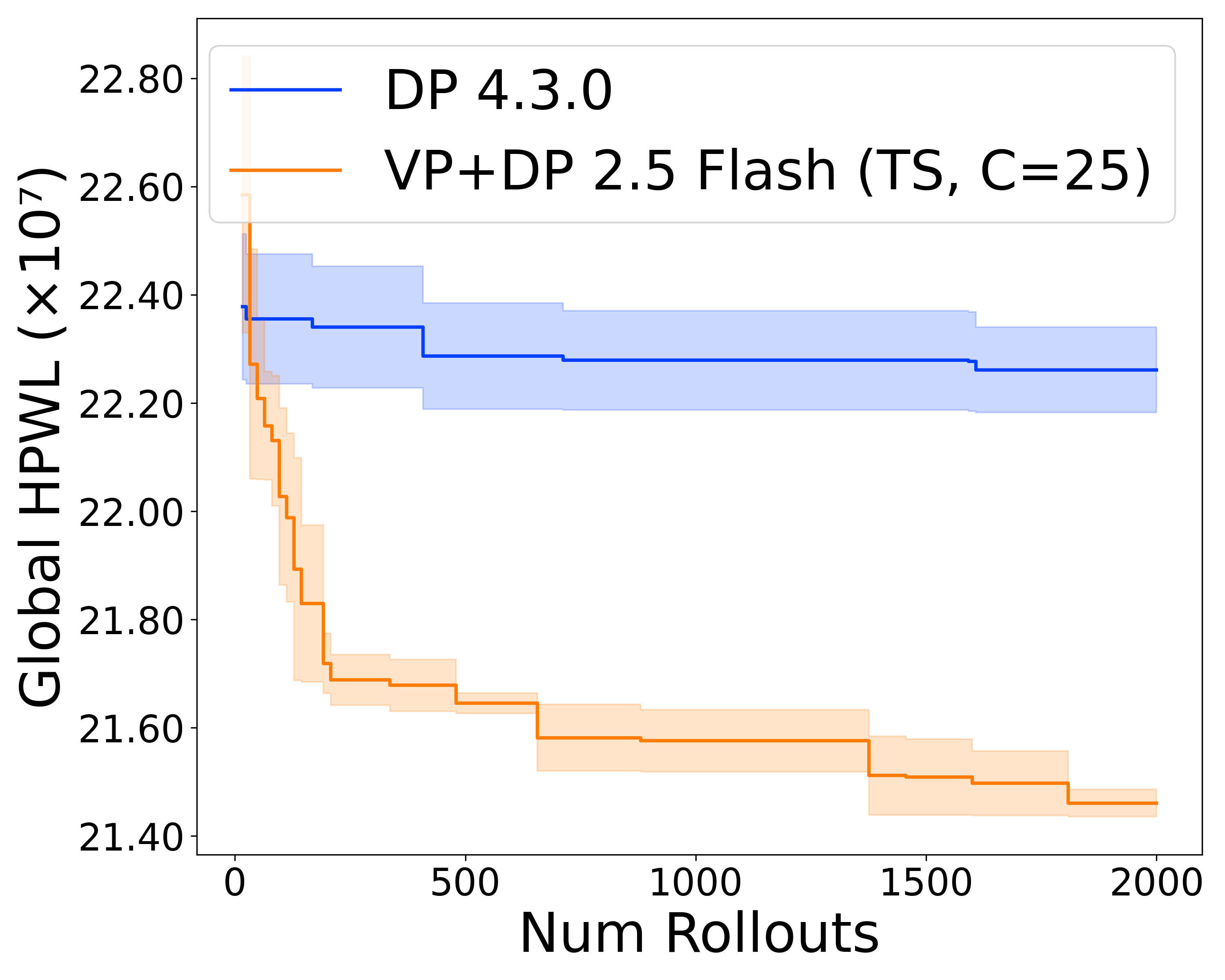}
        \caption{\texttt{superblue18}}
    \end{subfigure}
    \captionsetup{font=footnotesize}
    \caption{\method-guided DREAMPlace on (a) \texttt{superblue5}, (b) \texttt{superblue16}, and (c) \texttt{superblue18}. VP+DP denotes \method{} with DREAMPlace; DP~4.3.0 is standalone DREAMPlace. We report mean and standard error of the best gHPWL observed ($\times 10^7$, lower is better) across three random seeds with 2,000 rollouts each, using Gemini 2.5 Flash. The continued decrease in gHPWL demonstrates how \method{} improves placement quality over the fixed rollout budget, whereas standalone DREAMPlace converges early.}
    \label{fig:superblue_wirelength_progress}
\end{figure}

\begin{figure}[!htbp]
    \centering
    \includegraphics[width=0.75\textwidth]{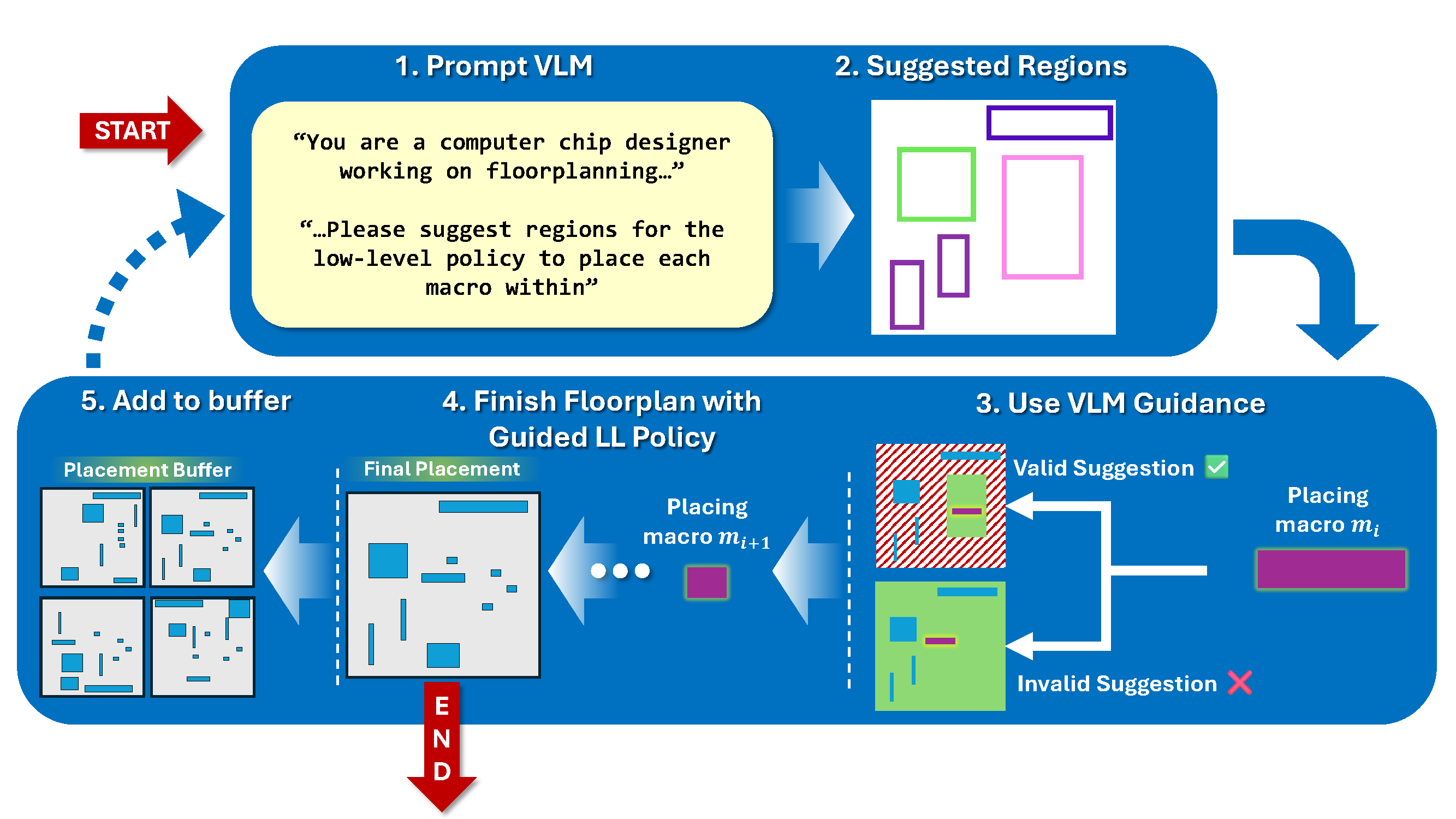}
    \captionsetup{font=footnotesize}
    \caption{\method framework overview. The VLM suggests placement regions (1-2) to constrain a low-level placer (3) for macro placement (4). A history buffer that stores the existing population of placements (5) facilitates evolutionary in-context improvement, creating a feedback loop to improve placement quality.}
    \label{fig:veoplace}
\end{figure}

We introduce \method (Visual Evolutionary Optimization Placement), a novel framework that uses a high-level VLM planner to guide a low-level placer by constraining it to promising regions. Crucially, these VLM proposals are iteratively refined through an evolutionary process, as visualized in \Cref{fig:veoplace}. \method requires no fine-tuning of the VLM (we use the public Gemini models \citep{team2023gemini}) and uses an independent low-level placer (e.g. ChiPFormer~\citep{chipformer} or DREAMPlace~\citep{lin_dreamplace_2019}). On open-source benchmarks (ISPD 2005~\citep{ispd2005_benchmarks}, ICCAD 2004~\citep{adya2002consistent_iccad04, adya2004unification_iccad04}, and Ariane~\citep{ariane_cva6}), \method boosts ChiPFormer on 9 of 10 benchmarks with peak wirelength reductions exceeding 32\%. Furthermore, we show \method generalizes to analytical placers, improving DREAMPlace performance on all eight ICCAD 2015 Superblue benchmarks, as illustrated in \Cref{fig:superblue_wirelength_progress} and quantified in \Cref{tab:superblue_results}, with gains up to 4.4\%. Our main contributions are:

\begin{itemize}[leftmargin=*]
    \item \textbf{Unified VLM-guided placement framework}: We introduce \method, a framework that uses multimodal foundation models to guide specialized placement algorithms through spatial reasoning, without requiring fine-tuning. We instantiate this idea with action masking for learning-based policies and soft anchor constraints for analytical placers.

    \item \textbf{Plug-and-play performance gains}: \method{} improves DREAMPlace~4.3.0 on all eight ICCAD 2015 Superblue benchmarks (up to 4.4\%) and boosts ChiPFormer on 9 of 10 benchmarks (up to 32\%), without VLM fine-tuning or modifying the underlying placer. This shows that VLM guidance can improve both analytical and learning-based placers while remaining compatible with existing EDA workflows.
\end{itemize}

\section{Related Work}
\label{sec:related_work}

\paragraph{Vision-Language Models for Decision-Making.} Vision-Language Models (VLMs) are trained on vast datasets of text and images, and therefore contain rich priors valuable for tasks requiring both vision and language \citep{driess2023palm_e}. The use of VLMs to perform decision-making has been explored in several fields, including robotics, where VLMs interpret natural language commands within a visual scene to guide robot actions or planning \citep{say_can, jiang2022vima, shridhar2022cliport, huang2022inner_monologue, brohan2023rt, kim2024openvla, team2025gemini, liang2024learning}. Systems such as SayCan \citep{say_can} and RT-2 \citep{brohan2023rt} demonstrate how VLMs can translate high-level instructions into actionable plans that low-level controllers can execute.

Our work leverages these VLM capabilities in a similar hierarchical approach. The VLM perceives chip placement images along with their performance metrics, analyzes spatial arrangements, and provides high-level guidance to a low-level placer in the form of suggested bounding regions for macros. These bounding regions constrain the low-level placer, effectively creating a division of labor where the VLM handles high-level spatial reasoning while a specialized placer executes precise placement decisions within these constraints.

\begin{wraptable}[21]{r}{0.4\textwidth}
    \vspace{-1.2em}
    \centering
    \captionsetup{font=small,skip=2pt}
    \small
    \renewcommand{\arraystretch}{0.98}
    \caption{Comparison of macro placement approaches. RL: Reinforcement Learning, MCTS: Monte Carlo Tree Search, IL: Imitation Learning, BBO: Blackbox Optimization, VLM: Vision-Language Model, LLM: Large Language Model.}
    \label{tab:placement_approaches_full}
    \begin{tabular}{@{}ll@{}}
        \toprule
        \textbf{Method}                                                  & \textbf{Category} \\
        \midrule
        DREAMPlace \citep{lin_dreamplace_2019}                           & Analytical        \\
        NTUPlace3 \citep{chen2008ntuplace3}                              & Analytical        \\
        RePlace \citep{cheng2018replace}                                 & Analytical        \\
        SP-SA \citep{murata2002vlsi}                                     & Packing           \\
        \midrule
        ChiPFormer \citep{chipformer}                                    & IL                \\
        DeepPR \citep{cheng2021joint_deep_pr}                            & RL                \\
        EfficientPlace \citep{rl_within_tree_search_for_macro_placement} & RL + MCTS         \\
        GraphPlace \citep{circuit_training_paper}                        & RL                \\
        MaskPlace \citep{lai2022maskplace}                               & RL                \\
        \midrule
        EvoPlace \citep{yao2025evolution_with_llm}                       & LLM + Analytical  \\
        MacroRegulate \citep{macro_regulator}                            & RL + Analytical   \\
        WireMaskBBO \citep{wiremask}                                     & BBO               \\
        \midrule
        \multirow{2}{*}{\method (Ours)}                                  & VLM + IL          \\
        & VLM + Analytical  \\
        \bottomrule
    \end{tabular}
\end{wraptable}

\paragraph{Automated Chip Floorplanning.} Automating computer chip floorplanning has been studied through various approaches (Table \ref{tab:placement_approaches_full}), including analytical methods \citep{lin_dreamplace_2019}, black-box optimization techniques such as simulated annealing \citep{new_algo_for_floorplanning_SA} and genetic algorithms \citep{opt_of_floorplanning_using_GA}, more recent guided black-box methods \citep{wiremask}, and various learning-based methods \citep{circuit_training_paper, chipformer, rl_within_tree_search_for_macro_placement}. Within the learning-based category, a prominent line of work formulates chip floorplanning as a reinforcement learning (RL) problem where macros are sequentially placed onto a chip canvas \citep{circuit_training_paper, lai2022maskplace, chipformer}. Alternatively, recent works have proposed learning to refine existing chip floorplans, employing techniques such as diffusion models \citep{chip_placement_with_diffusion} or RL algorithms \citep{macro_regulator} for post-processing. Our approach can be viewed as a generalization of learning-based approaches (and can explicitly leverage them in the inner loop) by using a high-level VLM to guide them at test-time.

\paragraph{Pairing LLMs with Evolution.}
Pairing LLMs with evolutionary search has achieved success in fields such as program generation~\citep{romera2024mathematical, hemberg2024evolving,liventsev2023fully}, planning and reasoning~\citep{lee2025evolving}, scientific discovery~\citep{aiscientist_v2,gottweis2025towards}, robotics~\citep{nasiriany2024pivot}, and chip design~\citep{alphaevolve,yao2025evolution_with_llm,macro_regulator,wiremask}.
Most relevant to our work, EvoPlace~\citep{yao2025evolution_with_llm} uses LLMs to evolve optimization algorithms for analytical global placement. In contrast, \method uses a VLM to evolve the global placement \textit{solutions} directly, suggesting where macros should go based on visual reasoning over prior placements. These suggestions are evaluated, and high-performing ones are selected to inform the VLM's next generation of proposals, given feedback from an objective function.
Our selection strategy samples a population of high-performing, geometrically similar placements.
That is, \method explicitly focuses on evolution in a local region, a principle that has been shown to be effective in sparse Gaussian processes \citep{wei2024scalable} and island models in genetic algorithms~\citep{romera2024mathematical,lee2025evolving,tanese1989distributed,cantu1998survey}.

\section{Preliminaries}
\label{sec:preliminaries}
\paragraph{Macro Placement.} We consider macro placement in chip floorplanning, where a set of \emph{macros} $M = \{m_1, \dots, m_N\}$, defined by their dimensions and connectivity, are placed on a 2D chip \emph{canvas}. Connectivity is given by a \emph{netlist} $G = (V, E)$, a hypergraph over circuit objects $V = M \cup C$, where $C$ denotes standard cells and each hyperedge (\emph{net}) connects a subset of objects. The objective is to find a macro placement $P = \{p_1, \dots, p_N\}$, where $p_i$ is the bottom-left corner of macro $m_i$, that minimizes wirelength. We measure wirelength using half-perimeter wirelength (HPWL), the standard proxy for the total interconnect needed to connect components within each net, and a crucial metric for a chip's performance, power, and area (PPA) \citep{lin_dreamplace_2019,circuit_training_paper}.

Learning-based macro placers often formulate placement as a Markov Decision Process (MDP)~\citep{circuit_training_paper, chipformer, lai2022maskplace}: macros are placed sequentially on a discretized canvas, with state $s_t$ encoding the partial placement, current macro features, and possibly netlist structure. An action $a_t$ selects the grid cell for the current macro's reference point. After all macros are placed, the terminal reward is based on HPWL, which is the sum over nets of the half-perimeter of the smallest bounding box enclosing each net's connected components. In this work, we report global HPWL (gHPWL) after standard-cell placement and legalization, ensuring that the evaluated placement is complete and overlap-free.

\paragraph{Inference-time Optimization.} 
Online RL requires many environment interactions and model updates to produce optimal placements for new netlists. 
Recent work suggests that offline RL pre-training provides strong zero-shot performance but benefits from fine-tuning on a small amount of online interaction~\citep{chipformer}. 
We consider an inference-time optimization setting that uses a \textbf{hierarchical approach}: we are allowed a fixed budget of placement evaluations, but do not fine-tune either the VLM (our high-level strategic guide) or the low-level placer. 

\paragraph{Analytical Placer.}
Analytical placers formulate placement as a nonlinear optimization problem, minimizing wirelength subject to density constraints. We use DREAMPlace~4.3.0~\citep{lin_dreamplace_2019}, a state-of-the-art GPU-accelerated analytical placer. In our framework, we guide DREAMPlace by incorporating VLM suggestions as soft anchor constraints in the optimization objective, pulling macros toward suggested target locations while still allowing the solver to find low-wirelength, density-aware arrangements.

\paragraph{Learning-Based Placer.}
ChiPFormer~\citep{chipformer} is an autoregressive Transformer that formulates macro placement as offline reinforcement learning via a Decision Transformer objective~\citep{chen2021decision,lee2022multi}, achieving state-of-the-art performance across multiple chip designs. We adopt it as our learning-based low-level placer due to its multi-task generality. Crucially, ChiPFormer outputs a probability distribution over grid cells for each macro. In our framework, the VLM constrains this distribution by masking out regions outside its suggested placement areas, steering the policy toward better design choices without requiring any fine-tuning.

\section{Method}
\label{sec:method}
In this section, we describe \method, our evolutionary framework for using VLM spatial reasoning in chip floorplanning. As shown in \Cref{fig:veoplace,alg:veoplace_overview}, \method maintains a history buffer of evaluated placements and repeatedly improves it over a fixed rollout budget. Algorithm~\ref{alg:veoplace_overview} initializes this buffer and iterates over batches of rollouts (Lines~1--2). Each iteration first samples an unguided batch from the base placer, legalizes each placement, evaluates its gHPWL, and stores the result in the history buffer (Lines~3--8). The method then builds a context from the current history and queries the VLM for a batch of macro-region suggestions (Lines~9--10). Finally, it runs a guided batch in which the low-level placer is constrained by those VLM suggestions, legalizes and evaluates the resulting placements, and adds them back to the history buffer (Lines~11--16). This feedback loop lets later VLM queries condition on previous successes and failures.

\Needspace{26\baselineskip}
\subsection{VLM and Low-Level Placer Interface}
\label{ssec:vlm_and_ll_policy_interface}

\begin{wrapfigure}[16]{r}{0.50\textwidth}
    \vspace{-2.0\baselineskip}
    \centering
    \captionsetup{skip=0pt}
    \captionof{algorithm}{\method}
    \label{alg:veoplace_overview}
    \vspace{0.12em}
    \hrule
    \vspace{0.2em}
    \scriptsize
    \begin{algorithmic}[1]
    \REQUIRE $V$: VLM; $\mathcal{P}$: Placer; $G$: Netlist; $C$: Context length; $E$: Total rollout budget; $q$: Batch size
    \STATE Initialize history buffer $H \gets \emptyset$
    \FOR{$t = 1$ to $E/(2q)$}
    \STATE \textbf{Phase 1: Unguided batch}
    \FOR{$j = 1$ to $q$}
    \STATE $P_{t,j} \gets \textsc{PlaceMacros}(\mathcal{P},G,\emptyset)$
    \STATE $P_{t,j} \gets \textsc{Legalize}(P_{t,j})$
    \STATE $H \gets H \cup \{(P_{t,j},\mathrm{gHPWL}(P_{t,j}))\}$
    \ENDFOR
    \STATE context $\gets$ \textsc{BuildContext}$(H, C)$
    \STATE $\{S_1,\ldots,S_q\} \gets V(\text{context}, G)$
    \STATE \textbf{Phase 2: VLM-guided batch}
    \FOR{$j = 1$ to $q$}
    \STATE $P_{t,j} \gets \textsc{PlaceMacros}(\mathcal{P},G,S_j)$
    \STATE $P_{t,j} \gets \textsc{Legalize}(P_{t,j})$
    \STATE $H \gets H \cup \{(P_{t,j},\mathrm{gHPWL}(P_{t,j}))\}$
    \ENDFOR
    \ENDFOR
    \RETURN~best placement in $H$
    \end{algorithmic}
    \vspace{0.35em}
    \hrule
\end{wrapfigure}

\noindent The manner in which VLM suggestions are incorporated depends on the low-level placer: for analytical placers, suggestions become soft anchor constraints in the optimization objective; for learning-based policies, suggestions mask the action space to constrain sampling. We demonstrate \method with both approaches.
Algorithm \ref{alg:veoplace_overview} summarizes the overall procedure, while \textsc{PlaceMacros} is instantiated for analytical and learning-based placers in Algorithms \ref{alg:placemacros_analytical} and \ref{alg:placemacros_learning}, respectively.
We set $q=8$ because each Gemini query returns eight candidate suggestions. Thus, one \method iteration consists of eight unguided rollouts, one VLM query, and eight guided rollouts. With $E=2{,}000$ total rollouts, this yields $E/(2q)=125$ VLM calls.

We use a 256-macro VLM interface, which matches ChiPFormer's sequence length in the learning-based track and keeps prompting cost fixed across both tracks. When a design contains more than 256 macros, we choose the guided subset by sorting macros by area, with connectivity used as a tie-breaker, so the largest and most connected macros receive VLM guidance first. After each full placement, we legalize placement $P$ using DREAMPlace's built-in legalizer to ensure a valid macro and standard-cell placement.

\begin{wrapfigure}[9]{r}{0.50\textwidth}
    \vspace{-2\baselineskip}
    \centering
    \captionsetup{skip=0pt}
    \scriptsize
    \captionof{algorithm}{\textsc{PlaceMacros} (Analytical)}
    \label{alg:placemacros_analytical}
    \vspace{0.12em}
    \hrule
    \vspace{0.2em}
    \begin{algorithmic}[1]
    \REQUIRE G: Netlist; $S$: VLM suggestions (possibly empty)
    \IF{$S = \emptyset$}
    \STATE Solve standard DREAMPlace objective to obtain $P$
    \ELSE
    \STATE Convert each $s_i \in S$ to anchor $\hat{x}_i$
    \STATE Solve anchored objective with $A$ to obtain $P$
    \ENDIF
    \RETURN~$P$
    \end{algorithmic}
    \vspace{0.2em}
    \hrule
\end{wrapfigure}

\subsubsection{Guiding Analytical Placers}
\label{ssec:analytical_placer}

\begin{figure*}[t]
    \centering
    \includegraphics[width=\textwidth]{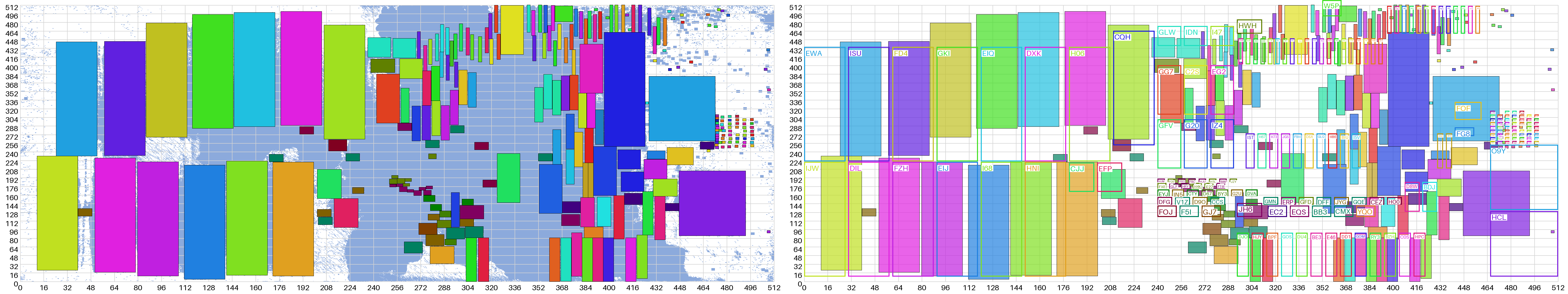}
    \caption{VLM guidance visualization on \texttt{superblue1}. \textbf{Left:} Final macro placement produced by \method + DREAMPlace 4.3.0. \textbf{Right:} The same placement with VLM-suggested regions overlaid. The VLM proposes target positions for selected macros, and DREAMPlace's loss function is modified with anchor weights to keep macros close to VLM suggestions while allowing standard cells to optimize freely (standard cells removed for visual clarity).}
    \label{fig:vlm_guidance_main}
\end{figure*}

For analytical placers, we interpret VLM suggestions as target macro locations. Let $x$ denote macro locations, $y$ denote standard-cell locations, $E$ denote the set of nets, and $M$ denote the set of macros. The standard DREAMPlace objective minimizes $\sum_{e \in E} \mathrm{WL}(e; x, y) + \lambda \, D(x, y)$, where $D$ is the density penalty. We incorporate VLM suggestions by adding an anchor term with weight $\lambda_A$:
\begin{equation}
    \min_{x, y} \; \sum_{e \in E} \mathrm{WL}(e; x, y) + \lambda \, D(x, y) + \lambda_A \, A(x; \{\hat{x}_i\}),
\end{equation}
where $A$ penalizes distance from each guided macro $i$ to its suggested location $\hat{x}_i$. Since macro locations are represented by their bottom-left reference coordinate, we convert each VLM-proposed macro box into an anchor by taking its bottom-left corner. Let $M_g \subseteq M$ be the set of macros with valid VLM suggestions. We use a quadratic anchor, $A(x; \{\hat{x}_i\}_{i \in M_g}) = \sum_{i \in M_g} \lVert x_i - \hat{x}_i \rVert_2^2$, which pulls guided macros toward the suggested targets. The weight $\lambda_A$ controls the strength of VLM guidance, where $\lambda_A = 0$ recovers unconstrained DREAMPlace. DREAMPlace then solves this anchored optimization where macros are softly pulled toward VLM targets (visualized in \Cref{fig:vlm_guidance_main}).

\subsubsection{Guiding Learning-Based Placers}
\label{ssec:learning_based_policies}

\begin{wrapfigure}[9]{r}{0.50\textwidth}
    \vspace{-2.5\baselineskip}
    \centering
    \captionsetup{skip=0pt}
    \scriptsize
    \captionof{algorithm}{\textsc{PlaceMacros} (Learned)}
    \label{alg:placemacros_learning}
    \vspace{0.12em}
    \hrule
    \vspace{0.2em}
    \begin{algorithmic}[1]
    \REQUIRE $\pi$: Policy; G: Netlist; $S$: VLM suggestions
    \STATE Order macros by size, connectivity
    \STATE Initialize $P \gets \emptyset$
    \FOR{macro $m_t$}
    \IF{$S$ provides valid suggestion $s_t$}
    \STATE $p_t \sim \pi(\cdot | m_t, P, s_t)$
    \ELSE
    \STATE $p_t \sim \pi(\cdot | m_t, P)$
    \ENDIF
    \STATE $P \gets P \cup \{ (m_t, p_t) \}$
    \ENDFOR
    \RETURN~$P$
    \end{algorithmic}
    \vspace{0.2em}
    \hrule
\end{wrapfigure}

For learning-based placers, the low-level policy $\pi$ places macros sequentially by sampling from a distribution over grid locations. When a macro has a VLM suggestion, we represent it as a bounding box region $s_t$ on the chip canvas (visualized in \Cref{fig:suggested_regions}) and constrain the policy to that region, denoted by $\pi(\cdot | m_t, P_e, s_t)$ (Algorithm \ref{alg:placemacros_learning}). We set out-of-region logits to $-\infty$ and renormalize over valid in-region cells before sampling, so $\pi$ still controls the exact placement coordinate within the VLM-suggested region.

Because the low-level policy autoregressively places macros, a VLM suggestion $s_t$ may be invalid for any $t>1$ due to already-placed macros overlapping the region. If no valid in-region cell exists, the macro $m_t$ is placed by sampling from the original, unconstrained policy distribution, $p_t \sim \pi(\cdot | m_t, P_e)$. Empirically, this fallback is triggered for ${\sim}20\%$ of suggestions once evolution stabilizes (\Cref{fig:invalid_suggestions}). After completing each placement $P_e$, we calculate its quality, $\mathrm{gHPWL}(P_e)$, and add $(P_e, \mathrm{gHPWL}(P_e))$ to the history buffer $H$.

\definecolor{rowgray}{gray}{0.95}

\begin{wraptable}[13]{r}{0.56\textwidth}
\vspace{-2em}
    \centering
    \captionsetup{font=scriptsize,skip=2pt}
    \scriptsize
    \caption{Components of the structured prompt for \method{}.}
    \label{tab:prompt_components}
    \rowcolors{2}{rowgray}{white}
    \begin{tabularx}{\linewidth}{@{} l X @{}}
  \toprule
  \textbf{Component} & \textbf{Description} \\
  \midrule
  \textbf{Grid Repr.} &
  \begin{itemize}[leftmargin=*, topsep=0pt, partopsep=0pt, parsep=0pt, itemsep=1pt]
      \item 84$\times$84 grid for ChiPFormer experiments (matching \citep{chipformer}); 512$\times$512 grid for DREAMPlace experiments.
      \item Macros positioned by bottom-left corner coordinates.
      \item Origin (0,0) at the bottom-left.
  \end{itemize} \\
  \textbf{Visual Repr.} &
  \begin{itemize}[leftmargin=*, topsep=0pt, partopsep=0pt, parsep=0pt, itemsep=1pt]
      \item Image of canvas showing placed macros and colors.
      \item Visual context for spatial relationships and patterns.
  \end{itemize} \\
  \textbf{Context} &
  \begin{itemize}[leftmargin=*, topsep=0pt, partopsep=0pt, parsep=0pt, itemsep=1pt]
      \item Grid specs and macro properties (dimensions, color).
      \item History of prior placements with performance metrics.
      \item Current state (locations) of placed macros.
  \end{itemize} \\
  \bottomrule
    \end{tabularx}
\end{wraptable}

\subsection{Structured Prompt}
\label{ssec:structured_prompt}

\method prompts a VLM to propose rectangular regions for the selected macro subset, conditioned on prior placements and their scores $\{(P_i,\mathrm{gHPWL}(P_i))\}$. Each VLM response specifies one region per selected macro, allowing a single query to provide coordinated spatial guidance for the entire selected subset rather than issuing separate queries per macro. The prompt is a simple structured scaffold: it specifies the objective, coordinate system, macro dimensions and color groups, prior placements with metrics, and the required rectangular-region output format. Its key characteristics are (1) this lightweight structure, which elicits spatial reasoning, and (2) the selection of in-context examples (detailed in \Cref{ssec:context_building_strategies}, with a full example in Appendix~\ref{appendix:example_prompt}). Our structured prompt guides the VLM with clear objectives, constraints, and a standardized format (Table \ref{tab:prompt_components}), transforming its general visual reasoning into useful spatial guidance. To ensure generalizability and avoid overfitting, we developed the prompt scaffold exclusively on \texttt{adaptec1}, using it only to validate the prompt interface (coordinate convention, output validity, and parser compatibility), and then applied the fixed scaffold unchanged to all other benchmarks.

\subsection{Context Selection Strategies for Evolution}
\label{ssec:context_building_strategies}

The core component of our evolutionary algorithm is prompting the VLM to generate a superior placement suggestion given a set of prior placements and their evaluations. Because each placement uses hundreds of tokens, only a small number can be provided to the model while maintaining reasonable inference cost. Given this limited budget, the examples should be (1) high-quality, so the model improves upon good placements, and (2) informative enough for the model to effectively deduce better placements via reasoning.

We compare multiple context selection strategies (FIFO, Random, Best, Diverse, Top Stratified) in \Cref{ssec:ablations}, and full definitions are provided in \Cref{appendix:context_selection_strategies}. We use \textbf{Top Stratified} as our default because it balances exploration (sampling across clusters) with exploitation (favoring high-performing placements within a promising cluster).

\section{Experiments}
\label{sec:experiments}

We design experiments to address four key questions: \textbf{(Q1)} Can VLM guidance improve the placement quality of both learning-based policies and analytical placers using only inference-time computation?
\textbf{(Q2)} Which \method{} design choices (prompt strategy, input modality, VLM, anchor weight, context selection strategy, and context length) most affect performance?
\textbf{(Q3)} How does \method{} perform on downstream PPA metrics?
\textbf{(Q4)} What runtime and API-cost overhead does VLM guidance introduce?

\textbf{Evaluation scope.} \method{} is a guidance framework rather than a standalone placer. We therefore evaluate the additional value of VLM guidance by comparing each underlying placer against the same placer augmented with VLM-generated spatial proposals, using the same rollout budget, legalization flow, and gHPWL evaluation protocol. This design tests whether an off-the-shelf VLM can serve as a history-conditioned spatial proposal operator for existing placers. This is nontrivial because the model must interpret prior placements and scores, identify useful geometric patterns, propose parseable macro regions under boundary and overlap constraints, and balance exploration with refinement without a benchmark-specific mutation heuristic.

\subsection{Experimental Setup}
\label{ssec:experimental_setup}

For both experimental tracks, \method generates an equal number of guided and unguided rollouts in each batch: 8 episodes from directly sampling the base placer, and 8 episodes guided by VLM suggestions (see \Cref{tab:gemini-hparams} for Gemini sampling parameters). All reported gHPWL values are computed after standard cells are placed and the full floorplan is legalized.

\subsubsection{Analytical Placer}
\label{sssec:setup_analytical}
We evaluate on the ICCAD 2015 Superblue benchmarks~\citep{kim2015iccad} using DREAMPlace 4.3.0~\citep{lin_dreamplace_2019} as the base placer. In this track, DREAMPlace is the low-level placer, so each rollout jointly optimizes macros and standard cells while incorporating VLM suggestions as soft anchor constraints in the DREAMPlace objective. We discretize the canvas to a $512 \times 512$ grid for VLM suggestions and use Gemini 3.6 Flash with $C{=}25$ context examples. VP+DP uses one global anchor weight $\lambda_A{=}0.01$ on all eight designs. All experiments run 2,000 rollouts across five seeds and report gHPWL (see \Cref{appendix:dreamplace_guidance_hyperparams} for full hyperparameters).

\subsubsection{Learning-Based Placer}
\label{sssec:setup_learning}
Following \citet{chipformer}, we evaluate on open-source chip benchmarks from the ISPD 2005 challenge~\citep{ispd2005_benchmarks} and ICCAD 2004~\citep{adya2002consistent_iccad04, adya2004unification_iccad04}, as well as the Ariane RISC-V CPU~\citep{ariane_cva6}.
For this track, the selected benchmarks vary in complexity, with tens to hundreds of macros or macro-like objects and up to hundreds of thousands of standard cells. We train ChiPFormer from scratch using the public repository\footnote{\url{https://github.com/laiyao1/chipformer}} and use an 84$\times$84 grid to match its original setup. We use Gemini 2.5 Flash with $C{=}1$, as even a single in-context example is enough to guide ChiPFormer to better solutions. We run 2,000 rollouts across three seeds (further details in \Cref{appendix:rollout_settings}; see \Cref{fig:suggested_regions} for a visualization of VLM-guided placement).

Our setup differs from \citet{chipformer} in two aspects.
(1) During the evolutionary search, we cluster standard cells for faster reward computation, following prior work~\citep{circuit_training_paper, chip_placement_with_diffusion}, but report unclustered results for final evaluation. (2) During final evaluation, ChiPFormer/\method macro locations are fixed and DREAMPlace places standard cells around them, following the two-stage evaluation protocol of \citet{circuit_training_paper}. This evaluates the macro placement produced by the learning-based track. Empirically, allowing movable macros results in significant changes to their final placements, confounding the actual efficacy of the base placer (see \Cref{fig:dreamplace_movement_analysis}).

\subsection{Q1: Does VLM Guidance Improve Placement Quality?}
\label{ssec:q1_main_results}

\begin{table}[H]
    \centering
    \captionsetup{font=scriptsize,skip=2pt}
    \scriptsize
    \renewcommand{\arraystretch}{0.92}
    \setlength{\tabcolsep}{1.3pt}
    \caption{\method-guided DREAMPlace (VP+DP) vs.\ DREAMPlace 4.3.0 (DP) and Superblue baselines reported by \citet{macro_regulator}. All columns report gHPWL ($\times 10^7$, lower is better). VP+DP uses Gemini 3.6 Flash and one global anchor weight $\lambda_A{=}0.01$ on all designs. VP+DP/DP report mean $\pm$ standard error of the best gHPWL observed across five seeds with 2{,}000 rollouts. Percentages compare VP+DP to matched DP 4.3.0.}
    \label{tab:superblue_results}
    \begin{tabularx}{\textwidth}{@{}l >{\centering\arraybackslash\hsize=1.45\hsize}X *{6}{>{\centering\arraybackslash\hsize=0.925\hsize}X}@{}}
\toprule
\textbf{Benchmark} & \textbf{VP+DP} & \textbf{DP 4.3.0} & \textbf{DP 4.0} & \textbf{AutoDMP} & \textbf{WireMask-EA} & \textbf{MaskPlace} & \textbf{MaskRegulate} \\
\midrule
superblue1 & \textbf{37.29$\pm$0.03}\,{\color{gain}\tiny($-$1.0\%)} & \underline{37.66$\pm$0.01} & 89.6$\pm$8.4 & 81.3$\pm$1.7 & 80.7$\pm$3.8 & 79.3$\pm$0.6 & 57.7$\pm$0.5 \\
superblue3 & \textbf{43.81$\pm$0.04}\,{\color{gain}\tiny($-$3.4\%)} & \underline{45.38$\pm$0.03} & 128.7$\pm$17.3 & 81.3$\pm$6.9 & 93.7$\pm$8.1 & 89.0$\pm$1.7 & 70.5$\pm$0.3 \\
superblue4 & \textbf{28.68$\pm$0.06}\,{\color{gain}\tiny($-$1.4\%)} & \underline{29.09$\pm$0.03} & 68.1$\pm$2.3 & 45.7$\pm$7.8 & 55.1$\pm$0.7 & 52.8$\pm$0.3 & 41.5$\pm$0.6 \\
superblue5 & \textbf{39.79$\pm$0.04}\,{\color{gain}\tiny($-$0.7\%)} & \underline{40.09$\pm$0.03} & 87.8$\pm$14.7 & 126.7$\pm$40.9 & 102.3$\pm$6.8 & 98.1$\pm$0.3 & 69.4$\pm$0.0 \\
superblue7 & \textbf{53.88$\pm$0.09}\,{\color{gain}\tiny($-$1.0\%)} & \underline{54.45$\pm$0.04} & 227.0$\pm$9.1 & 100.4$\pm$16.3 & 100.5$\pm$3.4 & 99.9$\pm$0.5 & 79.0$\pm$0.3 \\
superblue10 & \textbf{68.33$\pm$0.12}\,{\color{gain}\tiny($-$4.4\%)} & \underline{71.45$\pm$0.12} & 148.1$\pm$13.3 & 114.8$\pm$17.8 & 135.2$\pm$22.5 & 109.4$\pm$2.1 & 112.3$\pm$3.8 \\
superblue16 & \textbf{36.44$\pm$0.01}\,{\color{gain}\tiny($-$0.5\%)} & \underline{36.63$\pm$0.01} & 102.2$\pm$5.8 & 61.2$\pm$24.9 & 61.3$\pm$1.4 & 55.3$\pm$0.4 & 55.3$\pm$0.4 \\
superblue18 & \textbf{21.75$\pm$0.11}\,{\color{gain}\tiny($-$0.8\%)} & \underline{21.92$\pm$0.01} & 49.7$\pm$13.8 & 30.2$\pm$0.1 & 34.6$\pm$0.7 & 34.9$\pm$0.6 & 29.6$\pm$0.1 \\
\bottomrule
    \end{tabularx}
\end{table}

\paragraph{Analytical Placers.} As shown in \Cref{tab:superblue_results}, \method outperforms DREAMPlace 4.3.0 on all eight Superblue benchmarks, with improvements ranging from 0.5\% (superblue16) to 4.4\% (superblue10).
\method also achieves lower reported gHPWL than the published DREAMPlace 4.0, AutoDMP, WireMask-EA, MaskPlace, and MaskRegulate baselines included in \Cref{tab:superblue_results} on all eight benchmarks.
These published baselines have $1.35\times$ to $4.2\times$ the gHPWL of DREAMPlace 4.3.0 on every benchmark, showing that prior learning-based approaches have struggled to outperform the most recent state-of-the-art analytical placer.
Qualitative placement examples are shown in \Cref{fig:dp_visual_comparison}.
We ablate each design choice of \method{} in \Cref{ssec:ablations}.

\begin{table}[H]
    \centering
    \captionsetup{font=scriptsize,skip=2pt}
    \scriptsize
    \renewcommand{\arraystretch}{0.92}
    \setlength{\tabcolsep}{1.3pt}
    \caption{\method (VP) with ChiPFormer (CF). VP uses Gemini 2.5 Flash. Mean $\pm$ SE of best gHPWL ($\times 10^7$, lower is better) across three random seeds with 2{,}000 rollouts each. Percentages compare VP+CF to the better CF baseline. Best results are bolded.}
    \label{tab:main_results_hpwl_only}
    \begin{tabularx}{\textwidth}{@{}l >{\centering\arraybackslash\hsize=1.45\hsize}X *{2}{>{\centering\arraybackslash\hsize=0.775\hsize}X}@{}}
\toprule
& \textbf{VP+CF} & \multicolumn{2}{c}{\textbf{CF}} \\
\cmidrule(lr){3-4}
\textbf{Benchmark} & $(C{=}1)$ & No-FT & FT \\
\midrule
adaptec1 & 15.97$\pm$0.80\,{\color{red}\tiny(+12.2\%)} & \textbf{14.23$\pm$0.77} & 14.99$\pm$0.55 \\
adaptec2 & \textbf{11.82$\pm$0.89}\,{\color{gain}\tiny($-$13.8\%)} & 15.03$\pm$0.59 & 13.71$\pm$0.33 \\
adaptec3 & \textbf{24.10$\pm$0.21}\,{\color{gain}\tiny($-$6.0\%)} & 25.64$\pm$0.43 & 26.31$\pm$0.97 \\
adaptec4 & \textbf{19.63$\pm$0.20}\,{\color{gain}\tiny($-$8.7\%)} & 22.65$\pm$1.69 & 21.50$\pm$0.50 \\
\midrule
ibm01 & \textbf{0.31$\pm$0.01}\,{\color{gain}\tiny($-$22.5\%)} & 0.44$\pm$0.04 & 0.40$\pm$0.01 \\
ibm02 & \textbf{0.59$\pm$0.01}\,{\color{gain}\tiny($-$32.2\%)} & 0.91$\pm$0.02 & 0.87$\pm$0.06 \\
ibm03 & \textbf{0.80$\pm$0.02}\,{\color{gain}\tiny($-$16.7\%)} & 1.16$\pm$0.10 & 0.96$\pm$0.02 \\
ibm04 & \textbf{0.84$\pm$0.00}\,{\color{gain}\tiny($-$12.5\%)} & 1.15$\pm$0.06 & 0.96$\pm$0.02 \\
\midrule
ariane133 & \textbf{0.34$\pm$0.02}\,{\color{gain}\tiny($-$24.4\%)} & 0.46$\pm$0.00 & 0.45$\pm$0.02 \\
ariane136 & \textbf{0.34$\pm$0.02}\,{\color{gain}\tiny($-$26.1\%)} & 0.46$\pm$0.01 & 0.48$\pm$0.01 \\
\bottomrule
    \end{tabularx}
\end{table}

\paragraph{Learning-Based Placers.} This comparison isolates the value of VLM guidance as a test-time adaptation mechanism for ChiPFormer. Under the same 2,000-rollout budget, \method improves over both repeatedly sampling the fixed ChiPFormer policy (No-FT) and online fine-tuning of ChiPFormer itself (FT) on 9 of 10 benchmarks (\Cref{tab:main_results_hpwl_only}). The largest gains occur on \texttt{ibm02} (32\%) and \texttt{ariane136} (26\%), showing that VLM-guided spatial proposals can provide stronger adaptation than updating the low-level policy weights.

\subsection{Q2: Which design choices matter?}
\label{ssec:ablations}

We ablate the prompt strategy, input modality, VLM, anchor weight $\lambda_A$, context selection strategy, and context length $C$.
We also compare against two evolutionary search controls that replace the VLM.
The input modality, VLM, and search control experiments use all eight Superblue designs, and the others use \texttt{superblue1} (full plots in \Cref{fig:ablations}).

\noindent\textbf{Prompt Strategy.} We compare our default prompt against a \textbf{Greedy} variant that encourages minor refinements and an \textbf{Exploratory} variant that pushes for novel placements. On \texttt{superblue1}, the default prompt performs best, achieving $37.47 \pm 0.11$ versus $37.51 \pm 0.09$ for Greedy and $37.64 \pm 0.11$ for Exploratory (\Cref{tab:appendix_prompt_ablation}).

\noindent\textbf{Input Modality.} We remove either the chip canvas image or the textual context from each in-context example, or shuffle the images across placements, while keeping top-stratified selection and $C{=}25$ fixed. All four prompts improve over DREAMPlace on all eight Superblue designs and differ from each other by under 0.8\% (\Cref{tab:appendix_modality_ablation}), so the gain does not depend on a single input channel.

\noindent\textbf{VLM Choice.} With the open-weight Qwen3-VL 8B Instruct or Qwen3-VL 30B A3B Instruct served locally with vLLM in place of Gemini, VP+DP still improves over DREAMPlace on all eight designs and matches Gemini 3.6 Flash (\Cref{tab:open_models}).

\begin{table}[H]
\centering
\scriptsize
\setlength{\tabcolsep}{3pt}
\caption{VP+DP with open-weight VLMs on Superblue: gHPWL ($\times 10^6$, mean $\pm$ standard error over five seeds, lower is better). All runs use 2{,}000 rollouts. Bold is best and underlined is second best.
Both Qwen models are served locally with vLLM and use top-stratified history with $C{=}25$ and the global anchor weight $\lambda_A{=}0.01$.}
\label{tab:open_models}
\resizebox{\columnwidth}{!}{%
\begin{tabular}{@{}l *{8}{c}@{}}
\toprule
\textbf{VLM (VP+DP)} & \textbf{sb1} & \textbf{sb3} & \textbf{sb4} & \textbf{sb5} & \textbf{sb7} & \textbf{sb10} & \textbf{sb16} & \textbf{sb18} \\
\midrule
DP 4.3.0 & 376.63$\pm$0.13 & 453.75$\pm$0.33 & 290.92$\pm$0.26 & 400.88$\pm$0.30 & 544.45$\pm$0.35 & 714.47$\pm$1.18 & 366.34$\pm$0.12 & 219.24$\pm$0.07 \\
Gemini 3.6 Flash & \textbf{372.85$\pm$0.34} & 438.14$\pm$0.42 & 286.82$\pm$0.55 & \textbf{397.90$\pm$0.41} & \textbf{538.76$\pm$0.86} & \textbf{683.25$\pm$1.22} & \underline{364.43$\pm$0.09} & \underline{217.50$\pm$1.09} \\
Qwen3-VL 8B Instruct & \underline{374.52$\pm$0.41} & \underline{437.04$\pm$0.93} & \underline{284.03$\pm$0.75} & 400.50$\pm$0.26 & 539.85$\pm$0.38 & 686.85$\pm$0.51 & 364.46$\pm$0.38 & \textbf{217.25$\pm$0.28} \\
Qwen3-VL 30B A3B Instruct & 375.54$\pm$0.59 & \textbf{432.98$\pm$1.35} & \textbf{283.40$\pm$0.84} & \underline{400.29$\pm$0.34} & \underline{539.58$\pm$0.74} & \underline{686.25$\pm$1.26} & \textbf{364.40$\pm$0.39} & 217.59$\pm$0.32 \\
\bottomrule
\end{tabular}}
\end{table}

\noindent\textbf{Evolutionary Search Controls.} We perform ablations to test whether the improvement over DREAMPlace comes from our search loop or from the VLM.
In the first ablation, we replace the VLM's proposed regions with random regions.
In the second ablation, we replace VLM guidance by anchoring each rollout to the best placement found so far.
Both ablations run 4{,}000 rollouts, twice the 2{,}000 used by \method{}.
\method{} outperforms both ablations on all eight Superblue designs (\Cref{tab:search_controls}).

\noindent\textbf{Anchor Weight.} The anchor weight, $\lambda_A$, controls the strength of the auxiliary VLM anchor term relative to DREAMPlace's native wirelength and density terms.
As shown in \Cref{fig:ablation_anchor}, lower anchor weights perform better overall, while $\lambda_A \in \{0.1, 1.0\}$ degrades performance.
High anchor weights overconstrain macro locations and prevent standard cells from flowing freely around them. The main Superblue results use $\lambda_A{=}0.01$ on every benchmark (\Cref{appendix:dreamplace_guidance_hyperparams}).

\noindent\textbf{Context Selection Strategy.} We compare FIFO, Random, Best, Diverse, and Top Stratified strategies for selecting the $C$ in-context examples from the history buffer (definitions in \Cref{appendix:context_selection_strategies}). As shown in \Cref{fig:ablation_strategy}, each strategy continues to improve over rollouts, suggesting \method is robust to this design choice. Notably, ``Best'' performs worst, indicating that greedily selecting high-performing placements can trap the evolutionary loop in local minima.

\noindent\textbf{Invalid Suggestion Rate.} For learning-based placers, a suggested region can become invalid when it overlaps macros that were already placed earlier in the autoregressive sequence; in that case, the policy samples from its unconstrained distribution. For analytical placers, invalid suggestions correspond to malformed, out-of-bounds, or overlapping proposed macro boxes, and the corresponding anchor term is omitted. As shown in \Cref{fig:invalid_suggestions}, all strategies converge to ${\sim}20\%$ invalid suggestions.

\noindent\textbf{Context Length.} We test $C \in \{1, 10, 25\}$ examples.
As shown in \Cref{fig:ablation_context}, $C{=}25$ achieves lower gHPWL, suggesting that more in-context examples help the VLM better understand patterns associated with high-quality macro placements.

\subsection{Q3: How does \method{} perform on downstream PPA metrics?}
\label{ssec:ppa}

Because gHPWL is a placement-stage proxy, we also measure post-route power, performance, and area (PPA).
We use four designs with open Nangate45 collateral (Ariane-133, Ariane-136, BlackParrot, and BP-Quad), and push every method's macro placement through the identical OpenROAD flow up to detailed routing.
VP+DP has the shortest routed wirelength on Ariane-136, BlackParrot, and BP-Quad.
On all four designs, VP+DP also has the smallest area and the best or tied TNS and number of violating pins (\Cref{tab:ppa_main,tab:ppa_nangate45}).

\begin{table}[t]
\centering
\scriptsize
\setlength{\tabcolsep}{4pt}
\caption{Post-route PPA on four Nangate45 designs after the full OpenROAD flow.
The columns are routed wirelength (WL, $10^6\,\mu$m), worst negative slack (WNS, ns), total negative slack (TNS, $10^3$ ns), and power (mW).
\Cref{tab:ppa_nangate45} reports the full set of metrics.
VP+DP uses Gemini 3.5 Flash.
Bold is best and underlined is second best.
Error terms are standard deviations over three seeds.
An asterisk ($^{*}$) marks a mean over two seeds, because the third seed failed to route in OpenROAD.
Plain values are single runs.}
\label{tab:ppa_main}
\renewcommand{\arraystretch}{0.92}
\begin{tabularx}{\textwidth}{@{}l l *{4}{>{\centering\arraybackslash}X}@{}}
\toprule
\textbf{Method} & \textbf{Metric} & \textbf{Ariane-133} & \textbf{Ariane-136} & \textbf{BlackParrot} & \textbf{BP-Quad} \\
\midrule
\multirow{4}{*}{Hier-RTLMP} & Routed WL & 6.91 & 7.46 & 6.11 & 52.56 \\
 & WNS & -0.411 & \textbf{0.432} & -2.991 & -150.056 \\
 & TNS & -0.719 & \textbf{0.000} & \textbf{-0.003} & -38511.600 \\
 & Power & \textbf{254.3} & 162.1 & \underline{147.0} & \textbf{2008.7} \\
\midrule
\multirow{4}{*}{DP 4.3.0} & Routed WL & \textbf{5.93$\pm$0.19} & \underline{6.43$\pm$0.05} & \underline{5.71$\pm$0.02} & \underline{45.91$\pm$0.16} \\
 & WNS & \underline{-0.368$\pm$0.041} & 0.268$\pm$0.055 & -2.990$\pm$0.028 & \underline{-149.712$\pm$0.087} \\
 & TNS & \underline{-0.645$\pm$0.061} & \textbf{0.000} & \textbf{-0.003} & \underline{-38497.433$\pm$4.661} \\
 & Power & \underline{254.4$\pm$1.1} & \textbf{153.6$\pm$0.1} & 147.6$\pm$0.3 & 2018.1$\pm$6.6 \\
\midrule
\multirow{4}{*}{WireMask-EA} & Routed WL & 7.51$\pm$0.79$^{*}$ & 7.27$\pm$0.62 & 6.88$\pm$0.22 & 58.22 \\
 & WNS & -0.820$\pm$0.440$^{*}$ & 0.114$\pm$0.087 & \textbf{-2.938$\pm$0.049} & -150.507 \\
 & TNS & -2.002$\pm$0.827$^{*}$ & \textbf{0.000} & \textbf{-0.003} & -38547.600 \\
 & Power & 258.7$\pm$0.4$^{*}$ & 154.9$\pm$0.8 & 151.3$\pm$5.8 & 2058.5 \\
\midrule
\multirow{4}{*}{EGPlace} & Routed WL & 7.12$\pm$0.04$^{*}$ & 7.13$\pm$0.45$^{*}$ & 7.32$\pm$0.41 & 58.44$\pm$1.63 \\
 & WNS & -0.688$\pm$0.039$^{*}$ & -0.062$\pm$0.005$^{*}$ & -3.025$\pm$0.072 & -150.504$\pm$0.308 \\
 & TNS & -1.431$\pm$0.160$^{*}$ & \underline{-0.004$\pm$0.001}$^{*}$ & \textbf{-0.003} & -38552.300$\pm$11.772 \\
 & Power & 258.6$\pm$0.9$^{*}$ & 155.9$\pm$1.7$^{*}$ & 153.0$\pm$5.0 & 2073.1$\pm$22.3 \\
\midrule
\multirow{4}{*}{LaMPlace} & Routed WL & 7.49$\pm$1.21 & 7.59$\pm$0.51 & 7.33$\pm$0.29 & 63.92$\pm$2.14$^{*}$ \\
 & WNS & -0.496$\pm$0.171 & \underline{0.317$\pm$0.070} & -3.028$\pm$0.065 & -150.498$\pm$0.006$^{*}$ \\
 & TNS & -1.259$\pm$0.767 & \textbf{0.000} & \textbf{-0.003} & -38547.200$\pm$3.323$^{*}$ \\
 & Power & 256.9$\pm$2.2 & 160.7$\pm$2.8 & 153.3$\pm$2.7 & 2051.5$\pm$9.3$^{*}$ \\
\midrule
\multirow{4}{*}{VP+DP 3.5 Flash} & Routed WL & \underline{5.98$\pm$0.09} & \textbf{6.35$\pm$0.04} & \textbf{5.58$\pm$0.02} & \textbf{44.57$\pm$0.93} \\
 & WNS & \textbf{-0.339$\pm$0.026} & 0.277$\pm$0.121 & \underline{-2.965$\pm$0.014} & \textbf{-149.585$\pm$0.029} \\
 & TNS & \textbf{-0.549$\pm$0.039} & \textbf{0.000} & \textbf{-0.003} & \textbf{-38492.100$\pm$5.450} \\
 & Power & 255.3$\pm$0.5 & \underline{153.7$\pm$0.5} & \textbf{146.4$\pm$0.5} & \underline{2012.5$\pm$7.8} \\
\bottomrule
\end{tabularx}
\end{table}

\subsection{Q4: What runtime and API-cost overhead does VLM guidance introduce?}
\label{ssec:runtime_overhead}

Any iterative macro-placement method must repeatedly invoke a placer to position standard cells around macros and evaluate candidate layouts, so the additional cost introduced by \method comes from VLM inference. In both experimental tracks, each 2,000-rollout run uses 125 Gemini calls: one query per batch of eight guided rollouts, paired with eight unguided rollouts. Gemini 3.6 Flash averages 62 seconds per call, for about 2.1 hours of benchmark-independent latency per run. \Cref{tab:runtime_overhead} reports the resulting wall-clock overhead for one run on a single A100 GPU: 2\%--7\% over DREAMPlace, so DREAMPlace accounts for 93\%--98\% of a run. \method requires no VLM training, and under conservative token-count estimates, the API-cost overhead is about \$7 per 2,000-rollout experiment (see \Cref{app:api_cost}).

\section{Limitations \& Future Work}
\label{sec:limitations}
The search loop optimizes only gHPWL.
We evaluate post-route PPA only after the search, so the loop does not use PPA metrics.
This evaluation covers four Nangate45 designs (\Cref{tab:ppa_main}).
The full physical-design flow is too slow to run inside the loop (\Cref{appendix:wirelength_ppa}).
\method{} depends on VLM calls, which add the latency and API cost reported in \Cref{ssec:runtime_overhead}.
\method{} also runs with open-weight Qwen3-VL models served locally (\Cref{tab:open_models}).
The VLM guides at most 256 macros per design.
High-quality chips ultimately require optimizing additional PPA metrics such as timing and power.
Because \method{} relies on natural language prompting rather than a fixed loss function, these metrics can be incorporated directly into the VLM prompt and evolutionary scoring loop when available.
Future work will add more macros to the prompt and extend \method{} to multi-objective PPA optimization.

\section{Conclusion}
\label{sec:conclusion}
We proposed \method{}, an evolutionary framework that leverages vision-language models to enhance chip floorplanning. Through a structured prompting approach that requires no fine-tuning of the VLM, \method{} improves strong learning-based and analytical placers, providing a blueprint for integrating multimodal foundation models into complex engineering workflows as high-level strategic guides for computer-aided design.

\section*{Acknowledgments}
This work has been made possible in part by a gift from the Chan Zuckerberg Initiative Foundation to establish the Kempner Institute for the Study of Natural and Artificial Intelligence at Harvard University. Part of the computations in this paper were run on the FASRC Cannon cluster supported by the FAS Division of Science Research Computing Group at Harvard University.

\bibliographystyle{plainnat}
\bibliography{references}

\begin{thebibliography}{48}
\providecommand{\natexlab}[1]{#1}
\providecommand{\url}[1]{\texttt{#1}}
\expandafter\ifx\csname urlstyle\endcsname\relax
  \providecommand{\doi}[1]{doi: #1}\else
  \providecommand{\doi}{doi: \begingroup \urlstyle{rm}\Url}\fi

\bibitem[Adya and Markov(2002)]{adya2002consistent_iccad04}
Saurabh~N Adya and Igor~L Markov.
\newblock Consistent placement of macro-blocks using floorplanning and
  standard-cell placement.
\newblock In \emph{Proceedings of the 2002 international symposium on Physical
  design}, pages 12--17, 2002.

\bibitem[Adya et~al.(2004)Adya, Chaturvedi, Roy, Papa, and
  Markov]{adya2004unification_iccad04}
Saurabh~N Adya, Shubhyant Chaturvedi, Jarrod~A Roy, David~A Papa, and Igor~L
  Markov.
\newblock Unification of partitioning, placement and floorplanning.
\newblock In \emph{IEEE/ACM International Conference on Computer Aided Design,
  2004. ICCAD-2004.}, pages 550--557. IEEE, 2004.

\bibitem[Ahn et~al.(2022)Ahn, Brohan, Brown, Chebotar, Cortes, David, Finn, Fu,
  Gopalakrishnan, Hausman, et~al.]{say_can}
Michael Ahn, Anthony Brohan, Noah Brown, Yevgen Chebotar, Omar Cortes, Byron
  David, Chelsea Finn, Chuyuan Fu, Keerthana Gopalakrishnan, Karol Hausman,
  et~al.
\newblock Do as i can, not as i say: Grounding language in robotic affordances.
\newblock \emph{arXiv preprint arXiv:2204.01691}, 2022.

\bibitem[Brohan et~al.(2023)Brohan, Brown, Carbajal, Chebotar, Chen,
  Choromanski, Ding, Driess, Dubey, Finn, et~al.]{brohan2023rt}
Anthony Brohan, Noah Brown, Justice Carbajal, Yevgen Chebotar, Xi~Chen,
  Krzysztof Choromanski, Tianli Ding, Danny Driess, Avinava Dubey, Chelsea
  Finn, et~al.
\newblock Rt-2: Vision-language-action models transfer web knowledge to robotic
  control.
\newblock \emph{arXiv preprint arXiv:2307.15818}, 2023.

\bibitem[Cant{\'u}-Paz et~al.(1998)]{cantu1998survey}
Erick Cant{\'u}-Paz et~al.
\newblock A survey of parallel genetic algorithms.
\newblock \emph{Calculateurs paralleles, reseaux et systems repartis},
  10\penalty0 (2):\penalty0 141--171, 1998.

\bibitem[Chen et~al.(2021)Chen, Lu, Rajeswaran, Lee, Grover, Laskin, Abbeel,
  Srinivas, and Mordatch]{chen2021decision}
Lili Chen, Kevin Lu, Aravind Rajeswaran, Kimin Lee, Aditya Grover, Misha
  Laskin, Pieter Abbeel, Aravind Srinivas, and Igor Mordatch.
\newblock Decision transformer: Reinforcement learning via sequence modeling.
\newblock \emph{Advances in neural information processing systems},
  34:\penalty0 15084--15097, 2021.

\bibitem[Chen et~al.(2008)Chen, Jiang, Hsu, Chen, and Chang]{chen2008ntuplace3}
Tung-Chieh Chen, Zhe-Wei Jiang, Tien-Chang Hsu, Hsin-Chen Chen, and Yao-Wen
  Chang.
\newblock Ntuplace3: An analytical placer for large-scale mixed-size designs
  with preplaced blocks and density constraints.
\newblock \emph{IEEE Transactions on Computer-Aided Design of Integrated
  Circuits and Systems}, 27\penalty0 (7):\penalty0 1228--1240, 2008.

\bibitem[Cheng et~al.(2018)Cheng, Kahng, Kang, and Wang]{cheng2018replace}
Chung-Kuan Cheng, Andrew~B Kahng, Ilgweon Kang, and Lutong Wang.
\newblock Replace: Advancing solution quality and routability validation in
  global placement.
\newblock \emph{IEEE Transactions on Computer-Aided Design of Integrated
  Circuits and Systems}, 38\penalty0 (9):\penalty0 1717--1730, 2018.

\bibitem[Cheng and Yan(2021)]{cheng2021joint_deep_pr}
Ruoyu Cheng and Junchi Yan.
\newblock On joint learning for solving placement and routing in chip design.
\newblock \emph{Advances in Neural Information Processing Systems},
  34:\penalty0 16508--16519, 2021.

\bibitem[Driess et~al.(2023)Driess, Xia, Sajjadi, Lynch, Chowdhery, Ichter,
  Wahid, Tompson, Vuong, Yu, Huang, Chebotar, Sermanet, Duckworth, Levine,
  Vanhoucke, Hausman, Toussaint, Greff, Zeng, Mordatch, and
  Florence]{driess2023palm_e}
Danny Driess, Fei Xia, Mehdi S.~M. Sajjadi, Corey Lynch, Aakanksha Chowdhery,
  Brian Ichter, Ayzaan Wahid, Jonathan Tompson, Quan Vuong, Tianhe Yu, Wenlong
  Huang, Yevgen Chebotar, Pierre Sermanet, Daniel Duckworth, Sergey Levine,
  Vincent Vanhoucke, Karol Hausman, Marc Toussaint, Klaus Greff, Andy Zeng,
  Igor Mordatch, and Pete Florence.
\newblock Palm-e: an embodied multimodal language model.
\newblock In \emph{Proceedings of the 40th International Conference on Machine
  Learning}, ICML'23. JMLR.org, 2023.

\bibitem[Geng et~al.(2024)Geng, Wang, Liu, Xu, Tang, Yuan, Hao, Zhang, and
  Wu]{rl_within_tree_search_for_macro_placement}
Zijie Geng, Jie Wang, Ziyan Liu, Siyuan Xu, Zhentao Tang, Mingxuan Yuan, Jianye
  Hao, Yongdong Zhang, and Feng Wu.
\newblock Reinforcement learning within tree search for fast macro placement.
\newblock In \emph{Forty-first International Conference on Machine Learning},
  2024.

\bibitem[Gottweis et~al.(2025)Gottweis, Weng, Daryin, Tu, Palepu, Sirkovic,
  Myaskovsky, Weissenberger, Rong, Tanno, et~al.]{gottweis2025towards}
Juraj Gottweis, Wei-Hung Weng, Alexander Daryin, Tao Tu, Anil Palepu, Petar
  Sirkovic, Artiom Myaskovsky, Felix Weissenberger, Keran Rong, Ryutaro Tanno,
  et~al.
\newblock Towards an ai co-scientist.
\newblock \emph{arXiv preprint arXiv:2502.18864}, 2025.

\bibitem[Hagberg et~al.(2008)Hagberg, Swart, and Schult]{network_x}
Aric Hagberg, Pieter~J Swart, and Daniel~A Schult.
\newblock Exploring network structure, dynamics, and function using networkx.
\newblock Technical report, Los Alamos National Laboratory (LANL), Los Alamos,
  NM (United States), 2008.

\bibitem[Hemberg et~al.(2024)Hemberg, Moskal, and
  O’Reilly]{hemberg2024evolving}
Erik Hemberg, Stephen Moskal, and Una-May O’Reilly.
\newblock Evolving code with a large language model.
\newblock \emph{Genetic Programming and Evolvable Machines}, 25\penalty0
  (2):\penalty0 21, 2024.

\bibitem[Huang et~al.(2022)Huang, Xia, Xiao, Chan, Liang, Florence, Zeng,
  Tompson, Mordatch, Chebotar, et~al.]{huang2022inner_monologue}
Wenlong Huang, Fei Xia, Ted Xiao, Harris Chan, Jacky Liang, Pete Florence, Andy
  Zeng, Jonathan Tompson, Igor Mordatch, Yevgen Chebotar, et~al.
\newblock Inner monologue: Embodied reasoning through planning with language
  models.
\newblock \emph{arXiv preprint arXiv:2207.05608}, 2022.

\bibitem[Jiang et~al.(2022)Jiang, Gupta, Zhang, Wang, Dou, Chen, Fei-Fei,
  Anandkumar, Zhu, and Fan]{jiang2022vima}
Yunfan Jiang, Agrim Gupta, Zichen Zhang, Guanzhi Wang, Yongqiang Dou, Yanjun
  Chen, Li~Fei-Fei, Anima Anandkumar, Yuke Zhu, and Linxi Fan.
\newblock Vima: General robot manipulation with multimodal prompts.
\newblock \emph{arXiv preprint arXiv:2210.03094}, 2\penalty0 (3):\penalty0 6,
  2022.

\bibitem[Kim et~al.(2024)Kim, Pertsch, Karamcheti, Xiao, Balakrishna, Nair,
  Rafailov, Foster, Lam, Sanketi, et~al.]{kim2024openvla}
Moo~Jin Kim, Karl Pertsch, Siddharth Karamcheti, Ted Xiao, Ashwin Balakrishna,
  Suraj Nair, Rafael Rafailov, Ethan Foster, Grace Lam, Pannag Sanketi, et~al.
\newblock Openvla: An open-source vision-language-action model.
\newblock \emph{arXiv preprint arXiv:2406.09246}, 2024.

\bibitem[Kim et~al.(2015)Kim, Hu, Li, and Viswanathan]{kim2015iccad}
Myung-Chul Kim, Jin Hu, Jiajia Li, and Natarajan Viswanathan.
\newblock Iccad-2015 cad contest in incremental timing-driven placement and
  benchmark suite.
\newblock In \emph{2015 IEEE/ACM International Conference on Computer-Aided
  Design (ICCAD)}, pages 921--926. IEEE, 2015.

\bibitem[Lai et~al.(2022)Lai, Mu, and Luo]{lai2022maskplace}
Yao Lai, Yao Mu, and Ping Luo.
\newblock Maskplace: Fast chip placement via reinforced visual representation
  learning.
\newblock \emph{Advances in Neural Information Processing Systems},
  35:\penalty0 24019--24030, 2022.

\bibitem[Lai et~al.(2023)Lai, Liu, Tang, Wang, Hao, and Luo]{chipformer}
Yao Lai, Jinxin Liu, Zhentao Tang, Bin Wang, Jianye Hao, and Ping Luo.
\newblock Chipformer: Transferable chip placement via offline decision
  transformer.
\newblock In \emph{International Conference on Machine Learning}, pages
  18346--18364. PMLR, 2023.

\bibitem[Lee et~al.(2022)Lee, Nachum, Yang, Lee, Freeman, Guadarrama, Fischer,
  Xu, Jang, Michalewski, et~al.]{lee2022multi}
Kuang-Huei Lee, Ofir Nachum, Mengjiao~Sherry Yang, Lisa Lee, Daniel Freeman,
  Sergio Guadarrama, Ian Fischer, Winnie Xu, Eric Jang, Henryk Michalewski,
  et~al.
\newblock Multi-game decision transformers.
\newblock \emph{Advances in Neural Information Processing Systems},
  35:\penalty0 27921--27936, 2022.

\bibitem[Lee et~al.(2025)Lee, Fischer, Wu, Marwood, Baluja, Schuurmans, and
  Chen]{lee2025evolving}
Kuang-Huei Lee, Ian Fischer, Yueh-Hua Wu, Dave Marwood, Shumeet Baluja, Dale
  Schuurmans, and Xinyun Chen.
\newblock Evolving deeper llm thinking.
\newblock \emph{arXiv preprint arXiv:2501.09891}, 2025.

\bibitem[Lee et~al.(2024)Lee, Deng, Elzeiny, Abbeel, and
  Wawrzynek]{chip_placement_with_diffusion}
Vint Lee, Chun Deng, Leena Elzeiny, Pieter Abbeel, and John Wawrzynek.
\newblock Chip placement with diffusion.
\newblock \emph{arXiv preprint arXiv:2407.12282}, 2024.

\bibitem[Liang et~al.(2024)Liang, Xia, Yu, Zeng, Arenas, Attarian, Bauza,
  Bennice, Bewley, Dostmohamed, et~al.]{liang2024learning}
Jacky Liang, Fei Xia, Wenhao Yu, Andy Zeng, Montserrat~Gonzalez Arenas, Maria
  Attarian, Maria Bauza, Matthew Bennice, Alex Bewley, Adil Dostmohamed, et~al.
\newblock Learning to learn faster from human feedback with language model
  predictive control.
\newblock \emph{arXiv preprint arXiv:2402.11450}, 2024.

\bibitem[Lin et~al.(2019)Lin, Dhar, Li, Ren, Khailany, and
  Pan]{lin_dreamplace_2019}
Yibo Lin, Shounak Dhar, Wuxi Li, Haoxing Ren, Brucek Khailany, and David~Z.
  Pan.
\newblock {DREAMPlace}: {Deep} {Learning} {Toolkit}-{Enabled} {GPU}
  {Acceleration} for {Modern} {VLSI} {Placement}.
\newblock In \emph{Proceedings of the 56th {Annual} {Design} {Automation}
  {Conference} 2019}, pages 1--6, Las Vegas NV USA, June 2019. ACM.
\newblock ISBN 978-1-4503-6725-7.
\newblock \doi{10.1145/3316781.3317803}.
\newblock URL \url{https://dl.acm.org/doi/10.1145/3316781.3317803}.

\bibitem[Liventsev et~al.(2023)Liventsev, Grishina, H{\"a}rm{\"a}, and
  Moonen]{liventsev2023fully}
Vadim Liventsev, Anastasiia Grishina, Aki H{\"a}rm{\"a}, and Leon Moonen.
\newblock Fully autonomous programming with large language models.
\newblock In \emph{Proceedings of the Genetic and Evolutionary Computation
  Conference}, pages 1146--1155, 2023.

\bibitem[Lu et~al.(2015)Lu, Chen, Chang, Sha, Huang, Teng, and
  Cheng]{lu2015eplace}
Jingwei Lu, Pengwen Chen, Chin-Chih Chang, Lu~Sha, Dennis Jen-Hsin Huang,
  Chin-Chi Teng, and Chung-Kuan Cheng.
\newblock eplace: Electrostatics-based placement using fast fourier transform
  and nesterov's method.
\newblock \emph{ACM Transactions on Design Automation of Electronic Systems
  (TODAES)}, 20\penalty0 (2):\penalty0 1--34, 2015.

\bibitem[Mirhoseini et~al.(2021)Mirhoseini, Goldie, Yazgan, Jiang, Songhori,
  Wang, Lee, Johnson, Pathak, and Nova]{circuit_training_paper}
Azalia Mirhoseini, Anna Goldie, Mustafa Yazgan, Joe~Wenjie Jiang, Ebrahim
  Songhori, Shen Wang, Young-Joon Lee, Eric Johnson, Omkar Pathak, and Azade
  Nova.
\newblock A graph placement methodology for fast chip design.
\newblock \emph{Nature}, 594\penalty0 (7862):\penalty0 207--212, 2021.

\bibitem[Murata et~al.(1996)Murata, Fujiyoshi, Nakatake, and
  Kajitani]{murata2002vlsi}
Hiroshi Murata, Kunihiro Fujiyoshi, Shigetoshi Nakatake, and Yoji Kajitani.
\newblock Vlsi module placement based on rectangle-packing by the
  sequence-pair.
\newblock \emph{IEEE Transactions on Computer-Aided Design of Integrated
  Circuits and Systems}, 15\penalty0 (12):\penalty0 1518--1524, 1996.
\newblock \doi{10.1109/43.552084}.

\bibitem[Nam et~al.(2005)Nam, Alpert, Villarrubia, Winter, and
  Yildiz]{ispd2005_benchmarks}
Gi-Joon Nam, Charles~J Alpert, Paul Villarrubia, Bruce Winter, and Mehmet
  Yildiz.
\newblock The ispd2005 placement contest and benchmark suite.
\newblock In \emph{Proceedings of the 2005 international symposium on Physical
  design}, pages 216--220, 2005.

\bibitem[Nasiriany et~al.(2024)Nasiriany, Xia, Yu, Xiao, Liang, Dasgupta, Xie,
  Driess, Wahid, Xu, et~al.]{nasiriany2024pivot}
Soroush Nasiriany, Fei Xia, Wenhao Yu, Ted Xiao, Jacky Liang, Ishita Dasgupta,
  Annie Xie, Danny Driess, Ayzaan Wahid, Zhuo Xu, et~al.
\newblock Pivot: Iterative visual prompting elicits actionable knowledge for
  vlms.
\newblock In \emph{International Conference on Machine Learning}, pages
  37321--37341. PMLR, 2024.

\bibitem[Novikov et~al.(2025)Novikov, Vu, Eisenberger, Dupont, Huang, Wagner,
  Shirobokov, Kozlovskii, Ruiz, Mehrabian, Kumar, See, Chaudhuri, Holland,
  Davies, Nowozin, Kohli, and Balog]{alphaevolve}
Alexander Novikov, Ngân Vu, Marvin Eisenberger, Emilien Dupont, Po-Sen Huang,
  Adam~Zsolt Wagner, Sergey Shirobokov, Borislav Kozlovskii, Francisco J.~R.
  Ruiz, Abbas Mehrabian, M.~Pawan Kumar, Abigail See, Swarat Chaudhuri, George
  Holland, Alex Davies, Sebastian Nowozin, Pushmeet Kohli, and Matej Balog.
\newblock Alpha{E}volve: A coding agent for scientific and algorithmic
  discovery.
\newblock Technical report, Google Deepmind, 2025.

\bibitem[{OpenHW Group}(2024)]{ariane_cva6}
{OpenHW Group}.
\newblock {CVA6 RISC-V CPU}.
\newblock \url{https://github.com/openhwgroup/cva6}, 2024.

\bibitem[Rabaey et~al.(2003)Rabaey, Chandrakasan, and
  Nikolic]{rabaey2003digital}
Jan~M. Rabaey, Anantha Chandrakasan, and Borivoje Nikolic.
\newblock \emph{Digital Integrated Circuits: A Design Perspective}.
\newblock Prentice Hall, 2 edition, 2003.

\bibitem[Romera-Paredes et~al.(2024)Romera-Paredes, Barekatain, Novikov, Balog,
  Kumar, Dupont, Ruiz, Ellenberg, Wang, Fawzi, et~al.]{romera2024mathematical}
Bernardino Romera-Paredes, Mohammadamin Barekatain, Alexander Novikov, Matej
  Balog, M~Pawan Kumar, Emilien Dupont, Francisco~JR Ruiz, Jordan~S Ellenberg,
  Pengming Wang, Omar Fawzi, et~al.
\newblock Mathematical discoveries from program search with large language
  models.
\newblock \emph{Nature}, 625\penalty0 (7995):\penalty0 468--475, 2024.

\bibitem[Rousseeuw(1987)]{rousseeuw1987silhouettes}
Peter~J Rousseeuw.
\newblock Silhouettes: a graphical aid to the interpretation and validation of
  cluster analysis.
\newblock \emph{Journal of computational and applied mathematics}, 20:\penalty0
  53--65, 1987.

\bibitem[Shi et~al.(2023)Shi, Xue, Lei, and Qian]{wiremask}
Yunqi Shi, Ke~Xue, Song Lei, and Chao Qian.
\newblock Macro placement by wire-mask-guided black-box optimization.
\newblock \emph{Advances in Neural Information Processing Systems},
  36:\penalty0 6825--6843, 2023.

\bibitem[Shridhar et~al.(2022)Shridhar, Manuelli, and Fox]{shridhar2022cliport}
Mohit Shridhar, Lucas Manuelli, and Dieter Fox.
\newblock Cliport: What and where pathways for robotic manipulation.
\newblock In \emph{Conference on robot learning}, pages 894--906. PMLR, 2022.

\bibitem[Singha et~al.(2012)Singha, Dutta, and
  De]{opt_of_floorplanning_using_GA}
T~Singha, HS~Dutta, and M~De.
\newblock Optimization of floor-planning using genetic algorithm.
\newblock \emph{Procedia Technology}, 4:\penalty0 825--829, 2012.

\bibitem[Tanese(1989)]{tanese1989distributed}
Reiko Tanese.
\newblock \emph{Distributed genetic algorithms for function optimization}.
\newblock University of Michigan, 1989.

\bibitem[Team et~al.(2023)Team, Anil, Borgeaud, Alayrac, Yu, Soricut,
  Schalkwyk, Dai, Hauth, Millican, et~al.]{team2023gemini}
Gemini Team, Rohan Anil, Sebastian Borgeaud, Jean-Baptiste Alayrac, Jiahui Yu,
  Radu Soricut, Johan Schalkwyk, Andrew~M Dai, Anja Hauth, Katie Millican,
  et~al.
\newblock Gemini: a family of highly capable multimodal models.
\newblock \emph{arXiv preprint arXiv:2312.11805}, 2023.

\bibitem[Team et~al.(2025)Team, Abeyruwan, Ainslie, Alayrac, Arenas, Armstrong,
  Balakrishna, Baruch, Bauza, Blokzijl, et~al.]{team2025gemini}
Gemini~Robotics Team, Saminda Abeyruwan, Joshua Ainslie, Jean-Baptiste Alayrac,
  Montserrat~Gonzalez Arenas, Travis Armstrong, Ashwin Balakrishna, Robert
  Baruch, Maria Bauza, Michiel Blokzijl, et~al.
\newblock Gemini robotics: Bringing ai into the physical world.
\newblock \emph{arXiv preprint arXiv:2503.20020}, 2025.

\bibitem[Wei et~al.(2024)Wei, Zhuang, Soedarmadji, and Sui]{wei2024scalable}
Yunyue Wei, Vincent Zhuang, Saraswati Soedarmadji, and Yanan Sui.
\newblock Scalable bayesian optimization via focalized sparse gaussian
  processes.
\newblock \emph{arXiv preprint arXiv:2412.20375}, 2024.

\bibitem[Weste and Harris(2011)]{weste2011cmos}
Neil H.~E. Weste and David~Money Harris.
\newblock \emph{CMOS VLSI Design: A Circuits and Systems Perspective}.
\newblock Addison-Wesley, 4 edition, 2011.

\bibitem[Wong and Liu(1986)]{new_algo_for_floorplanning_SA}
DF~Wong and CL~Liu.
\newblock A new algorithm for floorplan design.
\newblock In \emph{23rd ACM/IEEE Design Automation Conference}, pages 101--107.
  IEEE, 1986.

\bibitem[Xue et~al.(2024)Xue, Chen, Lin, Shi, Kai, Xu, and
  Qian]{macro_regulator}
Ke~Xue, Ruo-Tong Chen, Xi~Lin, Yunqi Shi, Shixiong Kai, Siyuan Xu, and Chao
  Qian.
\newblock Reinforcement learning policy as macro regulator rather than macro
  placer.
\newblock \emph{Advances in Neural Information Processing Systems},
  37:\penalty0 140565--140588, 2024.

\bibitem[Yamada et~al.(2025)Yamada, Lange, Lu, Hu, Lu, Foerster, Clune, and
  Ha]{aiscientist_v2}
Yutaro Yamada, Robert~Tjarko Lange, Cong Lu, Shengran Hu, Chris Lu, Jakob
  Foerster, Jeff Clune, and David Ha.
\newblock The ai scientist-v2: Workshop-level automated scientific discovery
  via agentic tree search.
\newblock \emph{arXiv preprint arXiv:2504.08066}, 2025.

\bibitem[Yao et~al.(2025)Yao, Jiang, Zhao, Liao, Lin, and
  Yu]{yao2025evolution_with_llm}
Xufeng Yao, Jiaxi Jiang, Yuxuan Zhao, Peiyu Liao, Yibo Lin, and Bei Yu.
\newblock Evolution of optimization algorithms for global placement via large
  language models.
\newblock \emph{arXiv preprint arXiv:2504.17801}, 2025.

\end{thebibliography}

\appendix
\crefalias{section}{appendix}
\crefalias{subsection}{appendix}
\crefalias{subsubsection}{appendix}
\raggedbottom
\section*{Appendix}

\section{Additional Details}
\label{appendix:additional}

\subsection{Benchmark Statistics}
\label{appendix:benchmark_statistics}

\begin{table}[H]
\centering
\small
\caption{Benchmark statistics for all evaluated circuits. IBM benchmarks do not distinguish macros from standard cells; their macro counts come from DREAMPlace, which treats a movable object as a macro when its area exceeds 10 times the mean movable-cell area and its height spans more than two standard-cell rows.}
\label{tab:benchmark_statistics}
\begin{tabular}{@{}lrrrr@{}}
\toprule
\textbf{Benchmark} & \textbf{\# Objects} & \textbf{\# Macros} & \textbf{\# Nets} & \textbf{\# Pins} \\
\midrule
\multicolumn{5}{l}{\textit{ISPD 2005}} \\
adaptec1 & 211K & 63 & 221K & 944K \\
adaptec2 & 255K & 127 & 266K & 1.1M \\
adaptec3 & 452K & 58 & 467K & 1.9M \\
adaptec4 & 496K & 69 & 516K & 1.9M \\
\midrule
\multicolumn{5}{l}{\textit{ICCAD 2004}} \\
ibm01 & 13K & 246 & 14K & 51K \\
ibm02 & 20K & 271 & 20K & 81K \\
ibm03 & 23K & 290 & 27K & 94K \\
ibm04 & 28K & 295 & 32K & 106K \\
\midrule
\multicolumn{5}{l}{\textit{Ariane (Nangate45)}} \\
ariane133 & --- & 133 & --- & --- \\
ariane136 & --- & 136 & --- & --- \\
\midrule
\multicolumn{5}{l}{\textit{ICCAD 2015 Superblue}} \\
superblue1 & 1,215K & 424 & 1,216K & 3.8M \\
superblue3 & 1,219K & 565 & 1,225K & 3.9M \\
superblue4 & 802K & 300 & 803K & 2.5M \\
superblue5 & 1,090K & 770 & 1,101K & 3.2M \\
superblue7 & 1,938K & 441 & 1,934K & 6.4M \\
superblue10 & 984K & 1,629 & 1,898K & 5.6M \\
superblue16 & 986K & 99 & 1,000K & 3.0M \\
superblue18 & 772K & 201 & 772K & 2.6M \\
\bottomrule
\end{tabular}
\end{table}

\subsection{Visual Examples}
\label{appendix:visual_examples}

\subsubsection{Analytical Placer Comparison}
\begin{figure}[H]
    \centering
    \setlength{\tabcolsep}{4pt}
    \begin{tabular}{ccc}
        \small superblue1 & \small superblue4 & \small superblue16 \\
        \includegraphics[height=0.14\textwidth]{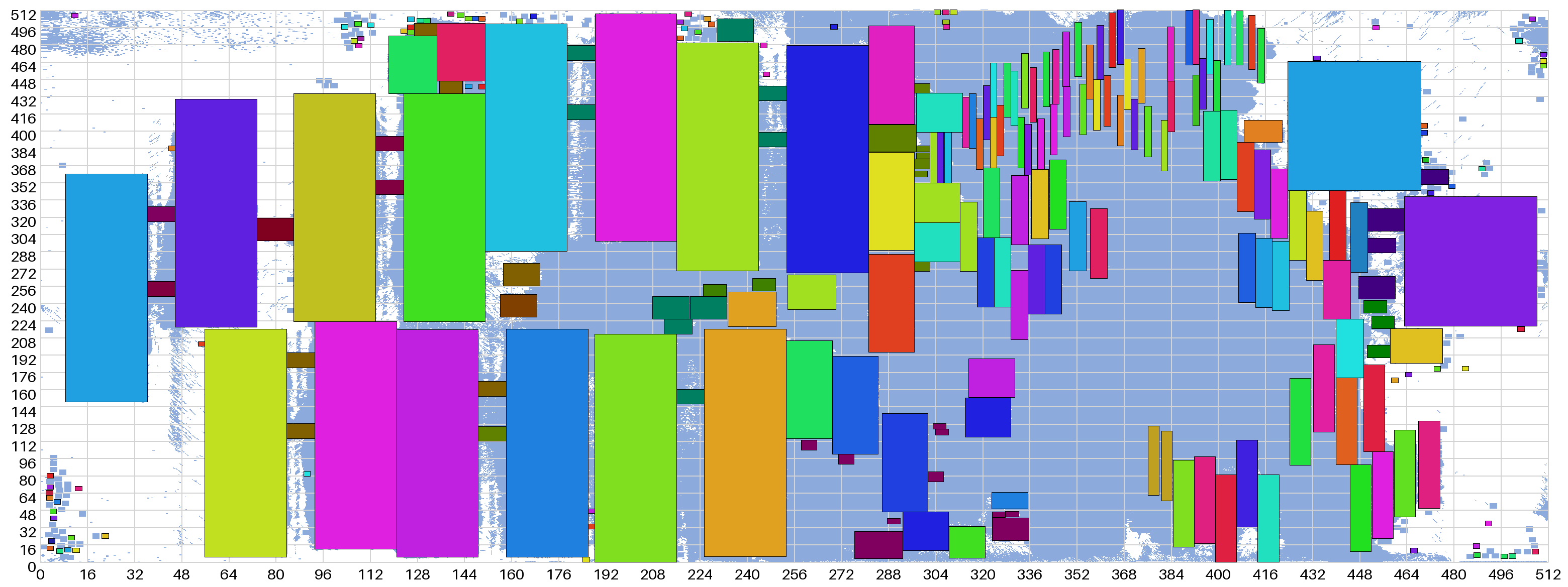} &
        \includegraphics[height=0.14\textwidth]{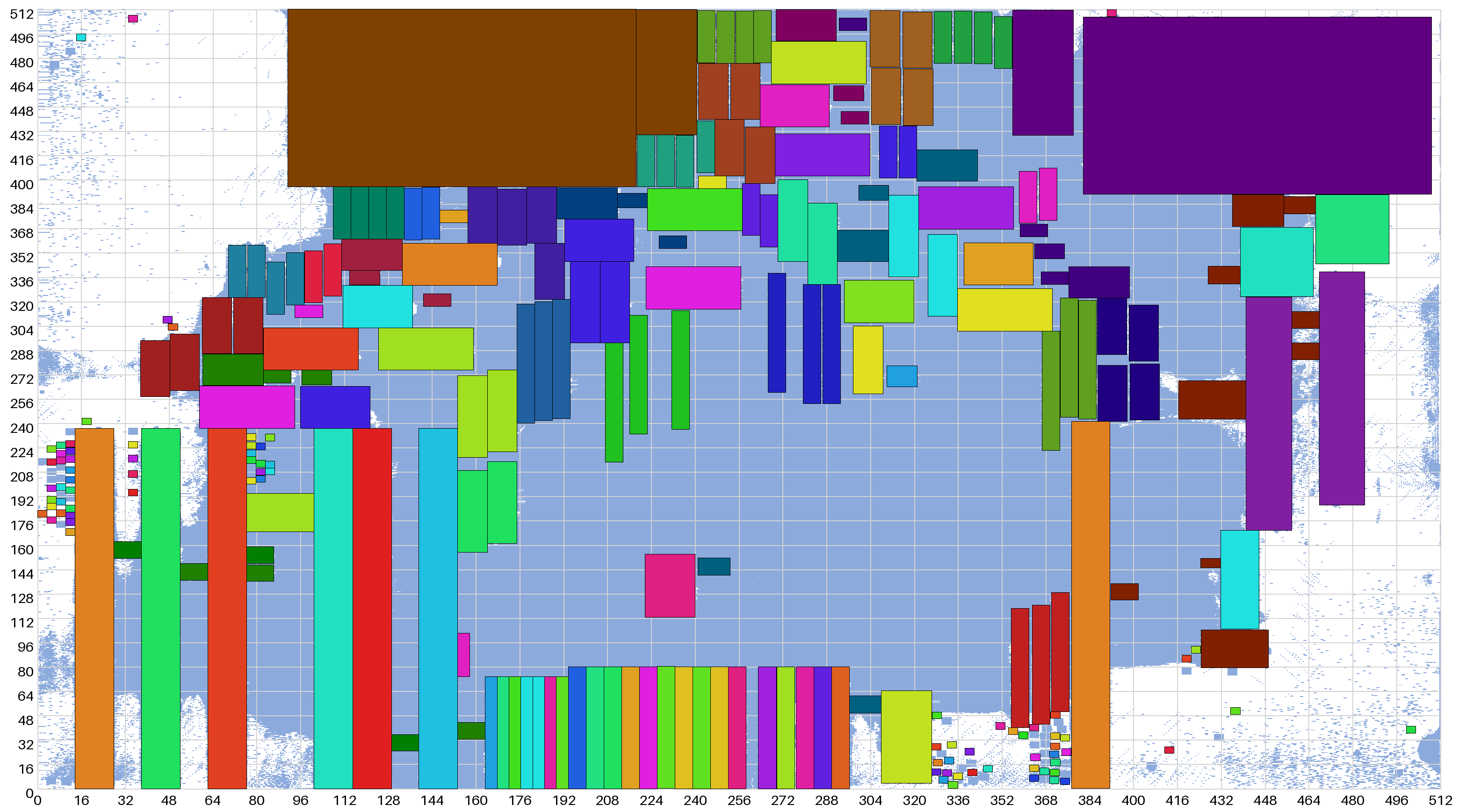} &
        \includegraphics[height=0.14\textwidth]{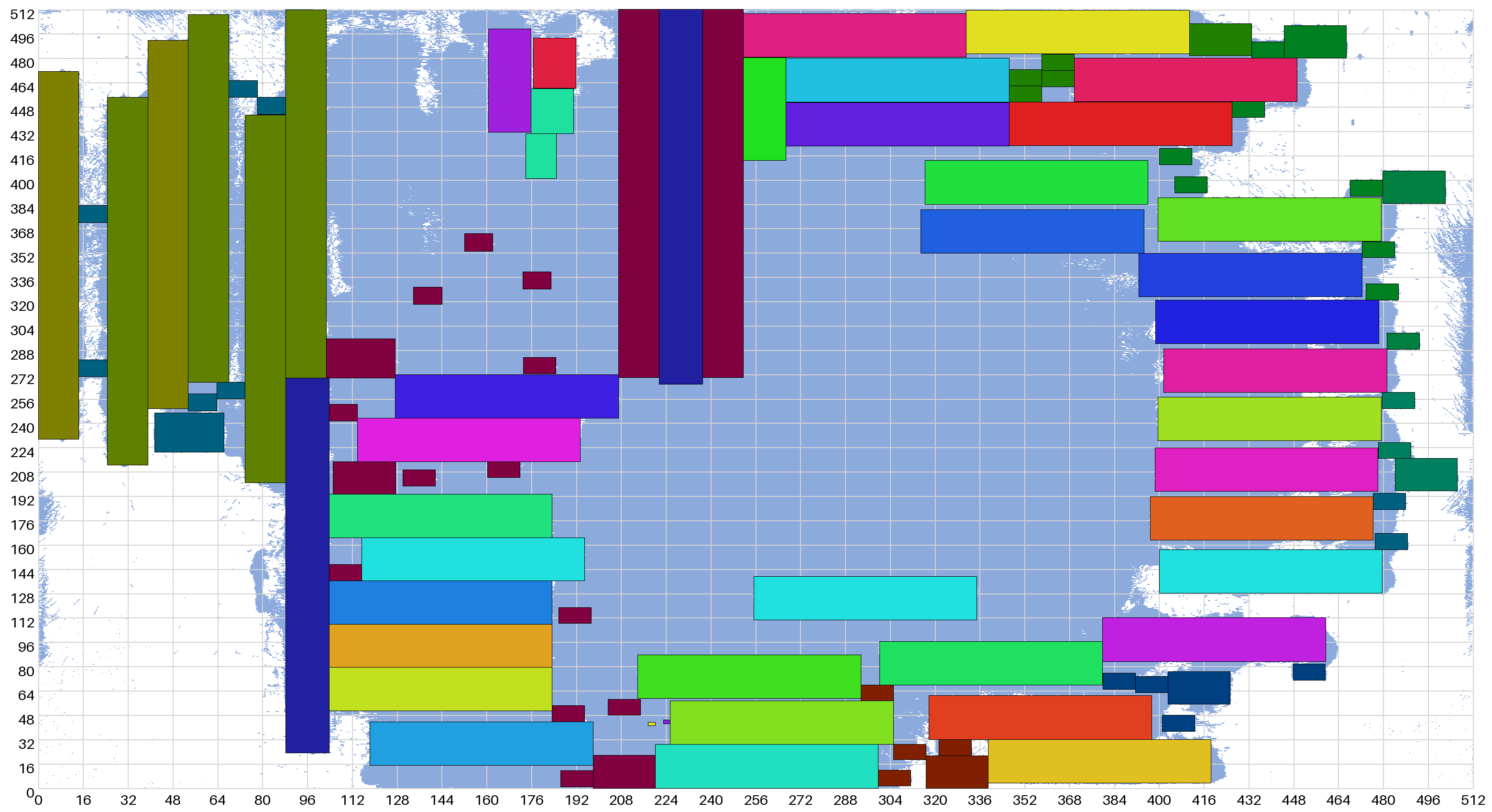} \\
        \includegraphics[height=0.14\textwidth]{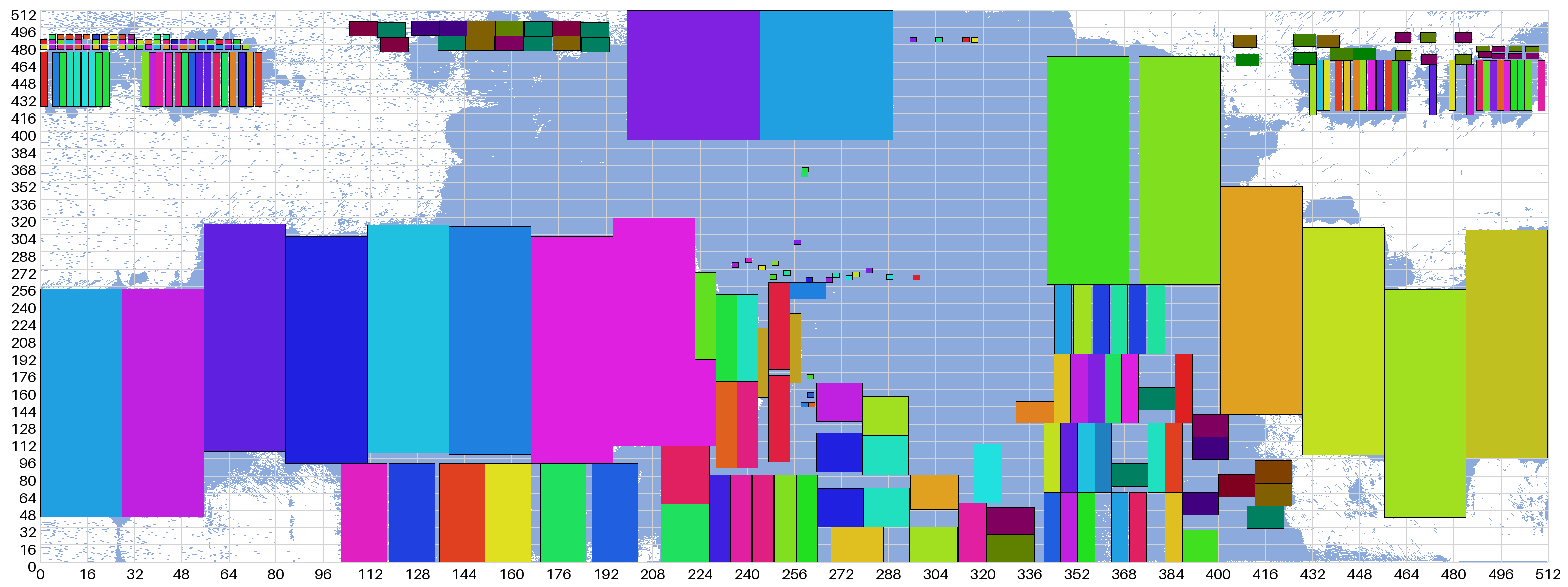} &
        \includegraphics[height=0.14\textwidth]{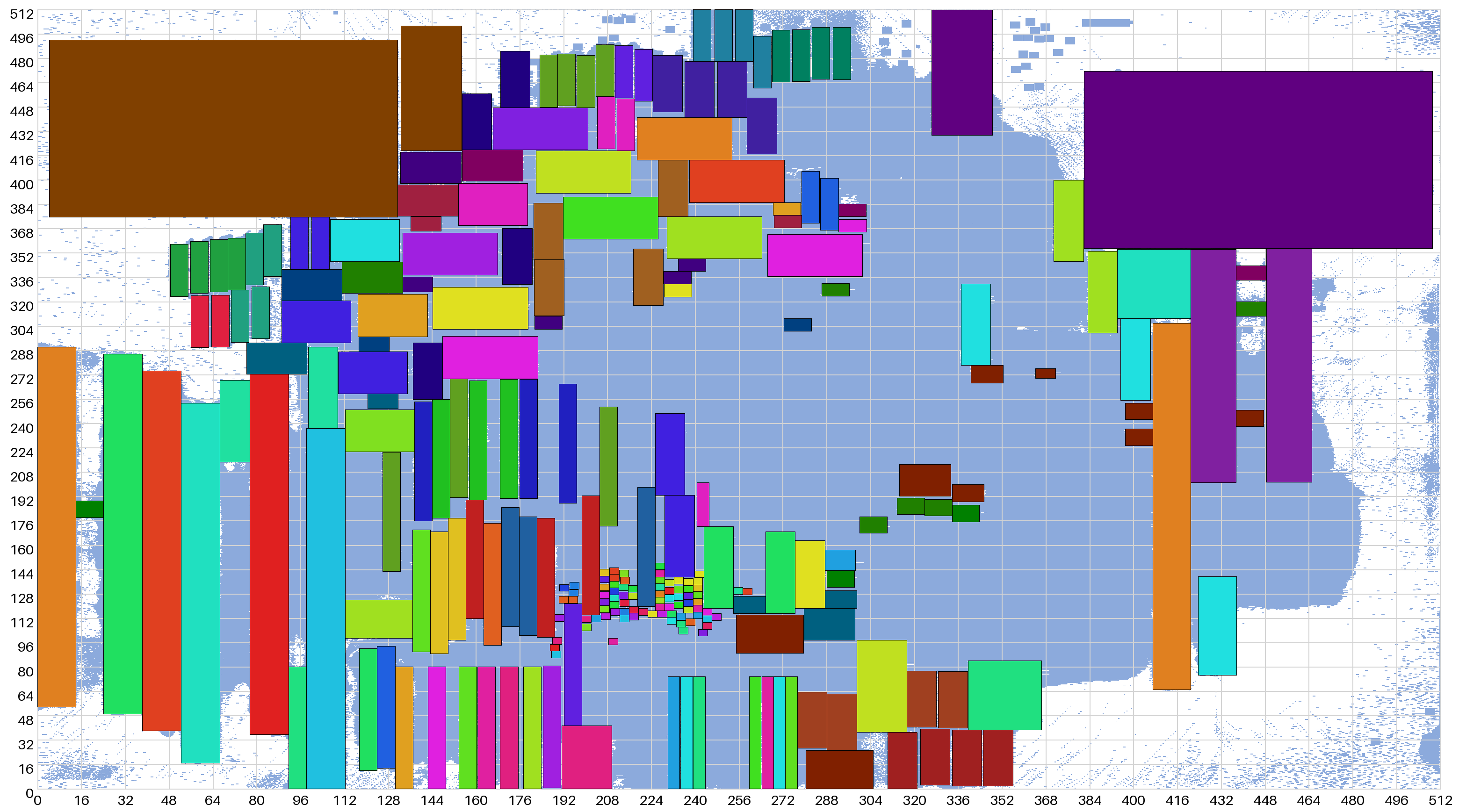} &
        \includegraphics[height=0.14\textwidth]{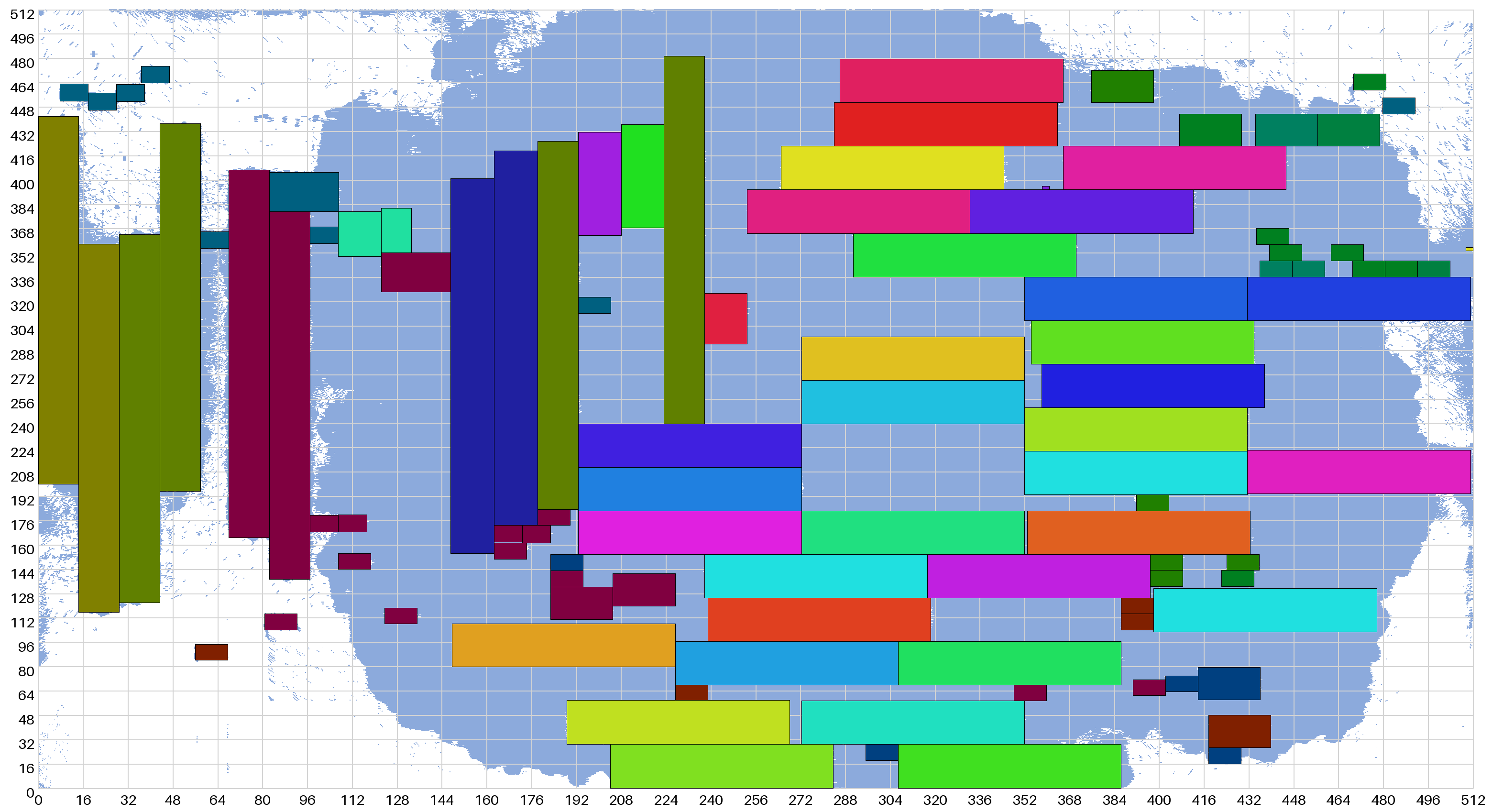} \\
    \end{tabular}
    \caption{Visual comparison of placements on selected Superblue benchmarks. \textbf{Top row:} DREAMPlace 4.3.0. \textbf{Bottom row:} \method-guided DREAMPlace. Blue clouds are individual standard cells; colored rectangles are macros. \method guidance improves gHPWL on all eight Superblue benchmarks.}
    \label{fig:dp_visual_comparison}
\end{figure}

\subsubsection{VLM-Guided Placement}
\begin{figure}[htbp]
    \centering
    \begin{subfigure}{0.32\textwidth}
        \centering
        \includegraphics[width=\linewidth]{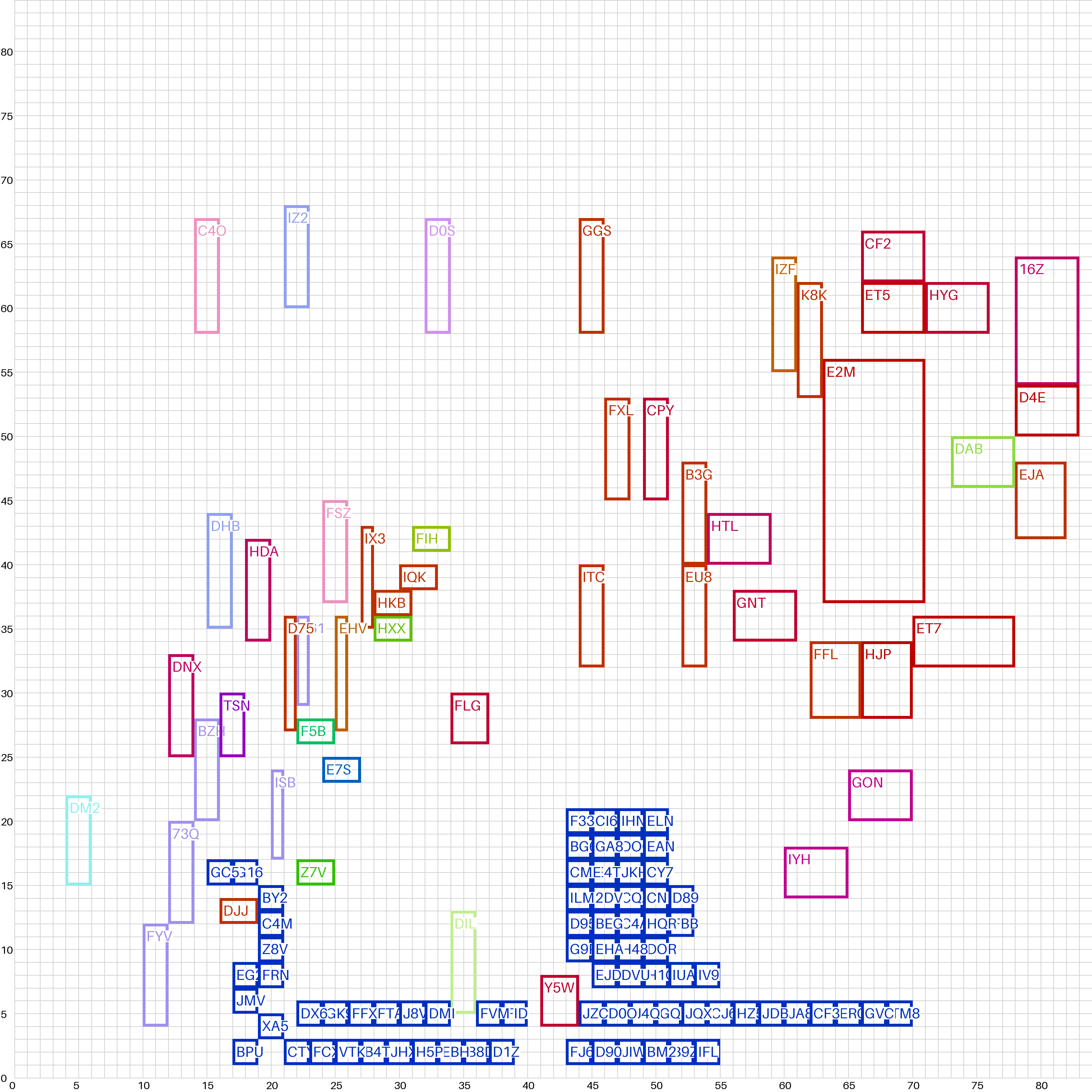}
        \caption{$t = 0$}
        \label{fig:progression_t0}
    \end{subfigure}
    \hfill
    \begin{subfigure}{0.32\textwidth}
        \centering
        \includegraphics[width=\linewidth]{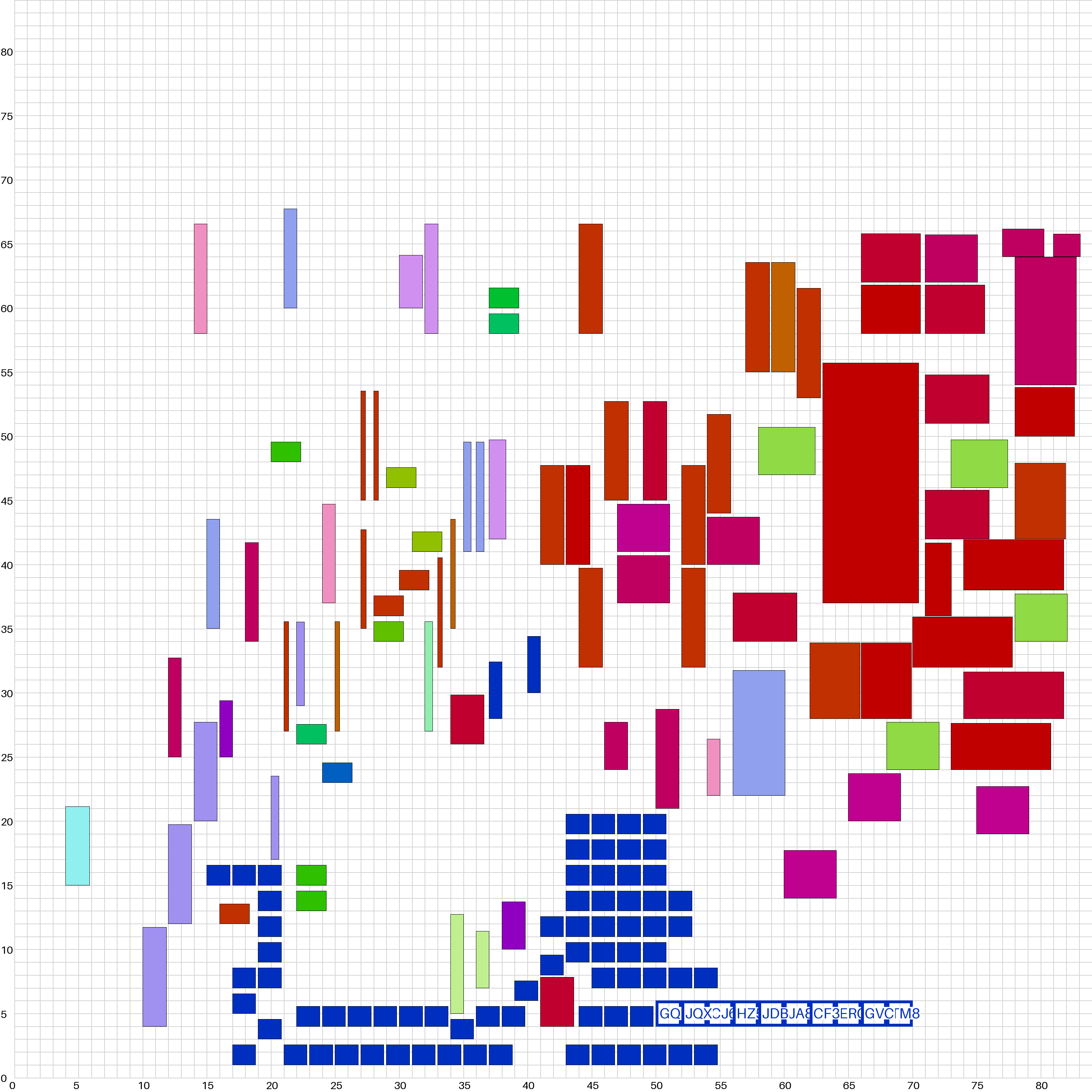}
        \caption{$t = \frac{T}{2}$}
        \label{fig:progression_t_half_T}
    \end{subfigure}
    \hfill
    \begin{subfigure}{0.32\textwidth}
        \centering
        \includegraphics[width=\linewidth]{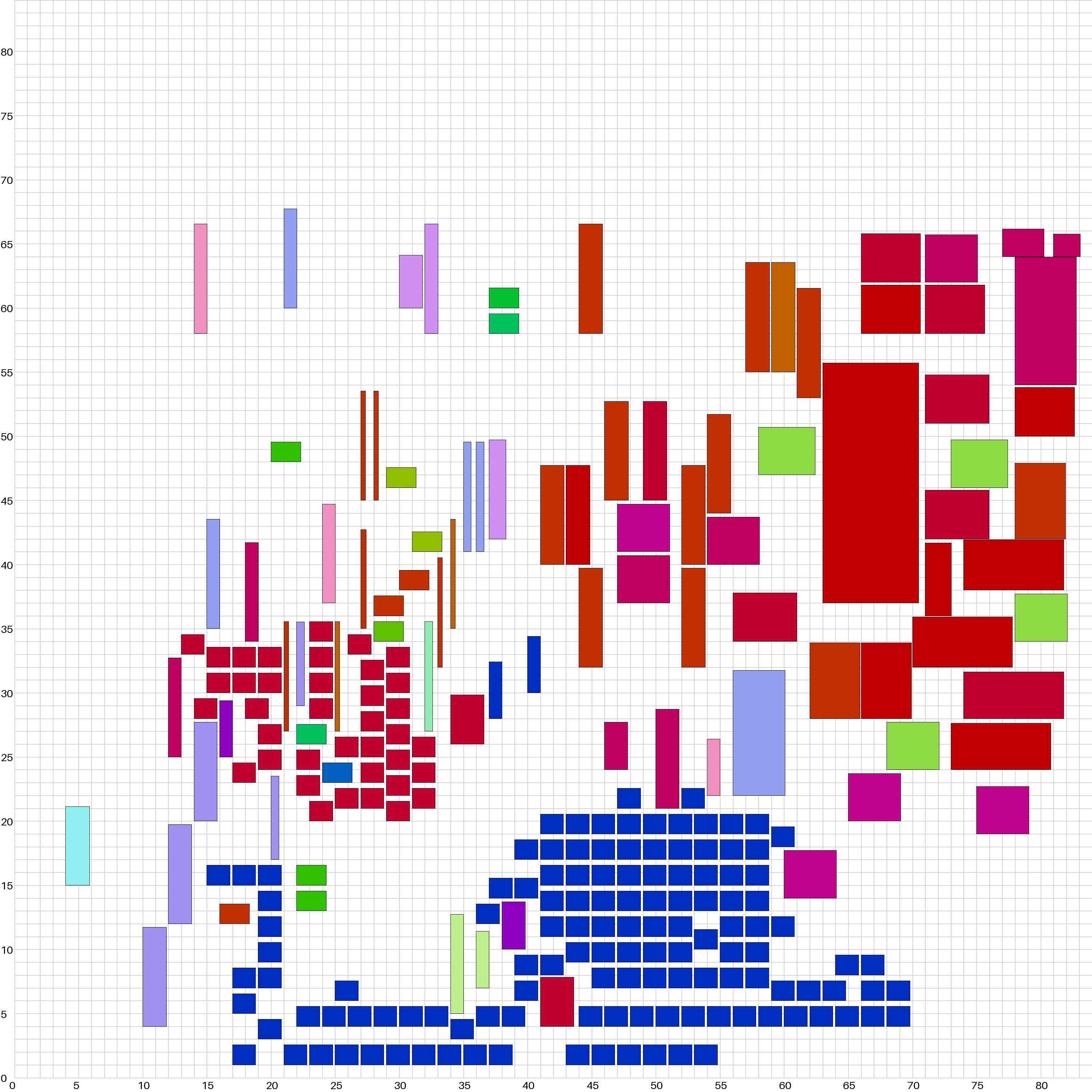}
        \caption{$t = T$}
        \label{fig:progression_t_T}
    \end{subfigure}
    \captionsetup{font=footnotesize}
    \caption{\method's VLM-guided placement on adaptec4. (a) VLM proposes initial regions ($t=0$); policy is unconstrained for macros without valid suggestions. (b) Mid-placement ($t=T/2$). (c) Final placement ($t=T$), with the policy operating within VLM constraints.}
    \label{fig:suggested_regions}
\end{figure}

To provide a qualitative illustration of our method, \Cref{fig:suggested_regions} visualizes a complete VLM-guided rollout on the \texttt{adaptec4} benchmark. This example highlights the hierarchical division of labor central to \method{}: the VLM provides a high-level spatial strategy, and the low-level policy executes the precise placement decisions within that strategy. The figure shows the initial VLM proposals ($t=0$), the state mid-placement ($t=T/2$), and the final floorplan ($t=T$) that results from this guided process.

\subsubsection{Anchor Weight Comparison}
\label{appendix:anchor_weight_visual}

\begin{figure}[htbp]
    \centering
    \begin{subfigure}{\textwidth}
        \centering
        \includegraphics[width=\linewidth]{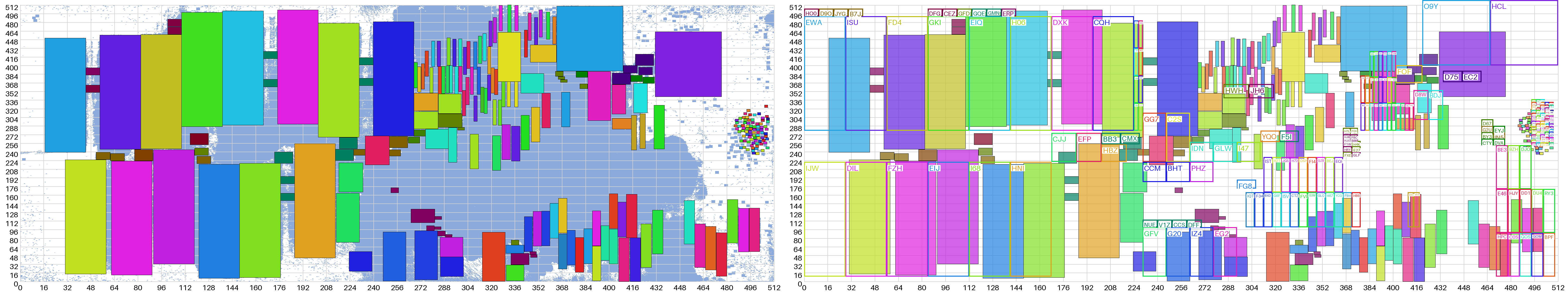}
        \caption{$\lambda_A = 0.001$ (best)}
        \label{fig:anchor_low}
    \end{subfigure}

    \vspace{1em}

    \begin{subfigure}{\textwidth}
        \centering
        \includegraphics[width=\linewidth]{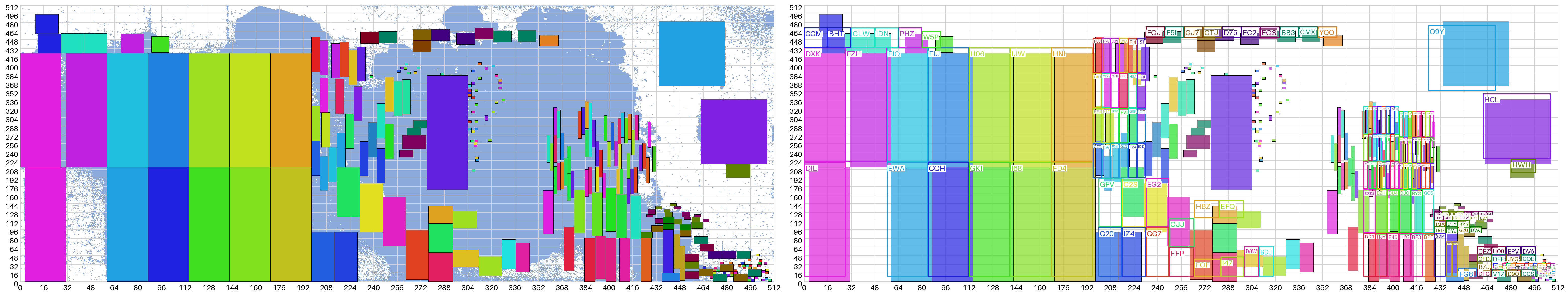}
        \caption{$\lambda_A = 1.0$ (too restrictive)}
        \label{fig:anchor_high}
    \end{subfigure}
    \caption{Visual comparison of placements at different anchor weights on \texttt{superblue1}. Lower anchor weights allow the analytical placer more flexibility to optimize standard cell positions around the macro placements.}
    \label{fig:anchor_weight_visual}
\end{figure}

To illustrate the effect of anchor weight on placement quality, \Cref{fig:anchor_weight_visual} shows the final placements for \texttt{superblue1} at different anchor weights. With low anchor weights ($\lambda_A = 0.001$), DREAMPlace has sufficient flexibility to route standard cells around the VLM-suggested macro positions. Higher anchor weights ($\lambda_A = 1.0$) overly constrain the optimization, resulting in suboptimal standard cell placement.

\subsection{Macro Coloring}
We employ a color-coding strategy to implicitly convey functional relationships between macros to the VLM. This process involves several steps:

First, we construct a \textbf{macro-connectivity graph} from the original netlist $G$. In this graph, nodes (excluding standard cells) represent the macros to be placed. An undirected, weighted edge is created between any two macros if they share one or more nets in $G$. The weight of such an edge is proportional to the number of nets these two macros commonly share. This construction effectively flattens the hypergraph structure of the netlist into a standard graph, where indirect connections through nets are represented as weighted between macros.

This macro-connectivity graph is embedded into a low-dimensional space (specifically, an 8-dimensional space in our implementation for k-means) using a spring-based graph layout algorithm.
Such algorithms, like the one implemented in the NetworkX \cite{network_x} Python library, position macros in the embedding space such that those with stronger connections in the graph are located closer to one another in space.

With macros represented as points in this embedding space, we apply k-means clustering to group them. To determine a suitable number of clusters, $k$, we iterate through a predefined range of potential $k$ values (e.g., from 2 to 30). For each $k$, we perform k-means clustering and evaluate the resulting cluster separation using the Silhouette score \cite{rousseeuw1987silhouettes}.
The value of $k$ (and its corresponding clustering) that yields the highest Silhouette score is selected as optimal.

Finally, macros are assigned colors based on their cluster membership: all macros within the same cluster receive the same unique color. Any macros that are not part of the main connectivity graph (e.g., isolated macros not sharing nets with other considered macros, if any) are assigned a default gray color. The coloring is used to guide VLM suggestions only: it provides the VLM with an approximate visual encoding of netlist connectivity so it can identify which macros are more closely related. The colors do not constrain the final placement.

\section{Experimental Setup}
\label{appendix:experimental_setup}
\subsection{Justification for Fixed Macros During Standard Cell Placement}
\label{app:fixed_macro_justification}

This section applies only to the learning-based evaluation, where the object being evaluated is the macro placement produced by ChiPFormer or \method{} (Section~\ref{sssec:setup_learning}). It does not apply to the analytical-track experiments, where DREAMPlace jointly optimizes macro and standard-cell positions with VLM soft-anchor guidance (Section~\ref{sssec:setup_analytical}).

In the learning-based evaluation, we report results with fixed macro placements, where the analytical placer only positions the standard cells around the macros (a two-stage flow). This differs from some prior work that allows the analytical placer to move all cells, including macros (a three-stage flow). We find that allowing macros to be movable during the analytical placement stage results in substantial displacements from their original \method-generated locations. This confounds the evaluation, as the final placement is no longer representative of \method's learned spatial reasoning but rather the output of the analytical placer.

\begin{figure}[H]
    \centering
    \begin{subfigure}[b]{0.49\textwidth}
        \centering
        \includegraphics[width=\textwidth]{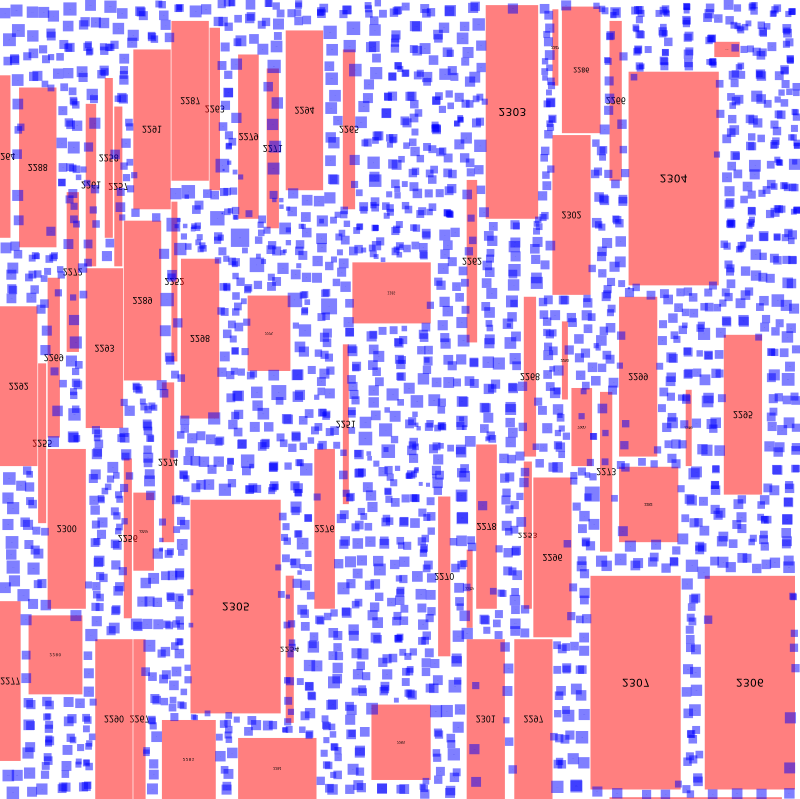} 
        \caption{Our two-stage flow (\method with fixed macros).}
        \label{fig:policy_placement}
    \end{subfigure}
    \hfill %
    \begin{subfigure}[b]{0.49\textwidth}
        \centering
        \includegraphics[width=\textwidth]{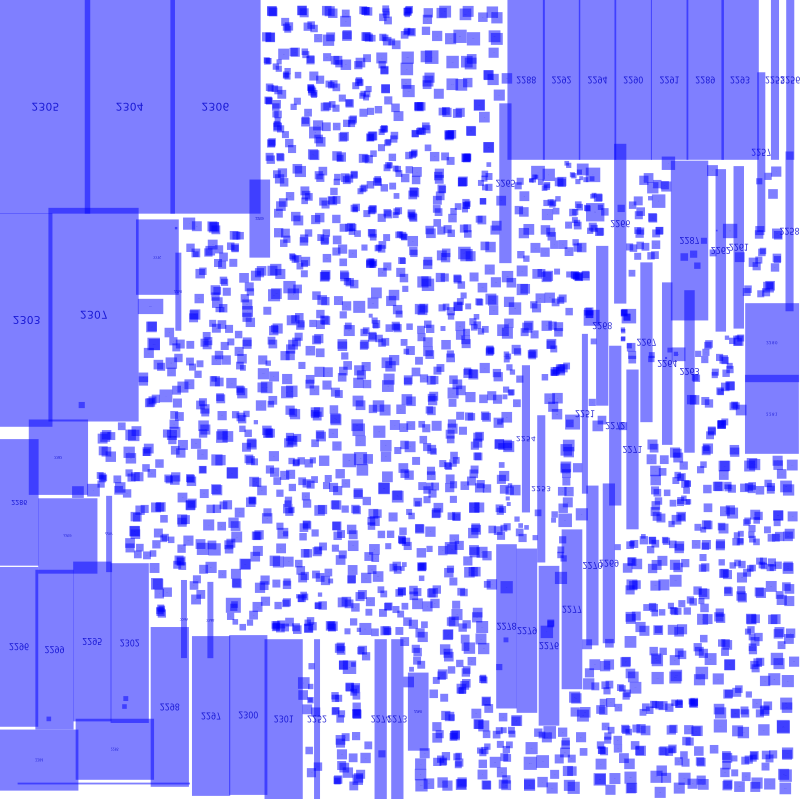} 
        \caption{Alternative three-stage flow (movable macros).}
        \label{fig:dreamplace_final_placement}
    \end{subfigure}
    
\caption{Visual comparison demonstrating why a fixed-macro flow is essential for a fair evaluation of \method. 
(a) The final layout from our two-stage flow, where macro placements generated by \method are \textbf{fixed}. 
(b) The result of an alternative three-stage flow, which takes the \textbf{exact same initial placement} from (a) as input but allows DREAMPlace to move all cells. The analytical placer's drastic rearrangement of macros in (b) shows that it effectively ignores the initial solution, making its final metrics an invalid measure of \method's contribution. Our fixed-macro approach ensures a direct and unconfounded evaluation. The flipped macro IDs are a DREAMPlace rendering artifact. We did not modify the DREAMPlace code.}
\label{fig:dreamplace_movement_analysis}
\end{figure}

\Cref{fig:dreamplace_movement_analysis} provides a clear visual demonstration of this effect on the \texttt{adaptec1} benchmark. The significant displacement of macros is evident, justifying our decision to use a fixed-macro evaluation to ensure we are measuring the direct efficacy of \method.

\subsection{Model Size}
The ChiPFormer decision transformer model contains approximately 3 million trainable parameters.

\subsection{Pre-training}
The pre-training phase was conducted on servers equipped with 4× NVIDIA A100 40GB GPUs. 

\subsection{Rollout and Inference}
Generating ChiPFormer macro-placement candidates is lightweight, requiring only forward passes through the trained 3M-parameter model on a single A100 GPU or equivalent. Each fine-tuning update trains the 3M-parameter model for 10 epochs on the 256-episode buffer and takes seconds, while the eight DREAMPlace evaluations in the same block take minutes, so fine-tuning is under 1\% of a run's wall clock. Reported gHPWL values are computed using the fixed-macro DREAMPlace evaluation flow described in \Cref{sssec:setup_learning}, which places standard cells and legalizes the full design.

\subsection{VLM Integration}
For VLM integration, we used Google's Gemini API: Gemini 2.5 Flash for the learning-based track and the \texttt{superblue1} ablations, Gemini 3.6 Flash for the Superblue results, and Gemini 3.5 Flash for the post-route PPA table. Our experiments are organized into VeoPlace iterations, where each iteration consists of 8 unguided rollouts, one VLM query, and 8 guided rollouts.

Over a 2,000-rollout experiment, this results in exactly 125 calls to the VLM. Each call to Gemini returns 8 candidate generations, with each generation being a complete set of suggested regions for the selected macros.

\subsubsection{Wall-Clock Time}
\label{app:runtime_overhead}
For the analytical-track Superblue experiments, the VLM overhead is dominated by Gemini response latency. Each 2{,}000-rollout run makes 125 queries. Gemini 3.6 Flash averages 62 seconds per query, for about 2.1 hours of added latency per run. \Cref{tab:runtime_overhead} reports the per-rollout DREAMPlace latency and total wall-clock time on a single A100 GPU.

Because the VLM latency is a fixed cost per run, its relative overhead decreases on larger designs, dropping from 7\% on \texttt{superblue18} to 2\% on \texttt{superblue10}.

\begin{table}[H]
    \centering
    \small
    \renewcommand{\arraystretch}{1.05}
    \setlength{\tabcolsep}{4pt}
    \caption{Wall-clock overhead for \method-guided DREAMPlace on Superblue. DP rollout is one vanilla DREAMPlace run; DP/VP+DP report total time for one 2{,}000-rollout run on one A100, with VP+DP including 125 Gemini 3.6 Flash queries at the measured 62 s per query.}
    \label{tab:runtime_overhead}
    \begin{tabularx}{\columnwidth}{@{}l *{4}{>{\centering\arraybackslash}X} @{}}
\toprule
\textbf{Benchmark} & \textbf{DP rollout (s)} & \textbf{DP (h)} & \textbf{VP+DP (h)} & \textbf{Overhead} \\
\midrule
superblue1 & 83 & 46 & 48 & +5\% \\
superblue3 & 168 & 93 & 95 & +2\% \\
superblue4 & 127 & 70 & 72 & +3\% \\
superblue5 & 95 & 53 & 55 & +4\% \\
superblue7 & 122 & 68 & 70 & +3\% \\
superblue10 & 190 & 106 & 108 & +2\% \\
superblue16 & 98 & 55 & 57 & +4\% \\
superblue18 & 52 & 29 & 31 & +7\% \\
\bottomrule
    \end{tabularx}
\end{table}

\subsubsection{API Cost}
\label{app:api_cost}
Using Gemini 2.5 Flash standard paid-tier pricing and conservative overestimates of 100K input tokens and 10K output tokens per query, the estimated cost per query is
\[
0.30 \times \frac{100{,}000}{1{,}000{,}000} + 2.50 \times \frac{10{,}000}{1{,}000{,}000} = 0.055
\]
USD. Across 125 queries, this yields an estimated API cost of
\[
125 \times 0.055 = 6.875
\]
USD, i.e., about \$7 per 2{,}000-rollout experiment. Because the token counts above are intentionally overestimated, the true API cost should be lower.

\section{Hyperparameters}
\label{appendix:hyperparameters}
\subsection{Gemini}
We use the public Gemini API endpoint for our experiments. \Cref{tab:gemini-hparams} shows these hyperparameters.

\begin{table}[H]
\caption{Gemini API hyperparameters.}
\label{tab:gemini-hparams}
\centering
\begin{tabular}{l c}
\toprule
Parameter   & Value \\
\midrule
Temperature & 0.7 \\
Top-$k$     & 64 \\
Top-$p$     & 0.95 \\
Candidates  & 8 \\
\bottomrule
\end{tabular}
\end{table}

\subsection{Standard Cell Grouping Parameters}
\label{app:grouping_params}

As discussed in \Cref{sec:experiments}, for the learning-based benchmarks (adaptec/ibm/ariane), we group the hundreds of thousands of standard cells into a smaller set of clusters to make training and inference tractable. This grouping is only used to compute the proxy wirelength reward during rollouts; all reported results are gHPWL computed by DREAMPlace with fixed macros.

We use the open-source codebase from Google's Circuit Training project for this task, which implements the grouping methodology first introduced by \citet{circuit_training_paper}. To ensure a consistent basis for comparison, we applied the same set of grouping hyperparameters across all learning-based benchmarks. The specific values for these parameters are detailed in Table~\ref{tab:grouping_params}.

\begin{table}[H]
\caption{Hyperparameters for the Standard Cell Grouping Algorithm.}
\label{tab:grouping_params}
\centering
\begin{tabular}{@{}lll@{}}
\toprule
\textbf{Parameter} & \textbf{Value} & \textbf{Description} \\
\midrule
Number of Groups & 2000 & The fixed number of clusters to group all standard cells into. \\
Cell Area Utilization & 1.25 & A target for the density of cells within a cluster. \\
Enable Group Breakup & True & A boolean flag that allows the algorithm to split larger groups. \\
\bottomrule
\end{tabular}
\end{table}

\subsubsection{Pretraining}

\paragraph{Circuit Tokens}
For pretraining the circuit token representation component using the Variational Graph Auto-Encoder (VGAE), we used the following hyperparameters:
\begin{itemize}
    \item Hidden layer dimensions: [32, 32]
    \item Learning rate: 0.01
    \item Training epochs: 800
\end{itemize}

\paragraph{Transformer}
Following ChiPFormer \citep{chipformer}, we use a reward-conditioned transformer with the following hyperparameters:
\begin{itemize}
    \item Number of transformer layers: 6
    \item Number of attention heads: 8
    \item Embedding dimension: 128
    \item Pretraining epochs: 1{,}500
    \item Batch size: 256
    \item Learning rate: $10^{-3}$
    \item Weight decay: $10^{-2}$
\end{itemize}

\subsubsection{Rollout Settings}
\label{appendix:rollout_settings}

\paragraph{Fine-tuning}
The FT baseline follows the ChiPFormer online fine-tuning settings:
\begin{itemize}
    \item Update schedule: after every 8 rollouts
    \item Epochs per update: 10
    \item Batch size: 64
    \item Replay buffer: the 256 best episodes
    \item Optimizer: AdamW with cosine annealing
\end{itemize}
The No-FT baseline is the same code path with the update disabled.

\paragraph{Returns-to-go}
We configured specific target returns-to-go for each benchmark netlist to guide the generated placements. Since our objective is to minimize wirelength, we define the reward as its negative value. Following the methodology of Decision Transformers, we set these values to ambitious targets, encouraging the model to generate high-quality placements with very low wirelengths. Table~\ref{tab:returns-to-go} shows the target returns-to-go values used for each benchmark circuit in our experiments.

\begin{table}[H]
\centering
\caption{Target Returns-to-Go for Different Benchmark Netlists}
\vspace{1em}
\label{tab:returns-to-go}
\renewcommand{\arraystretch}{1.2}
\begin{tabular}{c@{\hspace{2cm}}c}
\toprule
\textbf{Netlist} & \textbf{Return-to-go} \\
\midrule
adaptec1 & -2.86E+06 \\
adaptec2 & -2.91E+06 \\
adaptec3 & -5.90E+06 \\
adaptec4 & -6.37E+06 \\
\midrule
ibm01 & -7.00E+04 \\
ibm02 & -1.67E+05 \\
ibm03 & -2.52E+05 \\
ibm04 & -2.86E+05 \\
\midrule
ariane133 & -2.00E+05 \\
ariane136 & -2.00E+05 \\
\bottomrule
\end{tabular}
\end{table}

\subsection{DREAMPlace}
\label{appendix:dreamplace_hyperparams}

All experiments utilize DREAMPlace version \texttt{4.3.0}.

\subsubsection{VLM-Guided Placement (Superblue)}
\label{appendix:dreamplace_guidance_hyperparams}
For the VLM-guided DREAMPlace experiments on the Superblue benchmarks, we use anchor constraints to incorporate VLM suggestions with a single anchor weight $\lambda_A{=}0.01$ on every benchmark. We chose it on superblue1 by comparing placements at $\lambda_A \in \{1, 0.1, 0.01, 0.001\}$ and keeping the largest weight whose placements follow the VLM suggestions while standard cells still flow freely around the macros.

\Cref{tab:dreamplace_all_params} details the DREAMPlace hyperparameters for all benchmarks. All benchmarks use the Nesterov optimizer with a learning rate of 0.01.

\begin{table}[htb]
\centering
\caption{DREAMPlace 4.3.0 Hyperparameters for All Benchmarks}
\vspace{1em}
\label{tab:dreamplace_all_params}
\renewcommand{\arraystretch}{1.2}
\small
\begin{tabularx}{\textwidth}{l *{6}{>{\centering\arraybackslash}X}}
\toprule
\textbf{Benchmark} & \textbf{Target Density} & \textbf{Stop Overflow} & \textbf{Density Weight} & \textbf{Num Bins (X)} & \textbf{Num Bins (Y)} & \textbf{Iterations} \\
\midrule
\multicolumn{7}{l}{\textit{ICCAD 2015 Superblue}} \\
superblue1 & 1.00 & 0.10 & $8 \times 10^{-5}$ & 1024 & 1024 & 1000 \\
superblue3 & 1.00 & 0.10 & $8 \times 10^{-5}$ & 2048 & 2048 & 1000 \\
superblue4 & 1.00 & 0.10 & $8 \times 10^{-5}$ & 512 & 512 & 1000 \\
superblue5 & 1.00 & 0.10 & $8 \times 10^{-5}$ & 1024 & 1024 & 1000 \\
superblue7 & 1.00 & 0.10 & $8 \times 10^{-5}$ & 512 & 512 & 1000 \\
superblue10 & 1.00 & 0.10 & $8 \times 10^{-5}$ & 1024 & 1024 & 1000 \\
superblue16 & 1.00 & 0.10 & $8 \times 10^{-5}$ & 1024 & 1024 & 1000 \\
superblue18 & 1.00 & 0.10 & $8 \times 10^{-5}$ & 512 & 512 & 1000 \\
\midrule
\multicolumn{7}{l}{\textit{ISPD 2005}} \\
adaptec1 & 1.00 & 0.07 & $8 \times 10^{-5}$ & 512 & 512 & 1000 \\
adaptec2 & 1.00 & 0.07 & $8 \times 10^{-5}$ & 1024 & 1024 & 1000 \\
adaptec3 & 1.00 & 0.07 & $8 \times 10^{-5}$ & 1024 & 1024 & 1000 \\
adaptec4 & 1.00 & 0.07 & $8 \times 10^{-5}$ & 1024 & 1024 & 1000 \\
\midrule
\multicolumn{7}{l}{\textit{ICCAD 2004}} \\
ibm01 & 1.00 & 0.07 & $8 \times 10^{-5}$ & 512 & 512 & 1000 \\
ibm02 & 1.00 & 0.07 & $8 \times 10^{-5}$ & 512 & 512 & 1000 \\
ibm03 & 1.00 & 0.07 & $8 \times 10^{-5}$ & 512 & 512 & 1000 \\
ibm04 & 1.00 & 0.07 & $8 \times 10^{-5}$ & 512 & 512 & 1000 \\
\midrule
\multicolumn{7}{l}{\textit{Ariane (Nangate45)}} \\
ariane133 & 1.00 & 0.07 & $8 \times 10^{-5}$ & 512 & 512 & 1000 \\
ariane136 & 1.00 & 0.07 & $8 \times 10^{-5}$ & 512 & 512 & 1000 \\
\bottomrule
\end{tabularx}
\end{table}

\section{Additional Experiments}
\label{appendix:additional_experiments}

\subsection{Congestion}
We report a congestion proxy (RUDY) for the Superblue benchmarks to verify that \method's wirelength gains do not come at the expense of routability. The differences are small and mixed across benchmarks.

\begin{table}[H]
\centering
\small
\caption{Congestion proxy (RUDY, lower is better) for \method-guided DREAMPlace vs.\ DREAMPlace 4.3.0 on Superblue. VP+DP uses Gemini 2.5 Flash; mean $\pm$ standard error across three seeds. Differences are small and mixed (4--4 split), indicating that \method's gHPWL improvements do not materially change routability.}
\label{tab:superblue_congestion}
\begin{tabularx}{\columnwidth}{@{}l *{2}{>{\centering\arraybackslash}X} @{}}
\toprule
\textbf{Benchmark} & \textbf{VP+DP 2.5 Flash} & \textbf{DP 4.3.0} \\
\midrule
superblue1 & 0.93$\pm$0.01 & \textbf{0.92$\pm$0.00} \\
superblue3 & 1.06$\pm$0.02 & \textbf{1.04$\pm$0.00} \\
superblue4 & \textbf{0.91$\pm$0.02} & 0.93$\pm$0.00 \\
superblue5 & 0.85$\pm$0.02 & \textbf{0.80$\pm$0.00} \\
superblue7 & \textbf{1.03$\pm$0.00} & 1.06$\pm$0.00 \\
superblue10 & \textbf{1.08$\pm$0.02} & 1.09$\pm$0.01 \\
superblue16 & 1.13$\pm$0.02 & \textbf{1.09$\pm$0.02} \\
superblue18 & \textbf{1.02$\pm$0.01} & 1.10$\pm$0.01 \\
\bottomrule
\end{tabularx}
\end{table}

\subsection{Seed-Paired DREAMPlace Comparison}
\label{appendix:paired_seeds}
Each VP+DP run in \Cref{tab:superblue_results} shares its random seed with its DREAMPlace 4.3.0 baseline, so the comparison is paired by construction. \Cref{tab:superblue_paired} lists the per-seed differences. VP+DP improves on its seed-paired baseline in 38 of the 40 pairs; the two exceptions are superblue18 seeds 4 and 5.

\begin{table}[H]
\centering
\small
\caption{Per-seed paired gHPWL differences ($\times 10^6$), DREAMPlace 4.3.0 minus VP+DP, on Superblue.
Positive means VP+DP is better.
Green marks pairs where VP+DP is better, and red marks pairs where DREAMPlace 4.3.0 is better.
Each column pairs the two runs that share that seed.
The per-design means are the Table~\ref{tab:superblue_results} gaps.}
\label{tab:superblue_paired}
\begin{tabularx}{\columnwidth}{@{}l *{5}{>{\centering\arraybackslash}X} @{}}
\toprule
\textbf{Benchmark} & \textbf{Seed 1} & \textbf{Seed 2} & \textbf{Seed 3} & \textbf{Seed 4} & \textbf{Seed 5} \\
\midrule
superblue1 & \textcolor{gain}{4.26} & \textcolor{gain}{4.37} & \textcolor{gain}{2.41} & \textcolor{gain}{4.01} & \textcolor{gain}{3.86} \\
superblue3 & \textcolor{gain}{15.93} & \textcolor{gain}{13.53} & \textcolor{gain}{17.84} & \textcolor{gain}{16.00} & \textcolor{gain}{14.72} \\
superblue4 & \textcolor{gain}{2.47} & \textcolor{gain}{3.91} & \textcolor{gain}{4.36} & \textcolor{gain}{4.55} & \textcolor{gain}{5.18} \\
superblue5 & \textcolor{gain}{4.93} & \textcolor{gain}{3.39} & \textcolor{gain}{2.27} & \textcolor{gain}{2.51} & \textcolor{gain}{1.80} \\
superblue7 & \textcolor{gain}{6.39} & \textcolor{gain}{9.14} & \textcolor{gain}{6.59} & \textcolor{gain}{3.09} & \textcolor{gain}{3.25} \\
superblue10 & \textcolor{gain}{33.92} & \textcolor{gain}{32.16} & \textcolor{gain}{27.10} & \textcolor{gain}{32.07} & \textcolor{gain}{30.86} \\
superblue16 & \textcolor{gain}{2.00} & \textcolor{gain}{1.64} & \textcolor{gain}{1.83} & \textcolor{gain}{2.13} & \textcolor{gain}{1.96} \\
superblue18 & \textcolor{gain}{3.58} & \textcolor{gain}{3.34} & \textcolor{gain}{3.86} & \textcolor{red}{$-$0.99} & \textcolor{red}{$-$1.10} \\
\bottomrule
\end{tabularx}
\end{table}

\subsection{Placement Metrics and Downstream PPA}
\label{appendix:wirelength_ppa}

Macro placement is an early physical-design stage, so our evaluation follows the standard placement protocol of optimizing gHPWL after standard-cell placement and legalization, while also checking density or routability proxies~\citep{cheng2018replace,lin_dreamplace_2019,circuit_training_paper,chipformer,lai2022maskplace}. Full sign-off metrics such as WNS, TNS, DRV count, and power are determined by the complete downstream physical-design flow, including detailed placement, buffering, clock-tree synthesis, routing, and timing optimization. Running this flow can take weeks in realistic design settings, making it impractical as an inner-loop objective for evolutionary search. It would also conflate macro placement quality with downstream implementation choices, and many academic placement benchmarks do not provide the technology libraries and sign-off collateral needed for timing and power analysis.

Wirelength is the canonical objective for placement algorithms and directly targets physical drivers of downstream PPA. Dynamic power scales with switched capacitance, $P_{\mathrm{dyn}} \propto \alpha C V_{dd}^{2} f$, and wire capacitance is a major component of load capacitance in modern VLSI processes~\citep{rabaey2003digital,weste2011cmos}. Timing is similarly affected by interconnect RC delay, which grows with wire resistance and capacitance. Thus, reducing gHPWL addresses a central source of delay and power at the placement stage.

Because shorter wirelength can still coincide with worse routability, \Cref{tab:superblue_congestion} reports RUDY congestion on the Superblue benchmarks. The differences are small and mixed across designs, indicating that \method's wirelength improvements do not materially degrade this congestion proxy. If timing, power, or routing metrics are available in a deployment flow, they can also be included in the VLM prompt and used as additional scores in the same evolutionary loop.

We run the full flow on four designs with open Nangate45 collateral (Ariane-133, Ariane-136, BlackParrot, BP-Quad), pushing every method's macro placement through the identical OpenROAD flow through detailed routing (\Cref{tab:ppa_nangate45}). VP+DP has the shortest routed wirelength on Ariane-136, BlackParrot, and BP-Quad and the smallest area on all four designs, with best or tied TNS and violating pins on every design.

\begin{table}[H]
\centering
\footnotesize
\caption{Post-route PPA on four Nangate45 designs after the full OpenROAD flow.
The metrics are routed wirelength (WL, $10^6\,\mu$m), worst negative slack (WNS, ns), total negative slack (TNS, $10^3$ ns), number of violating pins (NVP), power (mW), area ($10^3\,\mu$m$^2$), and DRC violations.
VP+DP uses Gemini 3.5 Flash.
Bold is best and underlined is second best.
Error terms are standard deviations over three seeds.
An asterisk ($^{*}$) marks a mean over two seeds, because the third seed failed to route in OpenROAD.
Plain values are single runs.}
\label{tab:ppa_nangate45}
\renewcommand{\arraystretch}{1.05}
\begin{tabular}{@{}l l r r r r@{}}
\toprule
\textbf{Method} & \textbf{Metric} & \textbf{Ariane-133} & \textbf{Ariane-136} & \textbf{BlackParrot} & \textbf{BP-Quad} \\
\midrule
\multirow{7}{*}{Hier-RTLMP} & Routed WL & 6.91 & 7.46 & 6.11 & 52.56 \\
 & WNS & -0.411 & \textbf{0.432} & -2.991 & -150.056 \\
 & TNS & -0.719 & \textbf{0.000} & \textbf{-0.003} & -38511.600 \\
 & NVP & 2894 & \textbf{0} & \textbf{1} & \textbf{259747} \\
 & Power & \textbf{254.3} & 162.1 & \underline{147.0} & \textbf{2008.7} \\
 & Area & 727.7 & 748.6 & 645.2 & 5402.6 \\
 & DRC & \textbf{0} & \textbf{0} & \textbf{0} & \textbf{0} \\
\midrule
\multirow{7}{*}{DP 4.3.0} & Routed WL & \textbf{5.93$\pm$0.19} & \underline{6.43$\pm$0.05} & \underline{5.71$\pm$0.02} & \underline{45.91$\pm$0.16} \\
 & WNS & \underline{-0.368$\pm$0.041} & 0.268$\pm$0.055 & -2.990$\pm$0.028 & \underline{-149.712$\pm$0.087} \\
 & TNS & \underline{-0.645$\pm$0.061} & \textbf{0.000} & \textbf{-0.003} & \underline{-38497.433$\pm$4.661} \\
 & NVP & \underline{2838$\pm$127} & \textbf{0} & \textbf{1} & \textbf{259747} \\
 & Power & \underline{254.4$\pm$1.1} & \textbf{153.6$\pm$0.1} & 147.6$\pm$0.3 & 2018.1$\pm$6.6 \\
 & Area & \underline{726.9$\pm$1.4} & \underline{747.9$\pm$0.2} & \underline{643.5$\pm$0.0} & \underline{5355.6$\pm$0.7} \\
 & DRC & \textbf{0} & \textbf{0} & \textbf{0} & \textbf{0} \\
\midrule
\multirow{7}{*}{WireMask-EA} & Routed WL & 7.51$\pm$0.79$^{*}$ & 7.27$\pm$0.62 & 6.88$\pm$0.22 & 58.22 \\
 & WNS & -0.820$\pm$0.440$^{*}$ & 0.114$\pm$0.087 & \textbf{-2.938$\pm$0.049} & -150.507 \\
 & TNS & -2.002$\pm$0.827$^{*}$ & \textbf{0.000} & \textbf{-0.003} & -38547.600 \\
 & NVP & 4326$\pm$963$^{*}$ & \textbf{0} & \textbf{1} & \textbf{259747} \\
 & Power & 258.7$\pm$0.4$^{*}$ & 154.9$\pm$0.8 & 151.3$\pm$5.8 & 2058.5 \\
 & Area & 730.5$\pm$1.5$^{*}$ & 751.6$\pm$1.9 & 647.5$\pm$1.0 & 5454.7 \\
 & DRC & \textbf{0}$^{*}$ & \textbf{0} & \textbf{0} & \textbf{0} \\
\midrule
\multirow{7}{*}{EGPlace} & Routed WL & 7.12$\pm$0.04$^{*}$ & 7.13$\pm$0.45$^{*}$ & 7.32$\pm$0.41 & 58.44$\pm$1.63 \\
 & WNS & -0.688$\pm$0.039$^{*}$ & -0.062$\pm$0.005$^{*}$ & -3.025$\pm$0.072 & -150.504$\pm$0.308 \\
 & TNS & -1.431$\pm$0.160$^{*}$ & \underline{-0.004$\pm$0.001}$^{*}$ & \textbf{-0.003} & -38552.300$\pm$11.772 \\
 & NVP & 3321$\pm$62$^{*}$ & \underline{142$\pm$28}$^{*}$ & \textbf{1} & \textbf{259747} \\
 & Power & 258.6$\pm$0.9$^{*}$ & 155.9$\pm$1.7$^{*}$ & 153.0$\pm$5.0 & 2073.1$\pm$22.3 \\
 & Area & 731.1$\pm$0.9$^{*}$ & 750.8$\pm$1.8$^{*}$ & 648.6$\pm$1.6 & 5461.6$\pm$15.9 \\
 & DRC & \textbf{0}$^{*}$ & \textbf{0}$^{*}$ & \textbf{0} & \textbf{0} \\
\midrule
\multirow{7}{*}{LaMPlace} & Routed WL & 7.49$\pm$1.21 & 7.59$\pm$0.51 & 7.33$\pm$0.29 & 63.92$\pm$2.14$^{*}$ \\
 & WNS & -0.496$\pm$0.171 & \underline{0.317$\pm$0.070} & -3.028$\pm$0.065 & -150.498$\pm$0.006$^{*}$ \\
 & TNS & -1.259$\pm$0.767 & \textbf{0.000} & \textbf{-0.003} & -38547.200$\pm$3.323$^{*}$ \\
 & NVP & 3896$\pm$1424 & \textbf{0} & \textbf{1} & \textbf{259747}$^{*}$ \\
 & Power & 256.9$\pm$2.2 & 160.7$\pm$2.8 & 153.3$\pm$2.7 & 2051.5$\pm$9.3$^{*}$ \\
 & Area & 732.0$\pm$6.9 & 751.2$\pm$1.3 & 648.7$\pm$1.1 & 5512.0$\pm$16.3$^{*}$ \\
 & DRC & \textbf{0} & \textbf{0} & \textbf{0} & \textbf{0}$^{*}$ \\
\midrule
\multirow{7}{*}{VP+DP 3.5 Flash} & Routed WL & \underline{5.98$\pm$0.09} & \textbf{6.35$\pm$0.04} & \textbf{5.58$\pm$0.02} & \textbf{44.57$\pm$0.93} \\
 & WNS & \textbf{-0.339$\pm$0.026} & 0.277$\pm$0.121 & \underline{-2.965$\pm$0.014} & \textbf{-149.585$\pm$0.029} \\
 & TNS & \textbf{-0.549$\pm$0.039} & \textbf{0.000} & \textbf{-0.003} & \textbf{-38492.100$\pm$5.450} \\
 & NVP & \textbf{2523$\pm$74} & \textbf{0} & \textbf{1} & \textbf{259747} \\
 & Power & 255.3$\pm$0.5 & \underline{153.7$\pm$0.5} & \textbf{146.4$\pm$0.5} & \underline{2012.5$\pm$7.8} \\
 & Area & \textbf{726.2$\pm$0.8} & \textbf{747.7$\pm$0.1} & \textbf{643.3$\pm$0.2} & \textbf{5343.9$\pm$6.2} \\
 & DRC & \textbf{0} & \textbf{0} & \textbf{0} & \textbf{0} \\
\bottomrule
\end{tabular}
\end{table}

\subsection{Design Choice Ablations}
\label{appendix:design_choice_ablations}

\begin{figure}[H]
    \centering
    \begin{subfigure}[t]{0.24\textwidth}
        \centering
        \includegraphics[width=\linewidth]{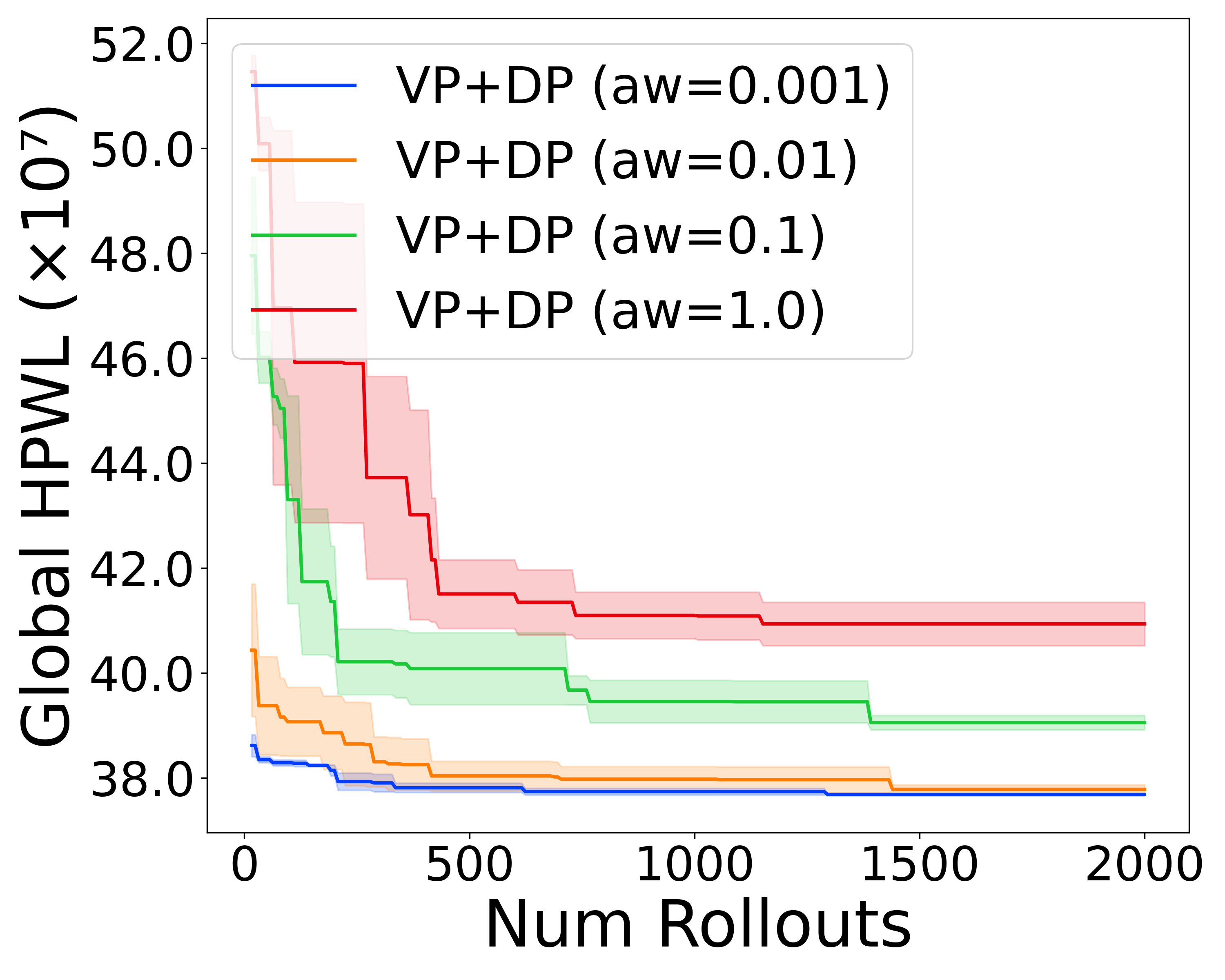}
        \caption{Anchor weight $\lambda_A$}
        \label{fig:ablation_anchor}
    \end{subfigure}
    \hfill
    \begin{subfigure}[t]{0.24\textwidth}
        \centering
        \includegraphics[width=\linewidth]{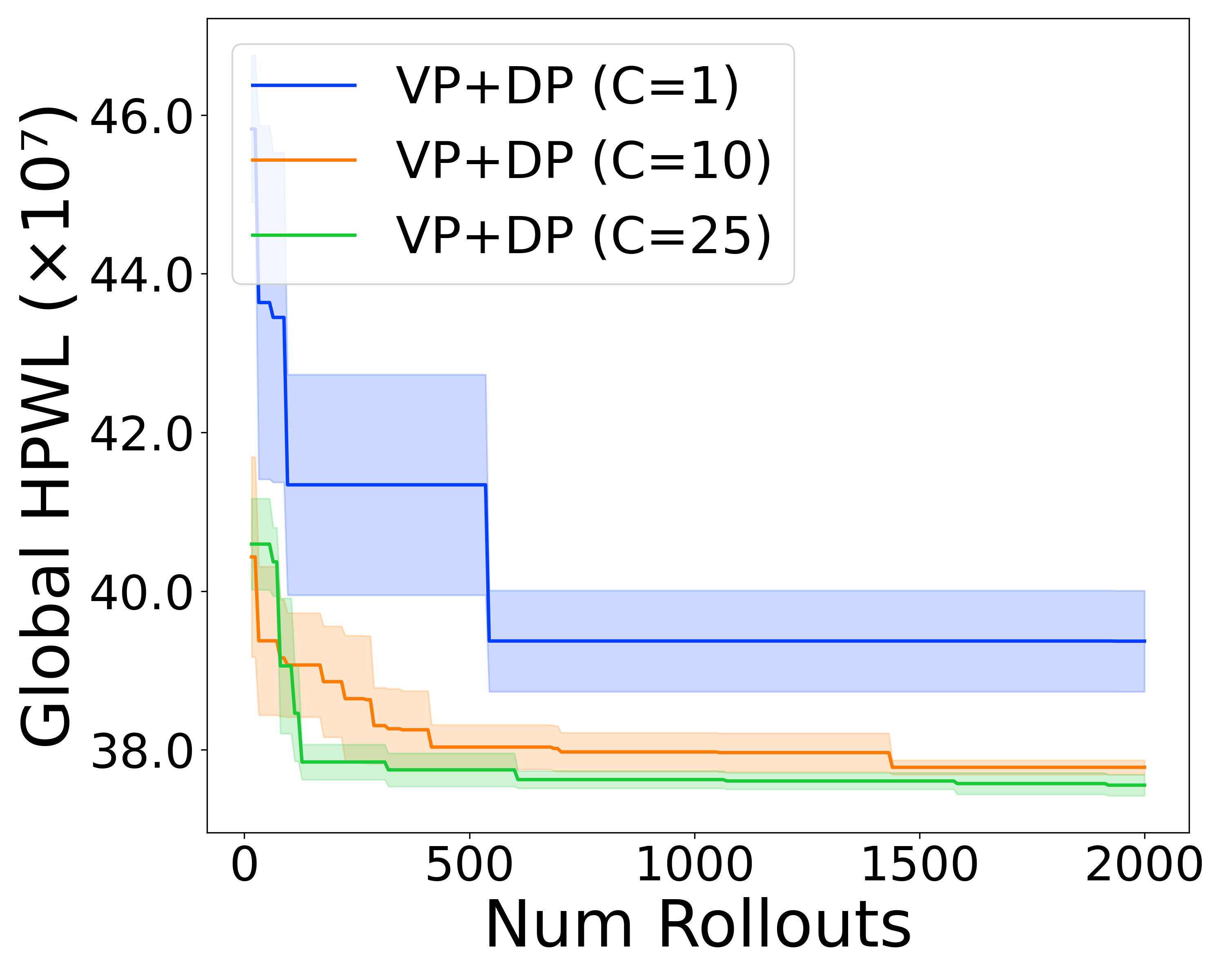}
        \caption{Context length $C$}
        \label{fig:ablation_context}
    \end{subfigure}
    \hfill
    \begin{subfigure}[t]{0.24\textwidth}
        \centering
        \includegraphics[width=\linewidth]{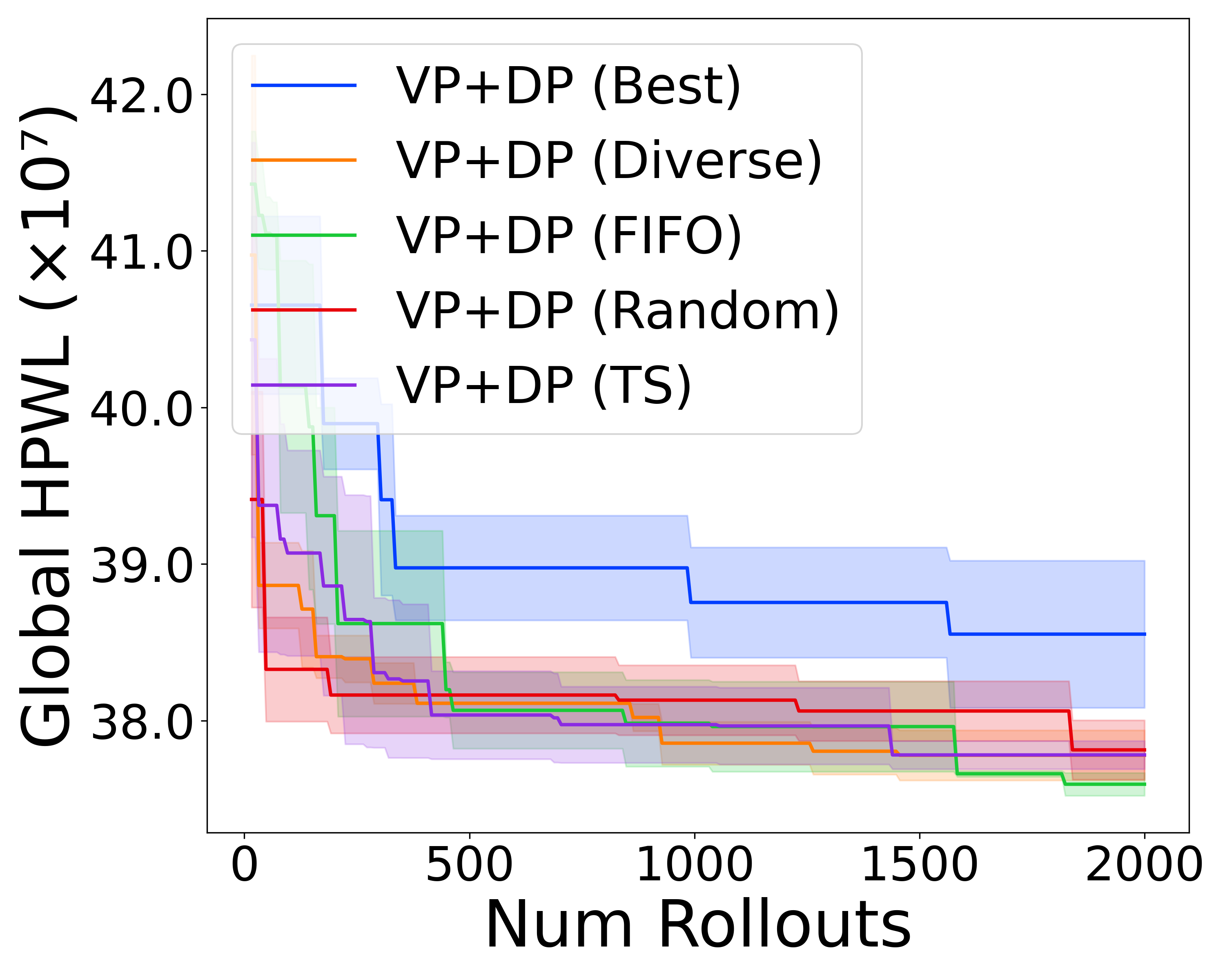}
        \caption{Context selection strategy}
        \label{fig:ablation_strategy}
    \end{subfigure}
    \hfill
    \begin{subfigure}[t]{0.24\textwidth}
        \centering
        \includegraphics[width=\linewidth]{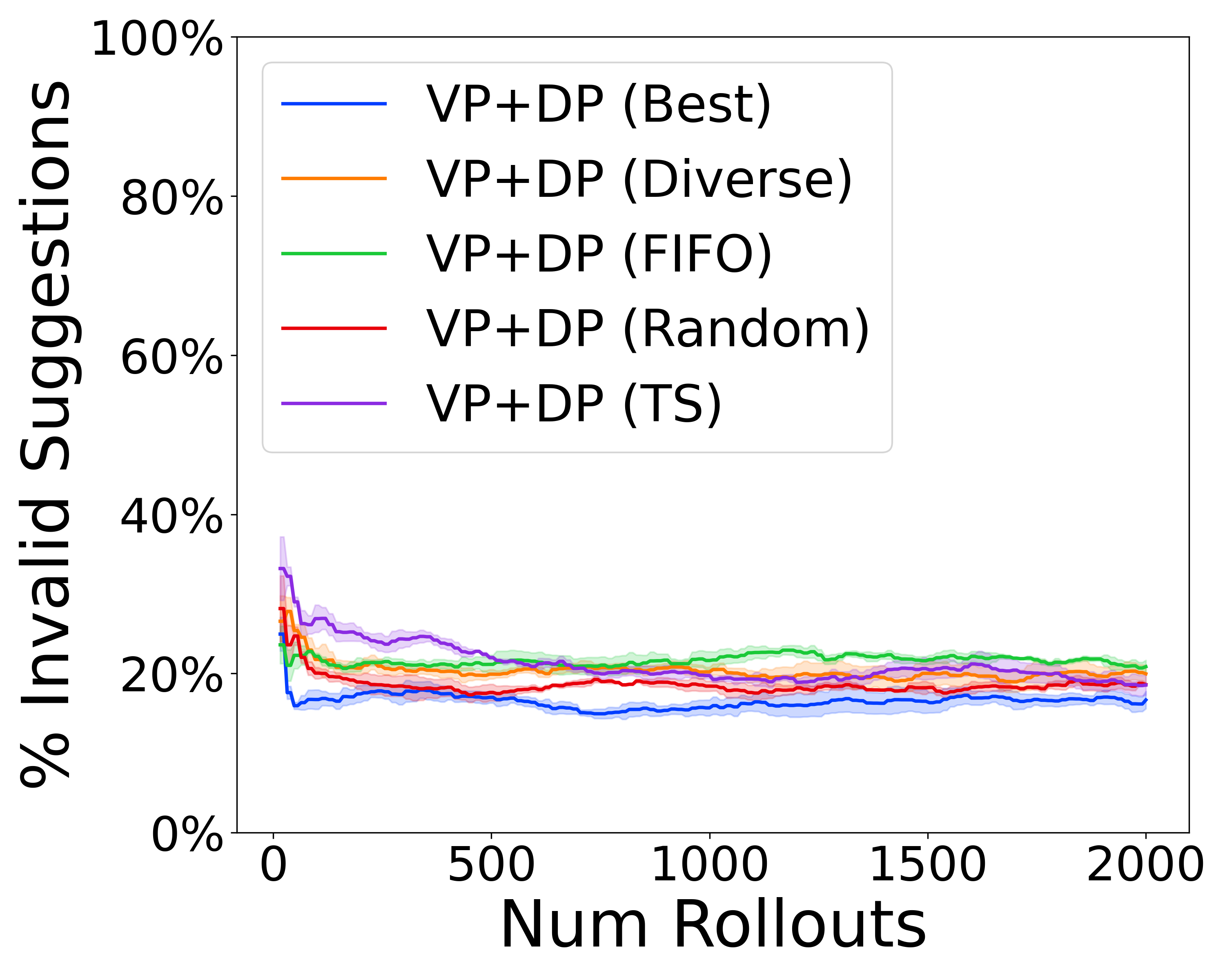}
        \caption{Invalid suggestion rate}
        \label{fig:invalid_suggestions}
    \end{subfigure}
    \caption{Ablations on \texttt{superblue1} with \method-guided DREAMPlace (Gemini 2.5 Flash, three seeds). Defaults: top stratified (TS), $C{=}10$, $\lambda_A{=}0.01$; main Superblue results use $C{=}25$. Lower anchor weights and longer context perform best; all context strategies improve over rollouts, with invalid suggestions converging to ${\sim}20\%$.}
    \label{fig:ablations}
\end{figure}

\Cref{fig:ablations} shows the full ablation plots discussed in \Cref{ssec:ablations}.

\subsection{Evolutionary Search Controls}
\label{appendix:search_controls}
The random-regions control replaces the VLM's proposals with regions drawn at random, matching the VLM arm in number of proposals per iteration, region format, and anchoring pipeline. The best-so-far-anchor control anchors each rollout to the best placement found so far with a soft spring at $\lambda_A{=}0.01$, a local perturbation of the best solution. Both controls use the same search loop and global $\lambda_A{=}0.01$ as \method and run 4{,}000 rollouts, twice the \method budget, on all eight Superblue designs with five seeds. VP+DP beats both controls on all eight designs (\Cref{tab:search_controls}): random regions sharply degrade placement quality, while best-so-far anchoring performs comparably to DREAMPlace alone.

\begin{table}[H]
    \centering
    \captionsetup{font=scriptsize,skip=2pt}
    \scriptsize
    \renewcommand{\arraystretch}{0.92}
    \setlength{\tabcolsep}{1.3pt}
    \caption{VP+DP (Gemini 3.6 Flash) vs.\ DREAMPlace 4.3.0 (DP) and two evolutionary search controls.
All columns report gHPWL ($\times 10^6$, lower is better), mean $\pm$ standard error across five seeds.
DP 4.3.0 and both controls run 4{,}000 rollouts.
VP+DP runs 2{,}000 rollouts.
Best results are bolded.}
    \label{tab:search_controls}
    \begin{tabularx}{\textwidth}{@{}l *{8}{>{\centering\arraybackslash}X}@{}}
\toprule
\textbf{Method (rollouts)} & \textbf{superblue1} & \textbf{superblue3} & \textbf{superblue4} & \textbf{superblue5} & \textbf{superblue7} & \textbf{superblue10} & \textbf{superblue16} & \textbf{superblue18} \\
\midrule
DP 4.3.0 (4{,}000) & \underline{376.61$\pm$0.12} & 453.64$\pm$0.27 & 290.90$\pm$0.25 & \underline{399.94$\pm$0.25} & \underline{544.36$\pm$0.31} & 714.05$\pm$1.30 & \underline{366.21$\pm$0.07} & \underline{219.13$\pm$0.05} \\
VP+DP 3.6 Flash (2{,}000) & \textbf{372.85$\pm$0.34} & \textbf{438.14$\pm$0.42} & \textbf{286.82$\pm$0.55} & \textbf{397.90$\pm$0.41} & \textbf{538.76$\pm$0.86} & \textbf{683.25$\pm$1.22} & \textbf{364.43$\pm$0.09} & \textbf{217.50$\pm$1.09} \\
Random regions (4{,}000) & 496.75$\pm$3.99 & 584.81$\pm$2.88 & 352.41$\pm$0.34 & 599.84$\pm$2.41 & 712.07$\pm$2.53 & 748.64$\pm$2.23 & 406.88$\pm$1.27 & 236.62$\pm$0.49 \\
Best-so-far anchor (4{,}000) & 382.79$\pm$0.72 & \underline{451.09$\pm$1.20} & \underline{289.09$\pm$0.43} & 406.20$\pm$0.74 & 544.93$\pm$0.28 & \underline{702.34$\pm$2.69} & 366.55$\pm$0.11 & 221.35$\pm$0.31 \\
\bottomrule
    \end{tabularx}
\end{table}

\subsection{Context Selection Strategies}
\label{appendix:context_selection_strategies}

We compare five strategies for selecting the $C$ in-context examples from the history buffer:

\paragraph{Most Recent (FIFO).} Select the $C$ most recently generated placements from the population. This strategy implements pure evolutionary search, where the VLM observes a temporally ordered sequence of recent attempts.

\paragraph{Random.} Randomly sample $C$ placements from the population buffer. This provides a baseline that makes no assumptions about which examples are most valuable for evolutionary search.

\paragraph{Best Performing.} Select the $C$ placements with the lowest gHPWL from the population to encourage the VLM to replicate successful patterns.

\paragraph{Diverse.} Represent each placement as a vector in $\mathbb{R}^{2T}$ by concatenating the $(x_i, y_i)$ coordinates of its $T$ macros. We then perform K-means clustering with $C$ clusters on the population and select the placement with the best gHPWL from each cluster. This promotes geometric diversity across the selected examples while still favoring high-quality designs.

\paragraph{Top Stratified.} Represent placements as coordinate vectors and cluster them by geometric similarity, as in the diverse strategy. We then focus on a single promising cluster by ranking all clusters by their best gHPWL and sampling one using a rank-based softmax distribution (with temperature $\tau = 0.43$). From the selected cluster, we choose the top $C$ performing layouts, supplementing from the nearest clusters by centroid distance if the selected cluster contains fewer than $C$ members. This strategy provides a set of geometrically similar examples that represent variations of a particular design pattern.

\subsection{Prompt Ablation}
To isolate the impact of the prompt's high-level strategic guidance, we conducted an ablation study to test the sensitivity of the VLM's performance to its core instructions. A key question is whether the VLM performs best when asked to explore novel design configurations, to greedily refine known high-quality solutions, or to follow a more balanced default instruction.

To investigate this, we created two strategic variants of our main prompt: a \textbf{Greedy} prompt that explicitly instructs the VLM to make only minor modifications to the best-performing examples provided in-context, and an \textbf{Exploratory} prompt that encourages the VLM to disregard prior examples and generate creative, novel placements. \Cref{fig:appendix_prompt_comparison} shows the exact strategy-specific prompt text used by these three variants.

The results of this ablation, presented in \Cref{tab:appendix_prompt_ablation} for the \texttt{superblue1} benchmark, show that prompt intent matters, but the strongest performance comes from the \textbf{default} prompt rather than from either extreme. The \textbf{Greedy} prompt is slightly worse, and the \textbf{Exploratory} prompt degrades further. This suggests that, for guiding DREAMPlace, the VLM works best when given balanced instructions that preserve useful inductive bias from prior examples without over-constraining the search.

\begin{figure}[H]
\centering
\small
\caption{Strategy-specific prompt text extracted from the implemented prompt templates. The \textbf{Greedy} prompt adds stay-close instructions, the \textbf{Default} prompt keeps the neutral scaffold, and the \textbf{Exploratory} prompt adds explicit novelty-seeking guidance.}
\label{fig:appendix_prompt_comparison}
\renewcommand{\arraystretch}{1.5}
\begin{tabularx}{\textwidth}{>{\raggedright\arraybackslash}X >{\raggedright\arraybackslash}X >{\raggedright\arraybackslash}X}
    \toprule
    \multicolumn{1}{c}{\bfseries Greedy Prompt} &
    \multicolumn{1}{c}{\bfseries Default Prompt} &
    \multicolumn{1}{c}{\bfseries Exploratory Prompt} \\
    \midrule
    \small \textit{Preamble addition:} ``Use the previous examples as a strong prior. Stay very close to the best known historical arrangements and prefer small, low-risk refinements over novel departures.'' \newline\newline \textit{Strategy section:} ``Refinement on Best Archetype: Propose exactly one small, low-risk improvement to the best known pattern. Stay very close to the successful historical arrangement and explain why this minimal change should help.'' &
    \small \textit{Preamble addition:} none. \newline\newline \textit{Strategy section:} ``Improvement on Best Archetype: Propose exactly one concrete change to the best pattern (e.g., shift position, adjust spacing, try different canvas region); if evidence is weak, label it `exploratory' but still commit.'' &
    \small \textit{Preamble addition:} ``Use the previous examples as context only, not as a constraint. Favor novel, diverse, and deliberately exploratory region proposals when they could plausibly lead to a better floorplan.'' \newline\newline \textit{Strategy section:} ``Deliberate Departure from Best Archetype: Propose exactly one concrete way your plan intentionally differs from the best known pattern. Explain why this novel configuration could outperform it; if evidence is weak, label it an `exploratory hypothesis' but still commit.'' \\
    \bottomrule
\end{tabularx}
\end{figure}

\begin{table}[H]
\centering
\small
\caption{Ablation study on the VLM prompt's strategic guidance (TS, $C{=}25$) on the \texttt{superblue1} benchmark. We report \method{} with DREAMPlace (VP+DP) using Gemini 2.5 Flash, mean $\pm$ standard error across three seeds. The \textbf{Default} prompt yields the best objective ($\times 10^7$, lower is better). The best result is \textbf{bolded}.}
\label{tab:appendix_prompt_ablation}
\renewcommand{\arraystretch}{1.1}
\begin{tabular}{@{}l l c c c@{}}
\toprule
\textbf{Benchmark} & \textbf{Method} & \textbf{Greedy} & \textbf{Default} & \textbf{Exploratory} \\
\midrule
\textbf{superblue1} & VP+DP 2.5 Flash & 37.51 $\pm$ 0.09 & \textbf{37.47 $\pm$ 0.11} & 37.64 $\pm$ 0.11 \\
\cmidrule{2-5}
& DP 4.3.0 (Baseline) & \multicolumn{3}{c}{37.95 $\pm$ 0.16} \\
\bottomrule
\end{tabular}
\end{table}

\subsection{Input Modality Ablation}
\label{appendix:modality_ablation}
We test how much each input channel contributes by removing either the chip canvas image or the textual context (the macro coordinates in text) from each in-context example, with top-stratified selection and $C{=}25$ fixed, on all eight Superblue designs with five seeds. The shuffled arm keeps the canvas images but pairs each one with a placement it did not come from. All four prompts improve over DREAMPlace 4.3.0 on all eight designs and differ from each other by under 0.8\% (\Cref{tab:appendix_modality_ablation}), so the gain does not depend on a single input channel. The shuffled arm matches the full prompt on every design, because the textual history already carries the exact macro coordinates.

\begin{table}[H]
\centering
\small
\caption{Input modality ablation on Superblue.
The values are gHPWL ($\times 10^6$, mean $\pm$ standard error over five seeds, lower is better).
VP+DP uses Gemini 3.6 Flash.
`Full' is the default multimodal prompt with both image and text.
`NoImg' removes the canvas image from every in-context example.
`NoTxt' removes the textual context from every in-context example.
`Shuffled' keeps the images but pairs each one with a placement it did not come from.
DREAMPlace 4.3.0 runs 2{,}000 rollouts.
The best result per design is in bold.}
\label{tab:appendix_modality_ablation}
\renewcommand{\arraystretch}{1.1}
\begin{tabularx}{\columnwidth}{@{}l *{5}{>{\centering\arraybackslash}X} @{}}
\toprule
\textbf{Benchmark} & \textbf{DP 4.3.0} & \textbf{VP+DP Full} & \textbf{VP+DP NoImg} & \textbf{VP+DP NoTxt} & \textbf{VP+DP Shuffled} \\
\midrule
superblue1 & 376.63$\pm$0.13 & \underline{372.85$\pm$0.34} & 374.18$\pm$0.36 & \textbf{372.73$\pm$0.39} & 373.54$\pm$0.38 \\
superblue3 & 453.75$\pm$0.33 & 438.14$\pm$0.42 & 438.31$\pm$0.73 & \textbf{436.81$\pm$0.22} & \underline{438.01$\pm$0.72} \\
superblue4 & 290.92$\pm$0.26 & 286.82$\pm$0.55 & \underline{285.88$\pm$0.22} & \textbf{284.64$\pm$0.59} & 285.91$\pm$0.21 \\
superblue5 & 400.88$\pm$0.30 & \textbf{397.90$\pm$0.41} & \underline{397.91$\pm$0.53} & 397.99$\pm$0.96 & 398.42$\pm$0.39 \\
superblue7 & 544.45$\pm$0.35 & 538.76$\pm$0.86 & \textbf{537.35$\pm$0.57} & 541.32$\pm$0.57 & \underline{537.64$\pm$0.90} \\
superblue10 & 714.47$\pm$1.18 & 683.25$\pm$1.22 & 684.01$\pm$1.31 & \underline{682.72$\pm$0.51} & \textbf{681.75$\pm$1.06} \\
superblue16 & 366.34$\pm$0.12 & 364.43$\pm$0.09 & \textbf{363.48$\pm$0.39} & 364.43$\pm$0.14 & \underline{364.10$\pm$0.29} \\
superblue18 & 219.24$\pm$0.07 & \underline{217.50$\pm$1.09} & 217.77$\pm$0.87 & \textbf{217.04$\pm$0.88} & 217.61$\pm$1.07 \\
\bottomrule
\end{tabularx}
\end{table}

\section{Prompt Details}
\label{appendix:prompts}
\subsection{Example Prompt}
\label{appendix:example_prompt}
We show a representative prompt from the \texttt{superblue1} DREAMPlace track. Repeated macro rows are omitted for readability, but the prompt structure and wording match the prompts used in our experiments.

\begin{tcolorbox}[breakable,colback=gray!5!white,colframe=gray!75!black,title=Prompt example: Superblue1]
\small
You are guiding a low-level macro placer for computer chip floorplanning. Your primary goal is to create a chip floorplan that minimizes wirelength (total interconnect length). Your task is to suggest rectangular regions for placing macros on the chip canvas, which has been divided into a grid. The low-level macro placer will choose the exact placement location within your suggested regions. If there is a macro in the netlist that you are not providing a suggestion for, the low-level macro placer will place that macro by itself.

\smallskip
The macros are grouped by colors based on their connectivity in the netlist graph, where macros with higher interconnectivity (more pin connections between them) are assigned similar colors. After macro placement, an analytical placer automatically positions the standard cells around them, so your macro placement decisions directly affect standard-cell placement quality.

\smallskip
This is a global optimization task where you need to consider:
\begin{itemize}[leftmargin=1.4em,itemsep=2pt,topsep=3pt]
    \item The impact of your suggested regions on macros that will be placed in the future.
    \item The overall arrangement of the selected macros that best supports a floorplan that minimizes wirelength (total interconnect length).
\end{itemize}

\medskip
\textbf{MACRO NAMES AND PROPERTIES FOR THIS NETLIST:}\par
\smallskip
\noindent
{\footnotesize
\setlength{\tabcolsep}{4pt}
\renewcommand{\arraystretch}{0.95}
\begin{minipage}[t]{0.48\linewidth}
\centering
\begin{tabular}{|l|l|l|}
\hline
\textbf{Macro} & \textbf{Color} & \textbf{WxH} \\
\hline
I68 & \#80e020 & 28 x 212 \\
DIL & \#e020e0 & 28 x 212 \\
IJW & \#c0e020 & 28 x 212 \\
H06 & \#a0e020 & 28 x 212 \\
FZH & \#c020e0 & 28 x 212 \\
DXK & \#e020e0 & 28 x 212 \\
\hline
\end{tabular}
\end{minipage}\hfill%
\begin{minipage}[t]{0.48\linewidth}
\centering
\begin{tabular}{|l|l|l|}
\hline
\textbf{Macro} & \textbf{Color} & \textbf{WxH} \\
\hline
EIQ & \#20c0e0 & 28 x 212 \\
ISU & \#6020e0 & 28 x 212 \\
FD4 & \#c0c020 & 28 x 212 \\
HNI & \#e0a020 & 28 x 212 \\
CQH & \#2020e0 & 28 x 212 \\
EIJ & \#2080e0 & 28 x 212 \\
\hline
\end{tabular}
\end{minipage}
\par}
\end{tcolorbox}

\begin{tcolorbox}[breakable,colback=gray!5!white,colframe=gray!75!black]
\small
\textbf{IMPORTANT PLACEMENT RULES:}
\begin{enumerate}[leftmargin=1.6em,itemsep=2pt,topsep=4pt]
    \item The chip canvas is 512$\times$512.
    \item Coordinate system:
    \begin{itemize}[leftmargin=1.4em,itemsep=1pt,topsep=2pt]
        \item Origin \texttt{(0,0)} is at the bottom-left corner.
        \item Top-left corner is \texttt{(0,512)}.
        \item Bottom-right corner is \texttt{(512,0)}.
        \item Top-right corner is \texttt{(512,512)}.
    \end{itemize}
    \item Suggested regions must be defined by bottom-left and top-right corners of the rectangle.
    \item Suggested regions must not overlap with each other.
    \item Suggestions are needed for these selected macros (name, size, color).
\end{enumerate}

\medskip
\textbf{PLACEMENT QUALITY METRICS:}
\begin{itemize}[leftmargin=1.4em,itemsep=2pt,topsep=3pt]
    \item Macro overlap must be zero (overlapping placements are invalid).
\end{itemize}

\medskip
\textbf{METRIC DIRECTIONS (REFERENCE ONLY):}
\begin{itemize}[leftmargin=1.4em,itemsep=2pt,topsep=3pt]
    \item Lower wirelength is better.
    \item Lower congestion is better.
    \item Lower density is better.
    \item Higher WNS and TNS is better.
\end{itemize}
\end{tcolorbox}

\begin{tcolorbox}[breakable,colback=gray!5!white,colframe=gray!75!black]
\small
\textbf{PREVIOUS PLACEMENT EPISODES:}

\smallskip
In each episode's canvas image, the selected macros are outlined in red and labeled with their names. Blue texture represents standard cells placed by the analytical placer. Dark gray rectangles are fixed I/O terminals.

\medskip
{\normalsize\textbf{Episode \#1}}\par
\smallskip
{\footnotesize\textbf{Position of Selected Macros:}}\par
\smallskip
{\footnotesize
\begin{itemize}[leftmargin=1.4em,itemsep=1pt,topsep=1pt]
    \item CQH: (262,300) to (290,512)
    \item EIJ: (148,0) to (176,212)
    \item DXK: (184,300) to (212,512)
    \item EIQ: (147,297) to (175,509)
    \item DIL: (83,10) to (111,222)
    \item ISU: (54,217) to (82,429)
    \item H06: (212,287) to (240,499)
    \item IJW: (46,6) to (74,218)
    \item I68: (176,4) to (204,216)
    \item HNI: (213,0) to (241,212)
    \item GKI: (119,278) to (147,490)
    \item FZH: (111,9) to (139,221)
\end{itemize}
}

\begin{center}
\includegraphics[width=0.82\linewidth]{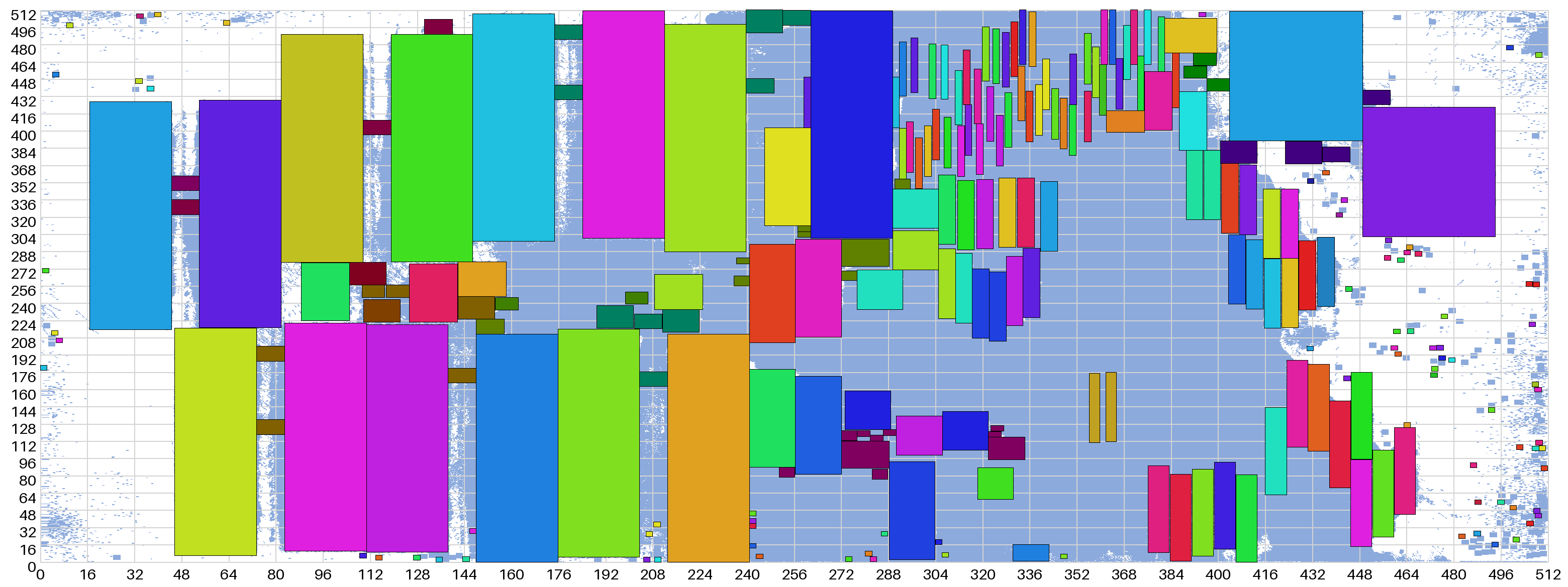}
\end{center}

{\footnotesize\textbf{Final Metrics:}}\par
\smallskip
{\footnotesize
\begin{itemize}[leftmargin=1.4em,itemsep=1pt,topsep=1pt]
    \item Wirelength: $3.93{\times}10^8$
    \item Congestion: $9.36{\times}10^{-1}$
    \item Density: $1.04$
    \item Macro overlap: $0$
\end{itemize}
}

\medskip
{\normalsize\textbf{Episode \#2}}\par
\smallskip
{\footnotesize\textbf{Position of Selected Macros:}}\par
\smallskip
{\footnotesize
\begin{itemize}[leftmargin=1.4em,itemsep=1pt,topsep=1pt]
    \item CQH: (243,284) to (271,496)
    \item EIJ: (150,244) to (178,456)
    \item DXK: (214,0) to (242,212)
    \item EIQ: (177,4) to (205,216)
    \item DIL: (178,300) to (206,512)
    \item ISU: (53,213) to (81,425)
    \item H06: (206,257) to (234,469)
    \item IJW: (122,277) to (150,489)
    \item I68: (121,17) to (149,229)
    \item HNI: (149,0) to (177,212)
    \item GKI: (93,0) to (121,212)
    \item FZH: (94,211) to (122,423)
\end{itemize}
}

\begin{center}
\includegraphics[width=0.82\linewidth]{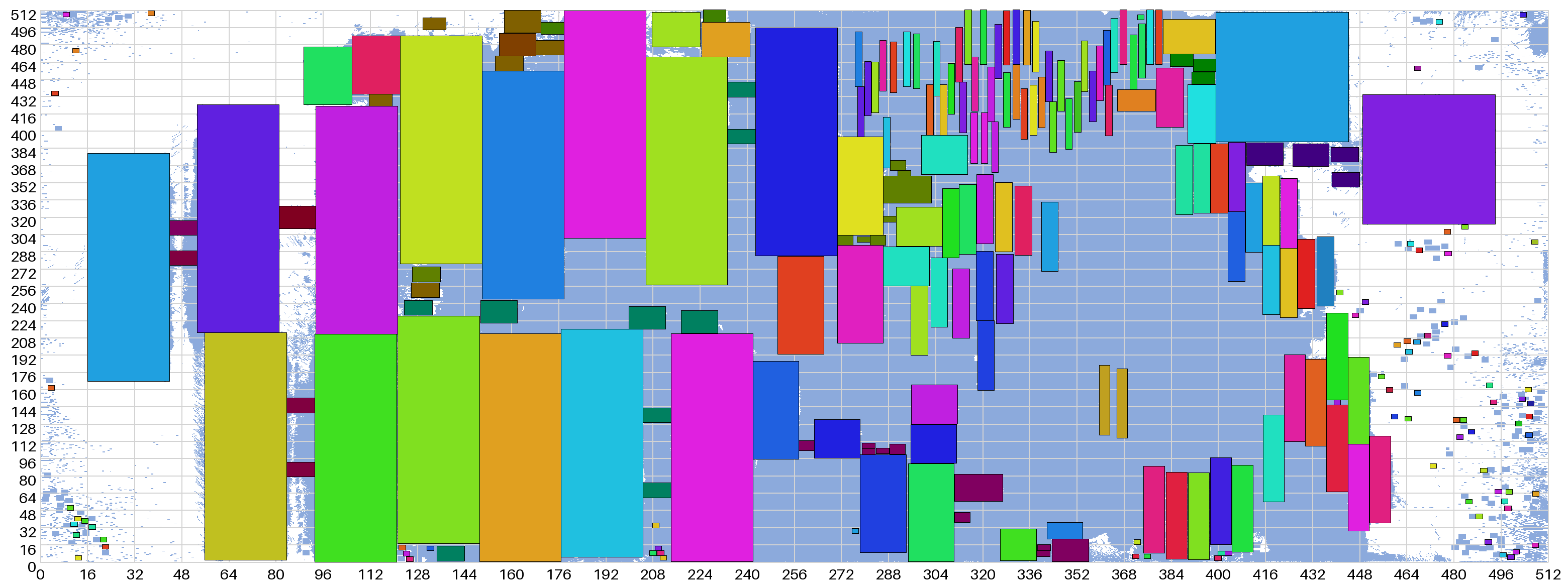}
\end{center}

{\footnotesize\textbf{Final Metrics:}}\par
\smallskip
{\footnotesize
\begin{itemize}[leftmargin=1.4em,itemsep=1pt,topsep=1pt]
    \item Wirelength: $4.20{\times}10^8$
    \item Congestion: $9.59{\times}10^{-1}$
    \item Density: $1.04$
    \item Macro overlap: $0$
\end{itemize}
}
\end{tcolorbox}

\begin{tcolorbox}[breakable,colback=gray!5!white,colframe=gray!75!black]
\small
\textbf{IMPORTANT OUTPUT FORMAT:}
\begin{enumerate}[leftmargin=1.6em,itemsep=2pt,topsep=4pt]
    \item All coordinates must be integers between 0 and 512.
    \item All regions must have non-zero width and height ($x_2 > x_1$ and $y_2 > y_1$).
    \item The orientation of macros cannot be changed. Do not try to rotate macros.
    \item All regions must be large enough to fit the macro while still being within the bounds of the canvas. For example, if a macro size is 3.1$\times$4.2, the region must be at least 4$\times$5.
    \item In the example below, replace text in square brackets with your own reasoning. Do not copy the text inside the brackets.
    \item Follow this example format exactly (without the dashed lines).
\end{enumerate}

\medskip
\textbf{HISTORY SUMMARY}\par
\smallskip
\begin{itemize}[leftmargin=1.4em,itemsep=3pt,topsep=2pt]
    \item \textbf{Placement Archetypes by Color Group:}
    \begin{itemize}[leftmargin=1.4em,itemsep=1pt,topsep=1pt]
        \item \char91 Identify 2--3 distinct placement strategies that appeared across episodes for major color groups \char93
        \item \char91 For each archetype: one representative episode + coordinates + wirelength \char93
        \item \char91 Note which strategy performed best and why \char93
    \end{itemize}
    \item \textbf{Largest Macros Analysis:}
    \begin{itemize}[leftmargin=1.4em,itemsep=1pt,topsep=1pt]
        \item \char91 Compare placement of largest macros in best vs worst episodes \char93
        \item \char91 Include coordinates and metrics only when there's clear performance difference \char93
    \end{itemize}
    \item \textbf{Spatial Patterns:}
    \begin{itemize}[leftmargin=1.4em,itemsep=1pt,topsep=1pt]
        \item \char91 Any adjacency, spacing, or positioning patterns that consistently helped performance \char93
        \item \char91 If no consistent pattern found, explain why patterns varied \char93
    \end{itemize}
    \item \textbf{Standard Cell Distribution:}
    \begin{itemize}[leftmargin=1.4em,itemsep=1pt,topsep=1pt]
        \item \char91 Observe light blue standard cells in canvas images - note where they cluster or get blocked \char93
        \item \char91 Which macro arrangements allowed standard cells to spread evenly vs get congested? \char93
    \end{itemize}
    \item \textbf{Key Principles (at least 3):}
    \begin{itemize}[leftmargin=1.4em,itemsep=1pt,topsep=1pt]
        \item \char91 Concise placement rules derived from analysis above \char93
        \item \char91 Cite evidence from history; if evidence is weak, mark as ``hypothesis'' \char93
    \end{itemize}
\end{itemize}

\medskip
\textbf{STRATEGY}\par
\smallskip
\begin{itemize}[leftmargin=1.4em,itemsep=3pt,topsep=2pt]
    \item \textbf{Overall Plan:} \char91 2--3 sentences describing your placement approach \char93
    \item \textbf{Improvement on Best Archetype:} \char91 Propose exactly one concrete change to the best pattern (e.g., shift position, adjust spacing, try different canvas region); if evidence is weak, label it ``exploratory'' but still commit \char93
    \item \textbf{Color Group Organization:} \char91 How different color groups will be distributed across the canvas \char93
    \item \textbf{Wirelength Rationale:} \char91 Why this arrangement minimizes wirelength \char93
\end{itemize}

\medskip
\textbf{REGIONS}\par
\smallskip
\begin{itemize}[leftmargin=1.4em,itemsep=2pt,topsep=2pt]
    \item \char91 List regions in DECREASING SIZE ORDER (largest macros first) to avoid overlaps \char93
    \item \char91 Format: \texttt{MACRO\_NAME: (x1,y1) to (x2,y2)} \char93
\end{itemize}

\smallskip
\textbf{Example format:}
\begin{itemize}[leftmargin=1.4em,itemsep=1pt,topsep=2pt]
    \item \texttt{ABC: (120,200) to (150,220)}
    \item \texttt{DEF: (45,310) to (65,326)}
    \item \texttt{RST: (280,150) to (288,162)}
\end{itemize}

\smallskip
Make your descriptions as detailed as possible so that the floorplan can be constructed without ambiguity.
\end{tcolorbox}

\subsection{Example Gemini Response}
\begin{tcolorbox}[breakable,colback=gray!5!white,colframe=gray!75!black]
\small
\textbf{Abridged response excerpt from the same prompt.}

\medskip
\textbf{HISTORY SUMMARY}
\begin{itemize}[leftmargin=1.4em,itemsep=3pt,topsep=3pt]
    \item \textbf{Placement Archetypes by Color Group:}
    \begin{itemize}[leftmargin=1.4em,itemsep=2pt,topsep=1pt]
        \item \textbf{Large Macros ($28\times212$):} The dominant archetype for these macros is dense vertical stacking in columns, typically occupying the left side of the chip. In Episode \#3 (WL: $3.92{\times}10^8$), these macros formed a very compact, multi-column block on the left. This pattern consistently yielded better wirelength. In contrast, Episode \#2 (WL: $4.20{\times}10^8$) showed more horizontal dispersion of these macros, leading to larger gaps and higher wirelength.
        \item \textbf{Top-Right Block (HCL, O9Y - $46\times121$):} These two macros, often with distinct colors, were consistently placed together in the top-right corner. In Episode \#3, O9Y (397,390) to (443,511) and HCL (452,316) to (498,437) were placed compactly, suggesting they are either I/O related or have high interconnectivity with each other and minimal with the main macro block.
        \item \textbf{Medium/Small Macros:} These macros were generally used to fill the remaining spaces, often forming smaller clusters or extending the larger blocks.
    \end{itemize}

    \item \textbf{Largest Macros Analysis:} The placement of the largest macros ($28\times212$) was critical. In the best episode (\#3, WL: $3.92{\times}10^8$), these macros were arranged into a very tight, rectangular block spanning almost the entire height of the left half of the chip. This dense packing minimized the overall footprint of the largest components, reducing their internal interconnect wirelength and creating a large, contiguous area for standard cells.

    \item \textbf{Key Principles:}
    \begin{enumerate}[leftmargin=1.6em,itemsep=2pt,topsep=1pt]
        \item \textbf{Maximize Macro Block Density:} Group the largest, most connected macros into a single, compact, and geometrically regular block to minimize inter-macro wirelength.
        \item \textbf{Utilize Chip Edges/Corners:} Place distinct macro groups, like HCL/O9Y, in corners to efficiently use peripheral space while maintaining a clear central area.
        \item \textbf{Fill Gaps with Smaller Macros by Color:} Systematically fill remaining spaces with smaller macros, prioritizing color-based clustering to localize connections and maintain overall floorplan density.
    \end{enumerate}
\end{itemize}

\medskip
\textbf{STRATEGY}
\begin{itemize}[leftmargin=1.4em,itemsep=3pt,topsep=3pt]
    \item \textbf{Overall Plan:} My strategy is to create a highly dense and rectangular block of the largest macros ($28\times212$) on the left side of the chip, spanning most of its height. The distinct HCL and O9Y macros will be placed compactly in the top-right corner. The remaining medium and small macros will then be systematically placed in contiguous blocks in the central and bottom-right regions, prioritizing color-based adjacency to minimize local interconnect.
    \item \textbf{Improvement on Best Archetype:} I will refine the ``dense block'' principle by ensuring the $28\times212$ macros are arranged in perfectly aligned vertical columns, minimizing any staggering or small gaps between them (exploratory). This aims for a more uniform macro-standard cell interface, which could improve routing predictability.
    \item \textbf{Wirelength Rationale:} This arrangement minimizes wirelength by physically co-locating highly connected macros, inferred by color and grouping. The dense packing of large macros ensures their internal connections are short.
\end{itemize}

\medskip
\textbf{REGIONS}
\begin{itemize}[leftmargin=1.4em,itemsep=1pt,topsep=3pt]
    \item O9Y: (466,391) to (512,512)
    \item HCL: (420,391) to (466,512)
    \item FOF: (380,470) to (398,503)
    \item D8W: (375,391) to (385,446)
    \item BDJ: (385,391) to (395,446)
    \item EWA: (0,0) to (28,212)
    \item ISU: (0,212) to (28,424)
    \item IJW: (28,0) to (56,212)
    \item DIL: (28,212) to (56,424)
    \item FZH: (56,0) to (84,212)
    \item GKI: (56,212) to (84,424)
    \item I68: (84,212) to (112,424)
    \item EIJ: (84,0) to (112,212)
    \item HNI: (112,0) to (140,212)
    \item H06: (112,212) to (140,424)
\end{itemize}
\end{tcolorbox}

\newpage
\section*{NeurIPS Paper Checklist}

\begin{enumerate}

\item {\bf Claims}
    \item[] Question: Do the main claims made in the abstract and introduction accurately reflect the paper's contributions and scope?
    \item[] Answer: \answerYes{}
    \item[] Justification: The abstract and introduction state that \method{} improves placement quality on open-source benchmarks using a VLM without fine-tuning, with specific win counts and peak improvement percentages that are substantiated by the experimental results in \Cref{sec:experiments}.

\item {\bf Limitations}
    \item[] Question: Does the paper discuss the limitations of the work performed by the authors?
    \item[] Answer: \answerYes{}
    \item[] Justification: \Cref{sec:limitations} states that the search loop optimizes only gHPWL and that post-route PPA is evaluated only after the search. It also covers the VLM latency and API cost, the 256-macro limit, and future work on more macros and multi-objective PPA optimization.

\item {\bf Theory assumptions and proofs}
    \item[] Question: For each theoretical result, does the paper provide the full set of assumptions and a complete (and correct) proof?
    \item[] Answer: \answerNA{}
    \item[] Justification: The paper does not include theoretical results. Our contribution is an empirical framework evaluated through experiments.

    \item {\bf Experimental result reproducibility}
    \item[] Question: Does the paper fully disclose all the information needed to reproduce the main experimental results of the paper to the extent that it affects the main claims and/or conclusions of the paper (regardless of whether the code and data are provided or not)?
    \item[] Answer: \answerYes{}
    \item[] Justification: We specify the VLM used for each experiment (Gemini 2.5, 3.5, or 3.6 Flash, and Qwen3-VL), all hyperparameters (context length, grid resolution, anchor weight, rollout count), random seeds, and benchmark sources. Full prompt templates and hyperparameter details are provided in the appendix. Code is included in the supplementary material.

\item {\bf Open access to data and code}
    \item[] Question: Does the paper provide open access to the data and code, with sufficient instructions to faithfully reproduce the main experimental results, as described in supplemental material?
    \item[] Answer: \answerYes{}
    \item[] Justification: Code is included in the supplementary material. All benchmarks used (ICCAD 2015 Superblue, ISPD 2005, ICCAD 2004, Ariane) are publicly available. Instructions for reproducing the experiments are provided with the code.

\item {\bf Experimental setting/details}
    \item[] Question: Does the paper specify all the training and test details (e.g., data splits, hyperparameters, how they were chosen, type of optimizer) necessary to understand the results?
    \item[] Answer: \answerYes{}
    \item[] Justification: \Cref{sec:experiments} specifies the VLM, grid resolution, context length, rollout count, and random seeds. Full hyperparameter tables for both DREAMPlace and ChiPFormer settings are provided in the appendix.

\item {\bf Experiment statistical significance}
    \item[] Question: Does the paper report error bars suitably and correctly defined or other appropriate information about the statistical significance of the experiments?
    \item[] Answer: \answerYes{}
    \item[] Justification: All main results report mean and standard error across random seeds (three or five, stated per table) with 2,000 rollouts each. The source of variability (random seed / initialization) and the error metric (standard error of the mean) are stated in each table caption.

\item {\bf Experiments compute resources}
    \item[] Question: For each experiment, does the paper provide sufficient information on the computer resources (type of compute workers, memory, time of execution) needed to reproduce the experiments?
    \item[] Answer: \answerYes{}
    \item[] Justification: We report per-rollout wall-clock time and total runtime on a single A100 GPU for each Superblue benchmark (\Cref{tab:runtime_overhead}), along with VLM API cost estimates (\Cref{app:api_cost}). The overhead ranges from 2\%--7\% over base DREAMPlace.

\item {\bf Code of ethics}
    \item[] Question: Does the research conducted in the paper conform, in every respect, with the NeurIPS Code of Ethics \url{https://neurips.cc/public/EthicsGuidelines}?
    \item[] Answer: \answerYes{}
    \item[] Justification: The research conforms with the NeurIPS Code of Ethics. We use publicly available benchmarks, a commercially available VLM API, and do not involve human subjects.

\item {\bf Broader impacts}
    \item[] Question: Does the paper discuss both potential positive societal impacts and negative societal impacts of the work performed?
    \item[] Answer: \answerNA{}
    \item[] Justification: This work improves macro placement in chip design, a specialized EDA task with no direct negative societal implications. It does not introduce new capabilities for surveillance, disinformation, or other harmful applications.

\item {\bf Safeguards}
    \item[] Question: Does the paper describe safeguards that have been put in place for responsible release of data or models that have a high risk for misuse (e.g., pre-trained language models, image generators, or scraped datasets)?
    \item[] Answer: \answerNA{}
    \item[] Justification: We do not release pre-trained models or scraped datasets. \method{} uses an existing commercial VLM API and publicly available EDA benchmarks, posing no risk for misuse.

\item {\bf Licenses for existing assets}
    \item[] Question: Are the creators or original owners of assets (e.g., code, data, models), used in the paper, properly credited and are the license and terms of use explicitly mentioned and properly respected?
    \item[] Answer: \answerYes{}
    \item[] Justification: We cite the original papers for all benchmarks (ICCAD 2015, ISPD 2005, ICCAD 2004, Ariane), DREAMPlace, and ChiPFormer. URLs to public repositories are provided where applicable.

\item {\bf New assets}
    \item[] Question: Are new assets introduced in the paper well documented and is the documentation provided alongside the assets?
    \item[] Answer: \answerYes{}
    \item[] Justification: We release the \method{} code in the supplementary material with documentation for reproducing experiments.

\item {\bf Crowdsourcing and research with human subjects}
    \item[] Question: For crowdsourcing experiments and research with human subjects, does the paper include the full text of instructions given to participants and screenshots, if applicable, as well as details about compensation (if any)?
    \item[] Answer: \answerNA{}
    \item[] Justification: This work does not involve crowdsourcing or research with human subjects.

\item {\bf Institutional review board (IRB) approvals or equivalent for research with human subjects}
    \item[] Question: Does the paper describe potential risks incurred by study participants, whether such risks were disclosed to the subjects, and whether Institutional Review Board (IRB) approvals (or an equivalent approval/review based on the requirements of your country or institution) were obtained?
    \item[] Answer: \answerNA{}
    \item[] Justification: This work does not involve human subjects research.

\item {\bf Declaration of LLM usage}
    \item[] Question: Does the paper describe the usage of LLMs if it is an important, original, or non-standard component of the core methods in this research? Note that if the LLM is used only for writing, editing, or formatting purposes and does \emph{not} impact the core methodology, scientific rigor, or originality of the research, declaration is not required.
    \item[] Answer: \answerYes{}
    \item[] Justification: The core method uses a Vision-Language Model (Gemini Flash models and Qwen3-VL) as a high-level planner for macro placement. The VLM, its role, and its usage are described in detail throughout \Cref{sec:method}.

\end{enumerate}

\end{document}